\pdfoutput=1
\UseRawInputEncoding
\documentclass[11pt]{article}

\usepackage[final]{acl}

\usepackage{times}
\usepackage{latexsym}

\usepackage[T1]{fontenc}

\usepackage[utf8]{inputenc}

\usepackage{microtype}

\usepackage{inconsolata}

\usepackage{graphicx}

\title{\raisebox{-0.22\height}{\includegraphics[height=1.35\fontcharht\font`X]{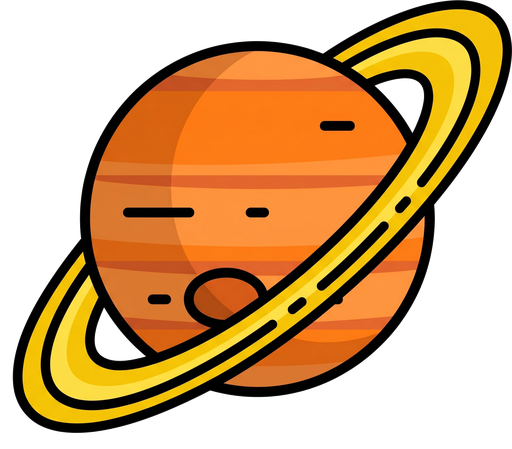}}\hspace{2pt}\method: Symbolic Spatial Reasoning for Multi-Perspective Grounding}

\author{
  \textbf{Danial Kamali\textsuperscript{1}},
  \textbf{Tanawan Premsri\textsuperscript{1}},
  \textbf{Shreya Rajpal\textsuperscript{1}},
\\
  \textbf{Amir Zadeh\textsuperscript{2}},
  \textbf{Chuan Li\textsuperscript{2}},
  \textbf{Parisa Kordjamshidi\textsuperscript{1}}
\\
  \textsuperscript{1}Michigan State University,
  \textsuperscript{2}Lambda Labs
\\
  \small{
    \texttt{\{kamalida, premsrit, rajpalsh, kordjams\}@msu.edu}
  }
}

\definecolor{green}{rgb}{0.0, 0.6, 0.0}
\definecolor{lightgreen}{HTML}{2DC75C}
\definecolor{lightblue}{HTML}{4DA8DA}
\definecolor{lightpurple}{HTML}{9A8AD2}
\definecolor{lightgray}{HTML}{F2F2F2}
\definecolor{program}{HTML}{f6dfde}
\definecolor{perceptual}{HTML}{ffe2c9}
\definecolor{symbolic}{HTML}{c6dcff}

\usepackage[most]{tcolorbox}
\usepackage{wrapfig}
\usepackage{multirow}
\usepackage{caption}
\usepackage{microtype}
\usepackage{xcolor}
\usepackage{graphicx}
\usepackage{amsmath}
\usepackage{enumitem}
\usepackage{xspace}
\usepackage[table]{xcolor}
\usepackage{tikz}
\usetikzlibrary{arrows.meta}
\usepackage{adjustbox}
\usepackage{booktabs}
\usepackage{amsfonts}
\usepackage{url}
\usepackage{booktabs}
\usepackage{amsfonts}
\usepackage{nicefrac}
\usepackage{microtype}
\usepackage{xcolor}
\usepackage{pifont}
\usepackage[utf8]{inputenc}
\usepackage[T1]{fontenc}
\usepackage{hyperref}
\usepackage{url}
\usepackage{booktabs}
\usepackage{amsfonts}
\usepackage{nicefrac}
\usepackage{microtype}
\usepackage{xcolor}
\usepackage{graphicx}
\usepackage{amsmath}
\usepackage{enumitem}
\usepackage{xspace}
\usepackage[table]{xcolor}
\usepackage{tikz}
\usetikzlibrary{arrows.meta}
\usepackage{adjustbox}
\definecolor{codegreen}{rgb}{0,0.6,0}
\definecolor{codegray}{rgb}{0.5,0.5,0.5}
\definecolor{codepurple}{rgb}{0.58,0,0.82}
\definecolor{backcolour}{rgb}{0.95,0.95,0.92}

\definecolor{planetbg}{HTML}{FFF0D6}
\definecolor{planetframe}{HTML}{F5B13A}
\tcbset{
  colback=planetbg,
  colframe=planetframe,
  boxrule=0.4pt,
  arc=2pt,
  left=6pt,
  right=6pt,
  top=6pt,
  bottom=6pt,
}
\tcbset{
  inlinebox/.style={
    enhanced,
    on line,
    tcbox raise base,
    boxrule=0pt,
    frame hidden,
    rounded corners,
    colback=#1!30,
    boxsep=0.5pt,
    left=2pt,right=2pt,top=0pt,bottom=0pt,
  }
}

\newtcolorbox{calloutbox}[1][]{
  enhanced,
  breakable,
  width=\columnwidth,
  title=\textbf{#1},
  fonttitle=\bfseries,
}

\newtcolorbox{promptbox}[1][]{
  enhanced,
  width=\columnwidth,
  colback=lightgray,
  colframe=black,
  boxrule=1.4pt,
  arc=4pt,
  outer arc=4pt,
  left=6pt,right=6pt,top=6pt,bottom=6pt,
  #1
}

\lstdefinestyle{mypython}{
    backgroundcolor=\color{backcolour},
    commentstyle=\color{codegreen},
    keywordstyle=\color{magenta},
    numberstyle=\tiny\color{codegray},
    stringstyle=\color{codepurple},
    basicstyle=\ttfamily\footnotesize,
    breakatwhitespace=false,
    breaklines=true,
    captionpos=b,
    keepspaces=true,
    numbers=left,
    numbersep=5pt,
    showspaces=false,
    showstringspaces=false,
    showtabs=false,
    tabsize=2
}
\lstdefinestyle{promptstyle}{
  basicstyle=\ttfamily\scriptsize,
  breaklines=true,
  breakatwhitespace=false,
  columns=fullflexible,
  keepspaces=true,
  showstringspaces=false,
  upquote=true,
  frame=single,
  xleftmargin=0pt,
  aboveskip=4pt,
  belowskip=4pt
}

\newcommand{\cmark}{\ding{51}\xspace}
\newcommand{\xmark}{\ding{55}\xspace}

\newcommand{\method}{{SATURN}\xspace}
\newcommand{\bench}{3D FORCE\xspace}
\newcommand{\benchfull}{3D Frames Of Reference Composition Evaluation\xspace}

\newcommand{\refbench}{\bench-REF\xspace}
\newcommand{\refbenchshort}{REF\xspace}
\newcommand{\puzzlebench}{\bench-SAG\xspace}
\newcommand{\puzzlebenchshort}{SAG\xspace}

\usepackage{wrapfig}

\begin{document}
\maketitle
\begin{abstract}
Vision-Language Models (VLMs) remain unreliable when spatial reasoning requires composing relations whose meanings depend on frames of reference.
Existing tool-augmented spatial reasoning methods make reasoning more explicit, but often rely on low-level geometric procedures and hard binary decisions over noisy perception.
We propose SATURN, a neuro-symbolic framework for perspective-aware compositional spatial reasoning. SATURN reconstructs an approximate 3D scene, derives soft perspective-aware spatial predicates, and composes them with a training-free Pythonic symbolic executor, separating perception from reasoning while preserving uncertainty through multi-hop inference.
We also introduce 3D FORCE, a diagnostic benchmark that controls reasoning depth, view, and perspective composition for spatial arrangement grounding (SAG) and referring expression grounding (REF).
On 3D FORCE, VLMs and spatially trained models degrade sharply as depth and perspective complexity increase, whereas SATURN degrades the least and outperforms every baseline at each depth.
On the real-world MindCube benchmark, SATURN achieves \(78.06\%\) overall accuracy, outperforming the strongest baseline by \(14\) percentage points.

\end{abstract}
\noindent\textbf{Code:} \url{https://github.com/HLR/SATURN}
\section{Introduction}
\begin{figure}[t]

    \centering
    \includegraphics[width=0.95\linewidth]{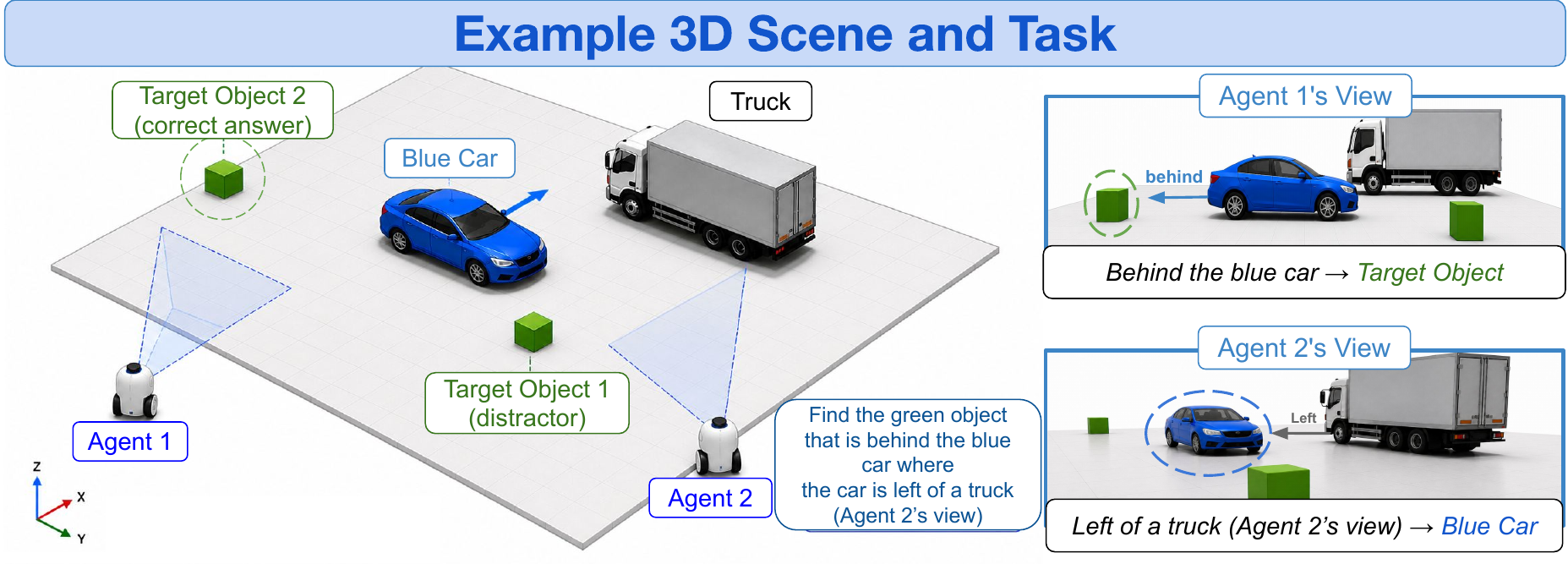}
    \caption{An example illustrating 3D scene understanding with multi-perspective reasoning.}
    \label{fig:FOR}
\end{figure}

Many visually grounded instructions require resolving spatial relations in different frames of reference (FoRs), such as the camera's view, an agent's view, or an object's intrinsic orientation as shown in Figure~\ref{fig:FOR}.
A relation such as ``left of the truck'' can change depending on whether the relation is interpreted from the camera, the agent, or the truck itself.
We define an FoR as the perspective induced by an anchor's 3D position and orientation; cameras and oriented objects can therefore induce distinct FoRs.

Existing spatial benchmarks have demonstrated VLMs' failure in spatial reasoning~\citep{yang2025mmsi,cvbench}.
Pure neural VLMs implicitly encode spatial relations, making it difficult to ensure that the same relation is applied consistently across different objects, views, and FoRs~\citep{zhang2026spinbenchperspectiverotationlens}.
Explicit reasoning approaches, including neuro-symbolic pipelines and tool-augmented agents, make spatial computation more exact.
However, in practice, these approaches often require low-level geometric programming and rely on pipelines of crisp decisions during multi-hop reasoning, making them less robust when perceptual outputs are noisy.
To overcome these limitations, we introduce \textbf{\method}, a neuro-symbolic framework for 3D perspective-aware compositional reasoning.
Rather than asking VLMs to infer multi-hop spatial relations implicitly or generate low-level geometric code, \method elevates the reasoning from raw 3D states to soft FoR-aware predicates.
This abstraction helps downstream programs compose relations without re-implementing geometry or thresholding perceptual evidence.
\method first approximates a coarse 3D scene using neural perception and derives predicates such as \texttt{left} and \texttt{front} under different FoRs.
A Pythonic symbolic executor then composes these predicates to answer multi-hop spatial queries.

While we target perspective-aware compositional spatial grounding, existing benchmarks do not cleanly isolate this capability.
Current spatial reasoning benchmarks largely emphasize either primitive spatial judgments, such as relative position, orientation, depth, and perspective-taking~\citep{zhang2026spinbenchperspectiverotationlens, wang2025spatial457, viewspatial}, or broad embodied, navigational, and long-horizon tasks~\citep{li2025industrynavexploringspatialreasoning, sohn2025embodied4cmeasuringmattersembodied, zhao2026espire, ma-etal-2026-breaking}.
Between these settings lies an intermediate capability: \emph{perspective-aware compositional spatial grounding}, where a model must ground objects while resolving the frame of reference associated with each spatial relation and composing multiple relations into a single decision. For example, the spatial component of an instruction such as ``Pick up the cup and place it on the desk on your left and in front of me'' requires resolving \emph{left} in the agent's FoR, \emph{front} in the user's FoR, and composing these relations over the relevant objects before any action can be taken.

To evaluate this intermediate reasoning layer, we introduce \benchfull (\textbf{\bench}), a diagnostic benchmark that controls relation depth, viewpoint, and FoR composition.
\bench contains two complementary subsets: \textsc{\puzzlebenchshort}, which evaluates whether a multi-hop spatial arrangement is satisfied, and \textsc{\refbenchshort}, which evaluates whether a model can ground the target object described by a perspective-aware compositional referring expression. On \bench, \method consistently outperforms general-purpose VLMs, spatially trained VLMs, and tool-augmented baselines across both SAG and REF.
This gain is most pronounced in settings that require composing relations across multiple objects, views, and FoRs.
While spatial-focused tuned VLMs improve local spatial judgments, their performance degrades as reasoning depth and FoR composition increase.
This suggests that current spatial supervision improves relation recognition more than systematic relation composition across FoRs.
We further evaluate \method on real-world multi-view benchmarks MMSI~\citep{yang2025mmsi} and MindCube~\citep{yin2025spatial}, showing its gains beyond the controlled setting of \bench.

In summary, we make three contributions. 1) We introduce SATURN, a neuro-symbolic framework that converts noisy 3D visual perception into reusable soft spatial predicates defined under camera-centric, object-centric, and virtual-viewer FoRs, then composes these predicates through Pythonic symbolic execution.
2) We introduce 3D FORCE, a diagnostic benchmark that isolates perspective-aware relation composition across spatial arrangement and referring-expression grounding while controlling relation depth, topology, view count, and FoR assignment.
3) We show that current VLMs become less reliable as relation depth and FoR composition increase. \method degrades the least while outperforming VLMs, spatially trained VLMs, and tool-augmented baselines on 3D FORCE, and these gains transfer to the real-world MindCube and MMSI benchmarks.

\begin{figure*}[t!]
    \centering
    \vspace{-1mm}
    \input{figures/main/pipeline/pipeline}
    \vspace{-5mm}
    \caption{\method overview. The framework combines scene estimation, perceptual grounding, geometry-based computation of spatial predicates, and symbolic program execution for FoR-aware compositional reasoning.}
    \label{fig:saturn}

\end{figure*}
\section{Related Work}
\label{sec:related_work}

\textbf{Spatial Reasoning in Vision-Language Models.}
Recent studies show that VLMs struggle with spatial reasoning, including relative position, depth, orientation, object localization, FoR ambiguity, and multi-view reasoning~\citep{kamath2023s,wang2024picture,liumllms25large,erik25mmspatial,wang2025spatial457,zhang2025do,yu2025far,yang2025mmsi,yin2025spatial}.
A line of work addresses these limitations by fine-tuning models on spatial-focused data, incorporating geometric priors such as depth and 3D structure during training, or injecting linguistic structure for grounded compositional generalization~\citep{chen2024spatialvlm,cheng2024spatialrgpt,wu2025spatial,li2025spatialladder,wu2025reinforcing,ma2025spatialllm,ma2025spatialreasoner,fan2025vlm,kamali-kordjamshidi-2023-syntax}.
While these methods improve performance on targeted benchmarks, they still encode spatial relations as learned model priors rather than deriving them explicitly from scene geometry, which can limit generalization beyond the training distribution~\citep{GCA,mirzaee-kordjamshidi-2023-disentangling}. \method instead externalizes spatial reasoning by reconstructing a coarse 3D scene, deriving FoR-aware predicates, and composing them through symbolic execution.

\noindent\textbf{Tool-Augmented Spatial Reasoning.}
Neuro-symbolic frameworks vary in how they integrate neural and symbolic components~\citep{sinha2026neurosymbolicframeworks,faghihi2021domiknows,rajpal2026spatialreasoningmodalityswitching}; recent work also uses agents to generate neuro-symbolic programs from task descriptions~\citep{nafar2026agentic}.
Tool-augmented neuro-symbolic methods translate visual queries into executable programs or tool calls over objects and image regions~\citep{Gupta2022VisProg,
suris2023vipergpt}. Recent methods extend visual programming toward 3D reasoning, but differ from \method in the level of abstraction exposed to the reasoner: VADAR~\citep{marsili2025visual} and TVP~\citep{wu2025transductivevisualprogrammingevolving} reason over 3D APIs, GCA~\citep{GCA} uses agentic tool calls to extract 3D object properties and invoke Python tools, pySpatial~\citep{pySpatial} combines 3D reconstruction with Python-based novel-view synthesis for perspective-aware reasoning, and TIGeR~\citep{tiger} fine-tunes a VLM to generate geometric tool calls rather than prompting a frozen model. In contrast, \method does not generate programs to perform low-level geometric calculations. It first derives soft FoR-aware spatial predicates from estimated scene state (objects and cameras), such as \textit{left} and \textit{behind}, under camera-centric, object-centric, or custom FoRs. The generated program then composes these soft predicate scores in Python, following NePTune~\citep{kamali2026neptune}. FoR-SALE~\citep{premsri2026forsaleframereferenceguidedspatial} also treats FoR explicitly, but for image editing: it converts intrinsic spatial expressions to the camera perspective and uses layout mismatches to guide diffusion-based corrections.

\noindent\textbf{Spatial Reasoning Benchmarks.}
Recent spatial reasoning benchmarks evaluate different levels of spatial understanding. Atomic benchmarks typically focus on primitive spatial abilities, such as spatial relation recognition, egocentric direction, relative position, depth estimation, and object orientation~\citep{spatialsense,kamath2023s,gsrbench,embspatialbench,basicspatialabilities,zhang2025do,mirzaee-etal-2021-spartqa,premsri-kordjamshidi-2025-forest,viewspatial,zhang2026spinbenchperspectiverotationlens,yang2025mmsi,yin2025spatial,allanglesbench,gholami2025spatial}.
Recent benchmarks evaluate broader embodied or long-horizon spatial skills, such as 3D awareness, 6D spatial relations, object localization, counting, functional knowledge, multi-view fusion, dynamic scene understanding, and perspective-aware reasoning~\citep{3dsrbench,wang2025spatial457,sphere,space-10,jia2025omnispatial,sat,legopuzzles,videomsr,s3bench}.
\textsc{\bench} fills a gap between primitive spatial tests and broad embodied tasks by evaluating compositional object-level spatial reasoning under camera-centric and object-centric FoRs.
It explicitly controls hop count, relation topology, number of views, and the FoR associated with each relation.
This benchmark consists of two complementary subsets: \refbenchshort for frame-aware referring expressions and \puzzlebenchshort for multi-view spatial arrangement grounding.
Together, they assess whether models can compose mixed-frame spatial relations within a single reasoning problem.

\section{Method}

We introduce \method, a neuro-symbolic framework for answering
perspective-aware spatial questions. As shown in Figure~\ref{fig:saturn}, given a natural-language query \(Q\) and a visual context \(C\), \method decomposes the reasoning into three stages. First, query-guided scene estimation identifies the objects required for the query and estimates their 3D positions and orientations. In addition, when the query states facts about the camera setup, such as two views sharing a position or being separated by a rotation, \method uses these facts to refine the estimated camera states. Second, the spatial engine converts the resulting object and camera states into soft spatial predicates (i.e., probability scores) under camera-centric, object-centric, or virtual-viewer frames of reference. Finally, a symbolic executor runs an LLM-generated Python program over semantic and spatial predicates.

\subsection{Query-Guided Scene Estimation}
\label{sec:scene_estimation}

The first stage estimates the query-relevant geometric state of the scene.
A VLM-based parser reads the query, given the context, and identifies the candidate objects, their descriptions, and the visible view needed for reasoning. The object descriptions are passed
to an object grounder, which detects candidate object instances in the visual context.

Let \(N\) be the number of candidate object instances kept after grounding,
and let \(V\) be the number of images (e.g., cameras). For
each candidate object \(i\), \method estimates a 3D position
\(\mathbf{x}_i\) and orientation \(\mathbf{R}_i\). For each selected camera
view \(v\), \method estimates an initial camera position \(\mathbf{x}_v^0\)
and orientation \(\mathbf{R}_v^0\) from the images. We denote these object
and camera states as
\[
P=\{(\mathbf{x}_i,\mathbf{R}_i)\}_{i=1}^{N},
\qquad
\Gamma^0=\{(\mathbf{x}_v^0,\mathbf{R}_v^0)\}_{v=1}^{V}.
\]
Here, \(P\) is the set of candidate object states, while \(\Gamma^0\) is
the set of initial image-derived camera states before any text-stated pose
constraints are applied. Together, \(P\) and \(\Gamma^0\) form the geometric
scene state passed to the spatial engine.

\paragraph{Text-Stated Pose Constraints}
\label{sec:pose_constraints}
Some multi-view queries explicitly describe how the camera views are
arranged. For example, a query may state that two images are taken from
the same position, or that one view is rotated by a fixed angle relative
to another. These statements provide constraints on the camera poses that
can improve the image-derived camera estimates. \method extracts these statements as pose constraints and uses them to refine the initial camera states \(\Gamma^0\) before computing spatial predicates. We consider two types of constraints: rotation constraints, which specify relative yaw (horizontal orientation) between camera views, and same-position constraints, which specify that multiple views share a camera position.

For each rotation constraint, \method refines the orientation of the constrained view by setting its yaw to the stated offset relative to its anchor view. For each same-position group, \method  moves all cameras in the group to the position of the group's lowest-indexed view. Camera views without extracted constraints keep their image-derived poses. The refined camera states \(\Gamma\), together with the object states \(P\), are then passed to the spatial engine. Additional details on constraint extraction and anchor-based propagation are provided in Appendix~\ref{app:pose_constraints}.
\subsection{Spatial Engine}
\label{sec:spatial_engine}

The spatial engine converts the geometric scene state \((P,\Gamma)\) into
soft qualitative spatial predicates. The key challenge is that spatial
relations depend on the FoR: object X may be to the left
of object Y from one viewpoint but to the right from another. \method handles this by representing each frame of reference as a local coordinate system attached to an anchor \(a=(\mathbf{x}_a,\mathbf{R}_a)\), where \(\mathbf{x}_a\) is the anchor's 3D position and \(\mathbf{R}_a\) is its orientation. The anchor can be a camera, an object, or a virtual viewer. As a result, camera-centric relations use a camera anchor, object-centric relations use the reference object as the anchor, and virtual-viewer relations use the virtual viewpoint as the anchor.

To define predicates uniformly across these cases, let
\(\mathcal{E}=\{e_1,\ldots,e_M\}\) denote the entities that can participate
in spatial predicates, including candidate objects, camera views, and
optional virtual viewers. Each entity \(e_i\) has an associated state
\((\mathbf{x}_i,\mathbf{R}_i)\), where \(\mathbf{x}_i\) is its 3D position
and \(\mathbf{R}_i\) is its orientation. For a target entity \(e_i\), a reference entity \(e_j\), and an anchor
\(a\), the spatial engine first expresses the displacement from \(e_j\) to
\(e_i\) in the local coordinate system of \(a\):
\[
\Delta^{a}_{ij}=\mathbf{R}_a^\top(\mathbf{x}_i-\mathbf{x}_j).
\]
Here, \(\Delta^{a}_{ij}\) is the target-reference displacement expressed from the perspective of anchor \(a\). Its signed coordinates provide score functions for directional relations such as \textsc{left}, \textsc{front}, and \textsc{above}. Some predicates also depend on orientation. For these predicates, we express the orientations of the target and reference entities in the same anchor frame:
\[
\mathbf{R}^{a}_i=\mathbf{R}_a^\top\mathbf{R}_i,
\qquad
\mathbf{R}^{a}_j=\mathbf{R}_a^\top\mathbf{R}_j.
\]
Thus, each relation can use both relative position and relative orientation
when computing the score.

Each relation \(r\) has a scoring function \(h_r\) that maps these local
geometric quantities to a scalar value. The evidence is then converted into
a predicate score:
\[
S^a_r[i,j]
=
\sigma\!\left(
\frac{h_r(\Delta^a_{ij}, \mathbf{R}^{a}_i, \mathbf{R}^{a}_j)-m_r}{\tau_r}
\right),
\]
where \(S^a_r[i,j]\in[0,1]\) measures how strongly entity \(e_i\) stands in
relation \(r\) to entity \(e_j\) from anchor \(a\)'s perspective. The margin
\(m_r\) controls how much evidence is needed for relation \(r\), and the
temperature \(\tau_r\) controls how sharply the score changes near that
margin. This formulation supports directional predicates such as \textsc{left}, \textsc{front}, and \textsc{above}; directional combinations such as \textsc{front-left}; and orientation-based predicates such as \textsc{perpendicular} and fixed-angle relations. Detailed predicate definitions are provided in Appendix~\ref{app:spatial_predicates}.

\begin{figure*}[t]
    \centering
    \includegraphics[width=\linewidth]{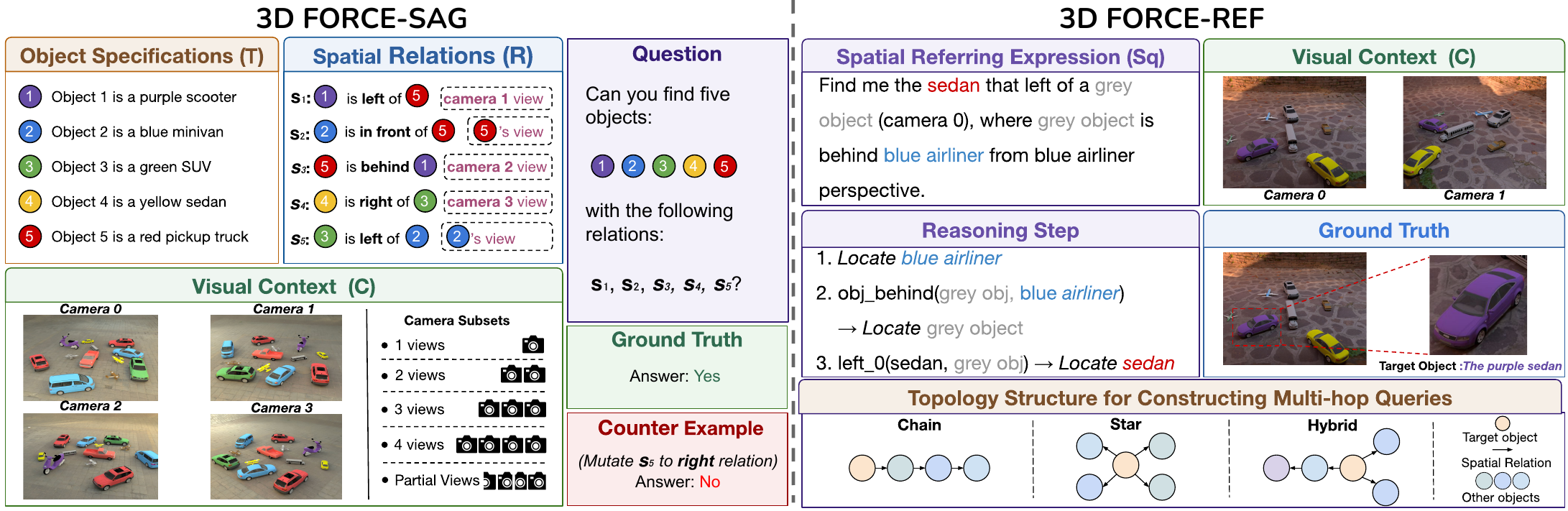}
    \caption{Overview of 3D FORCE benchmark containing two subsets: \textsc{\puzzlebenchshort}, evaluates multi-hop spatial arrangement grounding, and \textsc{\refbenchshort}, evaluates perspective-aware compositional referring expressions.}
    \label{fig:benchmark_overview}
\end{figure*}
\vspace{-2mm}
\subsection{Program Generation and Soft Execution}
\label{sec:code_generation}

Given the query \(Q\), this module produces an executable Python program that specifies how the query should be answered.
We use an LLM with few-shot examples and API documentation to generate the program.
The generated code does not perform low-level geometry.
Instead, it calls three interfaces exposed by \method: semantic scoring, open-ended querying, and spatial predicates from the spatial engine.

For semantic predicates, \method follows NePTune and provides a \texttt{score} interface. Given a natural-language predicate and the entity set \(\mathcal{E}\), \texttt{score} queries the backbone VLM over candidate objects and returns scores aligned with \(\mathcal{E}\). For unary predicates (\(n=1\)), it returns a score vector; for binary predicates (\(n=2\)), it returns a pairwise score matrix aligned with \(\mathcal{E}\times\mathcal{E}\). Entries involving invalid entity types, such as cameras for object-only predicates, are set to zero.

For example, \texttt{score("Is the object in the red box a car?", n=1)} scores each candidate object, while a binary predicate can ask whether two objects have the same color or shape. When the program needs an open-ended value rather than a probabilistic score, it calls \texttt{query}, which returns a textual answer such as an object's color or category.

The generated program is executed directly in Python. During execution, it binds variables to candidate entities, retrieves semantic evidence through \texttt{score} or \texttt{query}, retrieves FoR-aware spatial predicates from the spatial engine, and composes these values to produce the final answer. This explicit execution separates reasoning from direct VLM answer prediction: the VLM provides the atomic semantic concepts, while the program specifies how predicates should be combined. \method preserves uncertainty by keeping semantic and spatial predicate values as continuous scores in \([0,1]\), rather than thresholding them into hard predicates. The executor composes these scores with soft logical operators~\citep{zadeh1965fuzzy}, such as \(A \land B=\min(A,B)\). This allows multi-hop programs to propagate uncertainty from object grounding, 3D estimation, and spatial predicates. Additional composition functions are provided in Appendix~\ref{app:composition_functions}.

\section{Benchmark}

To isolate perspective-aware spatial composition, we introduce 3D FORCE, a diagnostic benchmark that keeps perception and object grounding simple while controlling reasoning depth, relation topology, view count, and FoR assignment. It combines high-resolution scenes with controlled multi-hop relation structures over objects, views, and FoRs.
Each instance is paired with a formal logical representation following prior work~\citep{hsu2024s}, making relation depth and composition structure explicit. Appendix~\ref{app:benchmark_comparison} provides detailed comparison with existing benchmarks.
\bench contains two complementary subsets: \refbenchshort for perspective-aware compositional referring expression grounding and \puzzlebenchshort for perspective-aware spatial arrangement grounding.  Figure~\ref{fig:benchmark_overview} illustrates examples of both subsets.
Answers are derived deterministically from the scene graphs used to render each scene. The full image-rendering setup, instance-generation and verification procedures, and dataset statistics (hop count, perspective settings, and view count) are provided in Appendix~\ref{app:benchmark_construction}.

\subsection{\puzzlebench}\label{sec:puzzle_bench}

The \puzzlebenchshort subset evaluates VLM spatial reasoning by introducing a visual-spatial arrangement grounding task.
Specifically, given a spatial configuration $S= \langle T, R\rangle$, where $T$ includes object specifications (e.g., \textit{object 1 is a blue minivan}) and $R$ describes a set of spatial relations between pairs of objects (e.g., \textit{object 1 is left of object 2}) {in natural language} text, the task is to determine whether the spatial configuration $S$ is satisfied given the visual context $C$. The output is a binary answer $A \in \{\text{Yes}, \text{No}\}$.

\subsection{\refbench}\label{sec:ref_bench}

The \refbenchshort subset evaluates spatial reasoning ability in terms of compositionality and multi-hop reasoning using a referring-expression grounding task.
Specifically, given a spatial referring expression $S_q$ and a visual context(s) $C$, the task is to determine a bounding box $B$ for the target object in $C$. The referring expression consists of a sequence of spatial relations that require multi-hop reasoning.

\section{Experiments}

We organize our experiments around five research questions.
\textbf{RQ1}: How well do current VLMs and tool-augmented methods perform on composing spatial relations with various
FoRs? (\S\ref{sec:e2e-exp})
\textbf{RQ2}: How does performance vary across camera-centric, object-centric, and mixed-FoR settings? (\S\ref{sec:frame-exp})
\textbf{RQ3}: How does performance vary with relation depth and topology? (\S\ref{sec:complexity-exp})
\textbf{RQ4}: What is the effect of soft symbolic reasoning on performance? (\S\ref{sec:uncertainty-exp})
\textbf{RQ5}: Do improvements on our synthetic benchmark transfer to real-world benchmarks? (\S\ref{sec:realworld-exp})

\subsection{Experimental Setting}
\label{sec:experimental_setting}

 We use \bench for RQ1--RQ4 because it provides controlled annotations for task type, FoR, reasoning depth, and relation topology while keeping perception and object grounding simple. These controlled axes let us isolate the effects of perspective and compositional complexity. For RQ5, we evaluate \method on the real-world multi-view benchmarks MindCube-1K~\citep{yin2025spatial} and MMSI~\citep{yang2025mmsi} to test whether the findings extend beyond synthetic scenes. Appendix~\ref{app:experimental_setting} describes these benchmarks and provides implementation details.

\noindent\textbf{Baselines.}
We compare \method against general-purpose VLMs, spatially trained VLMs, and tool-augmented spatial reasoning systems.
The general-purpose VLMs include Qwen3-VL~\citep{qwen2025qwen3vl}, Qwen3.5~\citep{qwen3.5}, InternVL3.5~\citep{internvl25}, Molmo~\citep{clark2026Molmo2openweightsdata}, Gemma 4~\citep{gemmateam2026gemma4}, GLM-4.5V~\citep{vteam2026glm45vglm41vthinkingversatilemultimodal}, GPT~\citep{singh2026openaigpt5card}, and Gemini~\citep{gemini25,gemini31}.
The spatially trained VLMs include SpaceQwen2.5-VL~\citep{remyxai2025spaceqwen25vl3b}, SpaceOm~\citep{remyxai2025spaceom}, SpaceThinker~\citep{remyxai2025spacethinkerqwen25vl3b}, and Cosmos-Reason~\citep{nvidia2025cosmosreason1physicalcommonsense}.
The tool-augmented group includes GCA, pySpatial, and TIGeR.
Unless otherwise stated, we use Qwen3-VL-8B-Instruct as the VLM backbone for parsing and semantic scoring and DeepSeek-V4-Flash~\citep{deepseekai2026deepseekv4highlyefficientmilliontoken} for program generation. Grounding, reconstruction, and orientation use SAM3~\citep{sam3}, VGGT~\citep{wang2025vggt}, and Orient Anything V2 (OAv2)~\citep{wang2026orientv2unifyingorientation}. GCA and pySpatial run on the same VLM, code generator, and perception modules, which controls for differences in the principal neural components and leaves the reasoning procedure as the main distinction. For TIGeR, we use its own fine-tuned weights. Appendix~\ref{app:symbolic_baselines} details each tool-augmented method's configuration. All \method numbers are the mean and standard deviation over three runs.

\noindent\textbf{Evaluation Metrics.}
For \puzzlebenchshort, MMSI, and MindCube, we report exact-match MCQ accuracy.
For \refbenchshort, we evaluate grounding using IoU between the predicted and ground-truth bounding boxes, counting a prediction as correct when IoU \(> 0.5\). For \refbenchshort baselines, all models are prompted to output a bounding box in the same format; malformed or missing boxes are counted as incorrect.

\subsection{Perspective-Aware Reasoning}
\label{sec:e2e-exp}
We evaluate how well current models compose spatial relations across frames of reference (RQ1). The three strongest VLMs cluster at about \(57.7\%\) overall (Table~\ref{tab:mindcube_merged}): Qwen3.5-9B, Qwen3.5-35B, and Gemini-3.1-Pro fall within \(0.12\) percentage points of one another, spanning a small open model, a mid-sized one, and a frontier proprietary system. Within Qwen3-VL, 8B, 32B, and 235B reach \(24.31\%\), \(28.81\%\), and \(26.44\%\), whereas Qwen3.5-9B reaches \(57.75\%\). Similarly, the evaluated spatially trained VLMs are the weakest group and fall below random choice on \textsc{\puzzlebenchshort}, suggesting that neither model scale nor spatial supervision alone is sufficient for FoR composition.

Tool augmentation does not close the gap either: despite using explicit 3D geometry, the strongest of these systems, pySpatial, reaches only \(44.84\%\). By comparison, \method reaches \(82.89\%\), exceeding pySpatial by \(38.1\) percentage points and the strongest VLM by \(25.1\) percentage points.

\begin{table}[t]
    \centering
    \scriptsize
    \setlength{\tabcolsep}{2.5pt}
    \adjustbox{width=\columnwidth}{
    \begin{tabular}{lcccc}
    \toprule
    \textbf{Method} & \textbf{Micro Avg.} & \textbf{SAG} & \textbf{REF} & \textbf{BBox} \\
    \midrule
    Random choice        & 18.80 & 50.17 & 1.53 & -- \\
    Qwen3-VL-8B          & 24.31 & 65.91 & 1.39 & 99.28 \\
    Qwen3-VL-32B    & 28.81 & 72.26 & 4.89 & 99.81 \\
    Qwen3-VL-235B        & 26.44 & 73.22 & 0.67 & 99.81 \\
    Qwen3.5-9B           & \textbf{57.75} & 74.87 & \textbf{48.32} & 98.75 \\
    Qwen3.5-35B          & \underline{57.63} & \textbf{76.35} & 47.32 & 99.76 \\
    InternVL3.5-8B       & 24.46 & 53.30 & 8.57 & 99.04 \\
    InternVL3.5-14B & 29.09 & 52.35 & 16.28 & 99.62 \\
    InternVL3.5-38B      & 30.82 & 58.26 & 15.71 & 99.81 \\
    Molmo2-4B       & 21.09 & 57.65 & 0.96 & 88.07 \\
    Molmo2-8B            & 19.89 & 54.61 & 0.77 & 98.61 \\
    Gemma4-2B       & 20.69 & 52.78 & 3.02 & 98.23 \\
    Gemma4-4B       & 21.62 & 55.74 & 2.83 & 98.95 \\
    Gemma4-31B      & 26.25 & 64.35 & 5.27 & 99.66 \\
    GPT-5.1              & 29.80 & 71.39 & 6.90 & 99.43 \\
    Gemini-3.1-Pro       & 57.63 & \underline{75.30} & \underline{47.89} & 91.19 \\
    \midrule
    SpaceQwen-3B         & 13.96 & 37.74 & 0.86 & 72.22 \\
    SpaceOm-3B           & 17.79 & 49.39 & 0.38 & 95.35 \\
    SpaceThinker-3B      & 17.51 & 48.00 & 0.72 & 97.13 \\
    Cosmos-Reason1-7B    & 13.47 & 37.04 & 0.48 & 78.74 \\
    \midrule
    GCA                  & 24.24 & \underline{51.00} & 9.50 & -- \\
    pySpatial       & \underline{44.84} & 48.35 & \underline{42.91} & -- \\
    TIGeR           & 20.38 & 30.43 & 14.85 & -- \\
    \rowcolor{orange!12}
    \method{} (ours)     & \textbf{82.89{\tiny{$\pm$}0.40}} & \textbf{85.88{\tiny{$\pm$}0.13}} & \textbf{81.24{\tiny{$\pm$}0.68}} & -- \\
    \method{}$_{\text{GT-BBox}}$ & 88.85{\tiny{$\pm$}0.55} & 87.57{\tiny{$\pm$}0.97} & 89.56{\tiny{$\pm$}0.67} & -- \\
    \method{}$_{\text{Oracle 3D}}$
                         & 95.94{\tiny{$\pm$}0.35} & 94.40{\tiny{$\pm$}0.68} & 96.79{\tiny{$\pm$}0.39} & -- \\
    \bottomrule
\end{tabular}
    }
    \caption{Overall results on \bench. BBox is the parsing success rate of output boxes. Oracle 3D uses ground-truth geometry. \method{} rows report the mean and standard deviation over three runs.}
    \label{tab:mindcube_merged}
\end{table}

\subsection{Frame of Reference Analysis}
\label{sec:frame-exp}
\begin{figure}[h]
    \centering
    \includegraphics[width=0.95\linewidth]{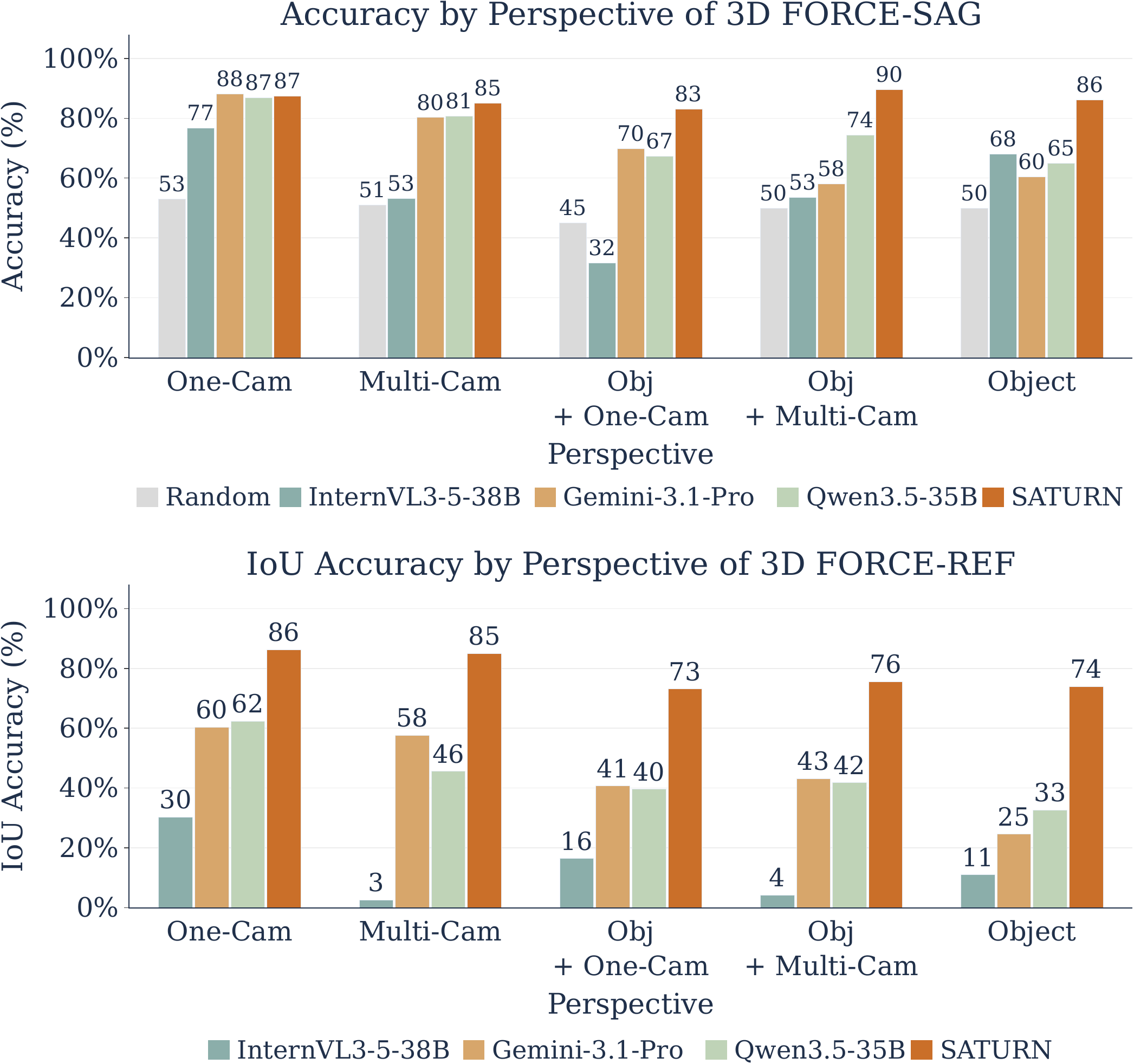}
    \caption{Accuracy on \puzzlebenchshort (top) and \refbenchshort (bottom) across perspective settings for top-performing models.}
    \label{fig:perspective_analysis}
\end{figure}

 The aggregate results in Table~\ref{tab:mindcube_merged} do not show how performance changes across FoRs. To answer RQ2, we group \puzzlebenchshort and \refbenchshort by the perspective required for reasoning and report the strongest baselines in Figure~\ref{fig:perspective_analysis}.
 Every VLM shown performs best in the single-camera setting and declines when reasoning involves an object frame or a second camera frame. Across the five settings, each VLM varies by at least \(22\) percentage points between its best and worst result, while \method varies by about \(7\) percentage points on \puzzlebenchshort and \(13\) percentage points on \refbenchshort.
 For example, Qwen3.5-35B falls from \(62\%\) in the single-camera setting to \(33\%\) in the object-centric setting, while \method falls from \(86\%\) to \(74\%\).

 The breakdown also distinguishes models with nearly identical aggregate performance. Qwen3.5-35B and Gemini-3.1-Pro both achieve about \(57.6\%\) overall, yet Gemini is \(8\) percentage points weaker on object-only \refbenchshort{} and \(12\) percentage points stronger on multi-camera \refbenchshort. Similar overall performance can therefore mask different sensitivities to perspective.

\subsection{Reasoning Complexity Analysis}
\label{sec:complexity-exp}
\begin{figure}[h]
    \centering
    \includegraphics[width=0.85\linewidth]{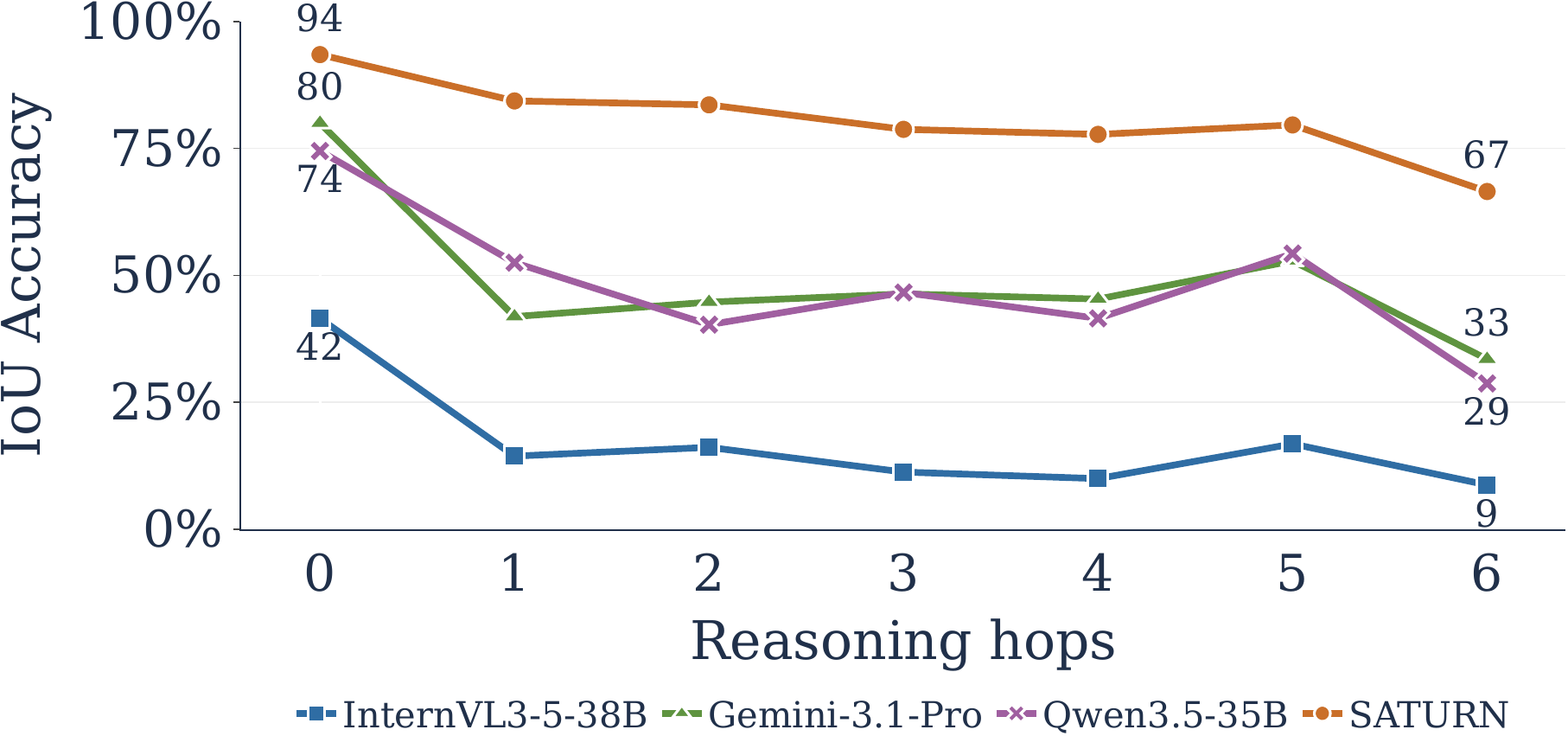}
    \caption{Accuracy on \refbenchshort{} by number of reasoning hops, pooled over topologies.}
    \label{fig:hop-analysis}
\end{figure}

In addition to FoR, compositional depth and relation topology also affect task difficulty. To answer RQ3, we group \refbenchshort by the number of reasoning hops. Figure~\ref{fig:hop-analysis} reports performance by hop count, pooled over topologies, while Appendix~\ref{app:topo} separates the results by topology. Every model performs substantially worse at six hops than at zero hops: \method falls from \(94\%\) to \(67\%\), Qwen3.5-35B from \(74\%\) to \(29\%\), Gemini-3.1-Pro from \(80\%\) to \(33\%\), and InternVL3.5-38B from \(42\%\) to \(9\%\). \method shows the smallest decline and outperforms every baseline at each hop count.

\begin{table*}[t]
\centering
\small
\setlength{\tabcolsep}{3.5pt}
\renewcommand{\arraystretch}{1.12}
\adjustbox{width={0.95\textwidth}}{
\begin{tabular}{lccccccccc}
\toprule
\multirow{2}{*}{\textbf{Method}} &
\multicolumn{4}{c}{\textbf{MindCube}} &
\multicolumn{5}{c}{\textbf{MMSI}} \\
\cmidrule(lr){2-5} \cmidrule(lr){6-10}
& \textbf{Overall}
& \textbf{Rotation}
& \textbf{Among}
& \textbf{Around}
& \textbf{Overall}
& \textbf{Positional}
& \textbf{Motion}
& \textbf{Attribute}
& \textbf{MSR} \\
\midrule

Qwen3-VL-8B-Instruct
& 38.86 & 43.50 & 32.83 & 49.60
& 29.40 & 32.18 & 27.33 & 31.54 & 22.22 \\

Qwen3-VL-235B-Thinking
& 47.30 & \underline{87.00} & 35.00 & 47.30
& 32.60 & 33.70 & 23.30 & 40.00 & 31.80 \\

GLM-4.5V
& 39.60 & 60.00 & 42.20 & 25.50
& 33.80 & 35.60 & 29.30 & 36.90 & 30.30 \\

Gemini-2.5-Pro
& 57.50 & \textbf{89.50} & 48.80 & 54.50
& 36.90 & 39.00 & 33.30 & 36.20 & \underline{34.30} \\

\midrule

SpaceQwen2.5-VL-3B
& 34.95 & 25.50 & 40.20 & 30.00
& 27.10 & 28.93 & 20.00 & 30.00 & 25.76 \\

SpaceOm
& 41.90 & 30.50 & 41.50 & 52.00
& 26.60 & 28.35 & 20.67 & 26.15 & 26.77 \\

SpaceThinker-3B
& 35.14 & 30.00 & 40.00 & 27.60
& 27.60 & 27.78 & 24.67 & 30.77 & 27.27 \\

Cosmos-Reason1-7B
& 36.67 & 39.50 & 34.20 & 40.40
& 26.90 & 28.54 & 22.67 & 28.46 & 24.75 \\

\midrule

TIGeR
& 26.95 & 18.50 & 27.50 & 32.40
& 25.50 & 28.35 & 22.00 & 23.85 & 21.72 \\

GCA\textsubscript{235B-Think}
& \underline{64.20} & 82.00 & 59.80 & 61.80
& 41.90 & \underline{46.74} & \underline{44.00} & 36.92 & 30.81 \\

GCA\textsubscript{8B}
& 32.19 & 34.00 & 34.00 & 26.40
& 37.69 & 40.88 & 42.18 & 30.23 & 30.81 \\

pySpatial
& 53.14 & 38.50 & 50.30 & \underline{71.60}
& 28.20 & 30.27 & 35.33 & 17.69 & 24.24\\

Qwen3VL + scene state
& 60.86 & 49.50 & \underline{60.50} & 70.80
& 39.40 & 41.38 & 42.67 & \underline{42.31} & 29.80 \\

\rowcolor{orange!12}
  \method\ (ours)
  & \textbf{78.06}{\tiny{$\pm$}1.05} & {85.67\tiny{$\pm$}2.93} & \textbf{77.00}{\tiny{$\pm$}1.02} & \textbf{74.53}{\tiny{$\pm$}2.31}
  & \textbf{48.77}{\tiny{$\pm$}0.61} & \textbf{53.19}{\tiny{$\pm$}0.73} & \textbf{50.44}{\tiny{$\pm$}1.39} & \textbf{45.38}{\tiny{$\pm$}2.04} &
  \textbf{38.05}{\tiny{$\pm$}0.29} \\
\bottomrule
\end{tabular}
}
\caption{Results on MindCube and MMSI. Refer to Appendix~\ref{app:significance} for paired confidence intervals.}
\label{tab:mindcube_mmsi}
\end{table*}

The pooled results are not monotonic because the available topologies change with hop count. Five-hop queries use only the \textit{chain} topology, while six-hop queries use only \textit{hybrid}; the increase at five hops therefore reflects a shift to an easier topology rather than improved performance at greater depth. The topology-specific results in Appendix~\ref{app:topo} support this interpretation: performance remains relatively stable across chain depth, while the VLMs decline as star depth increases. We also vary the number of images and camera frames composed by each query to test other dimensions of input complexity. Neither produces a systematic trend for \method, whose performance remains between \(76\%\) and \(93\%\) as shown in Appendix~\ref{app:nk}).

\subsection{Soft Symbolic Reasoning Effect}
\label{sec:uncertainty-exp}
\begin{figure}[h]
\centering
\includegraphics[width=0.8\linewidth]{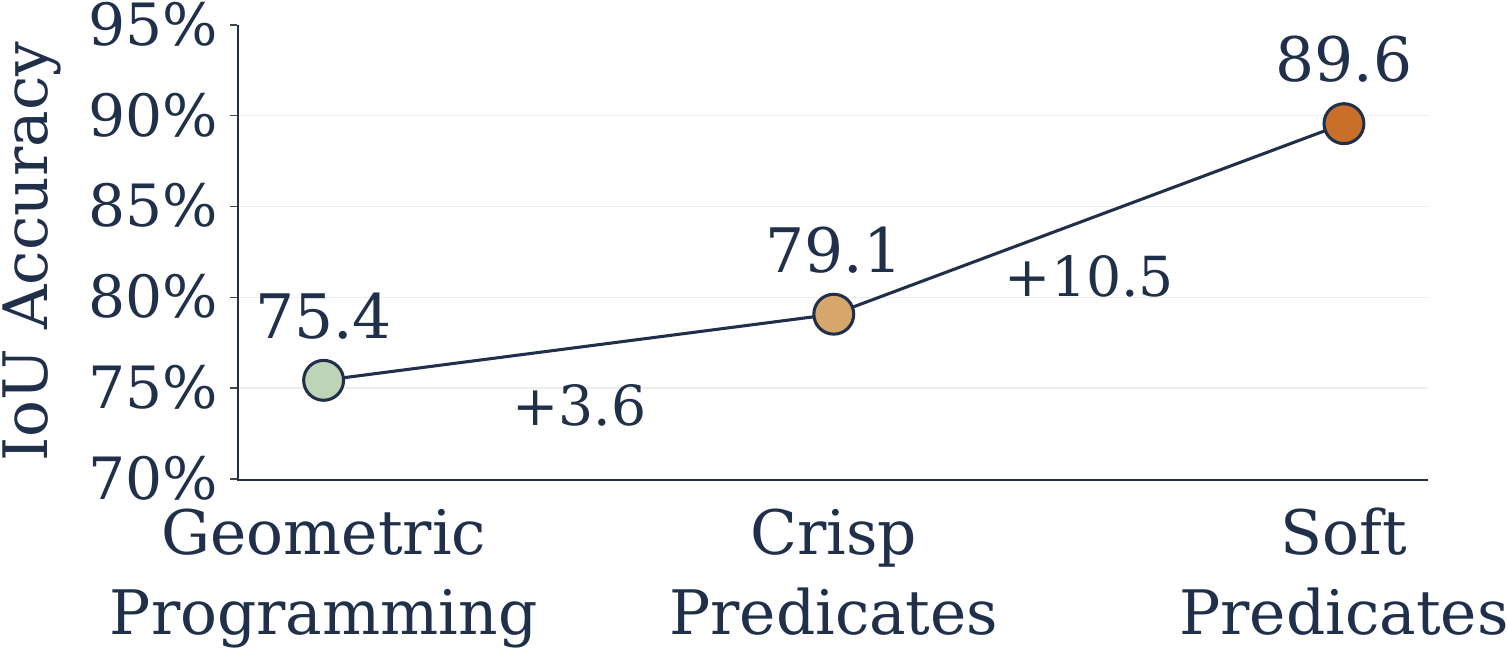}
\caption{Ablation of symbolic reasoning variants.}
\label{fig:symbolic_reasoning_ablation}
\end{figure}

To isolate the contributions of the declarative predicate interface and continuous predicate scores (RQ4), we compare three variants that use the same scenes built from the benchmark's ground-truth bounding boxes. \textit{Geometric Programming} has no declarative API and asks the backbone LLM to write low-level Python code. \textit{Crisp Predicates} provides the declarative interface but with binary 0 or 1 predicate scores. \textit{Soft Predicates} uses the same symbolic reasoning predicates but with continuous scores. The last two variants therefore differ only in thresholding.
Geometric Programming reaches \(75.43\%\) on \refbenchshort (Figure~\ref{fig:symbolic_reasoning_ablation}). Adding the declarative predicate interface raises performance to \(79.07\%\), a gain of \(3.6\) pp, while retaining continuous scores further raises it to \(89.6\%\), a gain of \(10.5\) pp. Both components contribute to performance, with the larger gain coming from preserving uncertainty.

\subsection{Real-World Multi-View Reasoning}
\label{sec:realworld-exp}

 To test whether the gains on \bench extend to real images (RQ5), we evaluate on the multi-view benchmarks MindCube and MMSI. As shown in Table~\ref{tab:mindcube_mmsi}, \method achieves the best overall performance on both benchmarks, reaching \(78.06\%\) on MindCube and \(48.77\%\) on MMSI. On MindCube, \method outperforms GCA by \(13.9\) percentage points in its released configuration and by \(45.9\) percentage points when GCA uses the same backbone as \method. \method also achieves the best performance on the \textit{Among} and \textit{Around} categories. On \textit{Rotation}, however, Gemini-2.5-Pro and Qwen3-VL-235B remain stronger. \method incorporates textual rotation cues through its pose-constraint module, but this category remains sensitive to errors in constraint extraction and view alignment.

On MMSI, \method outperforms all baselines overall and achieves the best performance in the positional, motion, attribute, and multi-step reasoning categories. The scene-state control helps separate the contributions of scene estimation and symbolic composition. Providing \method's estimated scene as text to Qwen3-VL-8B raises performance from \(38.86\%\) to \(60.86\%\) on MindCube and from \(29.40\%\) to \(39.40\%\) on MMSI. Composing predicates over the same scene further increases performance by \(17.2\) and \(9.4\) percentage points, respectively. Thus, scene estimation and symbolic composition each account for roughly half of the improvement over direct VLM inference. These results show that the gains on \bench extend to real images, while performance on rotation-heavy queries remains sensitive to view alignment.

\begin{table}[t]
  \centering
  \small
  \setlength{\tabcolsep}{3pt}
  \renewcommand{\arraystretch}{1.05}

  \begin{tabular}{lcccc}
  \toprule
  \textbf{Model} & \textbf{Overall} & \textbf{Rotation} & \textbf{Around} & \textbf{Among} \\
  \midrule
  VGGT
  & 79.14 {\tiny \textcolor{green}{(+1.71)}}
  & 84.50 {\tiny \textcolor{green}{(+6.00)}}
  & 77.20 {\tiny \textcolor{green}{(+2.40)}}
  & 78.17 {\tiny \textcolor{green}{(+0.00)}} \\
  DA3
  & 75.05 {\tiny \textcolor{green}{(+0.76)}}
  & 81.50 {\tiny \textcolor{green}{(+3.50)}}
  & 70.40 {\tiny \textcolor{green}{(+1.20)}}
  & 74.83 {\tiny \textcolor{red}{(-0.33)}} \\
  CUT3R
  & 76.25 {\tiny \textcolor{green}{(+2.53)}}
  & 85.00 {\tiny \textcolor{green}{(+9.00)}}
  & 72.00 {\tiny \textcolor{red}{(-0.40)}}
  & 75.10 {\tiny \textcolor{green}{(+1.60)}} \\
  \bottomrule
  \end{tabular}
  \caption{Per-backbone single runs on MindCube. Parentheses show the gain from pose constraints.}
  \label{tab:pose_constraint_ablation}
  \end{table}

\section{Analysis}
\label{sec:analysis}

\subsection{3D Backbone and Pose Constraints}
\label{sec:ablation-exp}

To test whether \method's performance depends on the 3D reconstruction backbone, we evaluate it with VGGT, DA3~\citep{depthanything3}, and CUT3R~\citep{cut3r} on MindCube. As shown in Table~\ref{tab:pose_constraint_ablation}, overall accuracy varies by at most \(4.1\) percentage points, so its performance is not tied to a single reconstruction backbone. Pose constraints also improve overall performance for every backbone, with the largest gains on Rotation, where accurate view alignment is crucial.

\subsection{Error Sources}
\label{sec:error-sources}
To identify the main sources of error in \method's end-to-end pipeline, we manually trace errors from one MindCube run to their earliest identifiable source. As shown in Table~\ref{tab:failure_attribution}, nearly half occur while connecting the question to the images: the query parser selects the relevant objects and views, while the grounder locates those objects; together, these steps account for \(46.6\%\) of failures. Another \(23.7\%\) arise from 3D reconstruction estimating object and camera positions and orientations, while VLM-based semantic scoring contributes \(18.6\%\). Only \(6.8\%\) come from program generation, indicating that the main bottleneck is supplying the program with the correct objects and geometry. We provide qualitative examples in Appendix~\ref{app:qualitative}.

\begin{table}[t]
\centering
\scriptsize
\setlength{\tabcolsep}{4pt}
\adjustbox{width=0.8\columnwidth}{
\begin{tabular}{lr}
\toprule
\textbf{Earliest identified source} & \textbf{Share} \\
\midrule
Visual grounding / detection           & 33.9\% \\
Reconstruction / camera pose    & 22.9\% \\
Semantic scoring                & 18.6\% \\
Query parsing / object selection                    & 12.7\% \\
Program generation              & 6.8\% \\
Ambiguous reference             & 4.2\% \\
Object orientation              & 0.8\% \\
\bottomrule
\end{tabular}}
\caption{Earliest identified source among manually categorized MindCube failures.}
\label{tab:failure_attribution}
\end{table}

\subsection{Cost}
\label{sec:cost-exp}
Table~\ref{tab:latency} reports end-to-end per-query latency on MindCube. \method is slower than pySpatial, direct inference, and TIGeR, but faster than GCA. Most of \method's cost is due to visual processing and can be amortized across queries related to the same scene; Appendix~\ref{app:cost} breaks the total down by stage. However, additional computation from the visual processing alone is not the only reason for the \method's performance gain. The VLM + scene state baseline using chain of thought prompting is slower end-to-end yet \(17.2\) percentage points less accurate than \method. Moreover, all methods in \S\ref{sec:uncertainty-exp} share the same perception tools, yet their accuracy ranges from \(75.4\%\) to \(89.6\%\).

\begin{table}[h]
\centering
\small
\setlength{\tabcolsep}{5pt}
\renewcommand{\arraystretch}{1.05}
\begin{tabular}{lrr}
\toprule
\textbf{Method} & \textbf{Median (s)} & \textbf{Mean (s)} \\
\midrule
pySpatial            & 7  & 7 \\
Qwen3-VL-8B (direct) & 12 & 15 \\
TIGeR                & 25 & 28 \\
\method              & 91 & 105 \\
VLM + scene state    & 114 & 174 \\
GCA\textsubscript{8B} & 123 & 205 \\
\bottomrule
\end{tabular}
\caption{Per-query end-to-end latency on MindCube.}
\label{tab:latency}
\end{table}

\section{Conclusion}
We introduced \method, a neuro-symbolic framework for FoR-aware spatial composition, and \benchfull, a diagnostic benchmark for this capability. \bench shows VLMs remain unreliable when composing relations across objects, views, and FoRs. \method addresses this limitation by composing soft camera- and object-centric predicates over an estimated 3D scene. It outperforms general-purpose and spatially trained VLMs and tool-augmented systems on \bench and performs best overall on the real-world multi-view benchmarks MindCube and MMSI. Together, these results identify FoR-aware composition as a distinct spatial reasoning challenge and show the value of explicit spatial predicates with soft truth values.

\section*{Limitations}
Although \method achieves robust performance, it remains a neuro-symbolic framework whose effectiveness is bounded by the capabilities of the underlying neural backbone and the reliability of external tools, which may introduce noisy intermediate predictions. SATURN depends on accurate camera and object orientation estimates; errors in view alignment can propagate directly into derived predicates.
\bench is a diagnostic benchmark for compositional grounding across multiple frames of reference. It is rendered from a controlled simulator, so results on it measure how well a system composes relations under multiple frames of reference. It does not establish generality to open embodied environments. In general, constructing benchmarks that fully capture the intricacies of real-world environments remains a fundamental limitation of this line of work and embodied AI applications.
Furthermore, our framework assumes access to substantial GPU resources to support VLM backbones of varying sizes. These computational demands may limit accessibility for researchers with constrained resources.

\section*{Acknowledgements}
This project is partially supported by the Office of Naval Research (ONR) grant N00014-23-1-2417. Any opinions, findings, and conclusions or recommendations expressed in this material are those of the authors and do not necessarily reflect the views of the Office of Naval Research.

\bibliography{references}

@INPROCEEDINGS{Gupta2022VisProg,
  author={Gupta, Tanmay and Kembhavi, Aniruddha},
  booktitle={2023 IEEE/CVF Conference on Computer Vision and Pattern Recognition (CVPR)}, 
  title={Visual Programming: Compositional visual reasoning without training}, 
  year={2023},
  volume={},
  number={},
  pages={14953-14962},
  doi={10.1109/CVPR52729.2023.01436}}

@article{hsu2024s,
  title={What’s left? concept grounding with logic-enhanced foundation models},
  author={Hsu, Joy and Mao, Jiayuan and Tenenbaum, Josh and Wu, Jiajun},
  journal={Advances in Neural Information Processing Systems},
  volume={36},
  year={2024}
}

@misc{suris2023vipergpt,
title={ViperGPT: Visual Inference via Python Execution for Reasoning},
    author={Sur\'is, D\'idac and Menon, Sachit and Vondrick, Carl},
    journal={Proceedings of IEEE International Conference on Computer Vision (ICCV)},
    year={2023}
}

@inproceedings{kamali-kordjamshidi-2023-syntax,
    title = "Syntax-Guided Transformers: Elevating Compositional Generalization and Grounding in Multimodal Environments",
    author = "Kamali, Danial and 
      Kordjamshidi, Parisa",
    editor = "Hupkes, Dieuwke  and
      Dankers, Verna  and
      Batsuren, Khuyagbaatar  and
      Sinha, Koustuv  and
      Kazemnejad, Amirhossein  and
      Christodoulopoulos, Christos  and
      Cotterell, Ryan  and
      Bruni, Elia",
    booktitle = "Proceedings of the 1st GenBench Workshop on (Benchmarking) Generalisation in NLP",
    month = dec,
    year = "2023",
    address = "Singapore",
    publisher = "Association for Computational Linguistics",
    url = "https://aclanthology.org/2023.genbench-1.10",
    doi = "10.18653/v1/2023.genbench-1.10",
    pages = "130--142",
}

@misc{sphere,
      title={SPHERE: Unveiling Spatial Blind Spots in Vision-Language Models Through Hierarchical Evaluation}, 
      author={Wenyu Zhang and Wei En Ng and Lixin Ma and Yuwen Wang and Junqi Zhao and Allison Koenecke and Boyang Li and Lu Wang},
      year={2025},
      eprint={2412.12693},
      archivePrefix={arXiv},
      primaryClass={cs.CV},
      url={https://arxiv.org/abs/2412.12693}, 
}

@misc{viewspatial,
      title={ViewSpatial-Bench: Evaluating Multi-perspective Spatial Localization in Vision-Language Models}, 
      author={Dingming Li and Hongxing Li and Zixuan Wang and Yuchen Yan and Hang Zhang and Siqi Chen and Guiyang Hou and Shengpei Jiang and Wenqi Zhang and Yongliang Shen and Weiming Lu and Yueting Zhuang},
      year={2025},
      eprint={2505.21500},
      archivePrefix={arXiv},
      primaryClass={cs.CV},
      url={https://arxiv.org/abs/2505.21500}, 
}

@misc{blink,
      title={BLINK: Multimodal Large Language Models Can See but Not Perceive}, 
      author={Xingyu Fu and Yushi Hu and Bangzheng Li and Yu Feng and Haoyu Wang and Xudong Lin and Dan Roth and Noah A. Smith and Wei-Chiu Ma and Ranjay Krishna},
      year={2024},
      eprint={2404.12390},
      archivePrefix={arXiv},
      primaryClass={cs.CV},
      url={https://arxiv.org/abs/2404.12390}, 
}

@inproceedings{johnson2017clevr,
  title={CLEVR: A Diagnostic Dataset for Compositional Language and Elementary Visual Reasoning},
  author={Johnson, Justin and Hariharan, Bharath and van der Maaten, Laurens
          and Fei-Fei, Li and Zitnick, C Lawrence and Girshick, Ross},
  booktitle={CVPR},
  year={2017}
}

@inproceedings{faghihi2021domiknows,
    title = "{DomiKnowS:} A Library for Integration of Symbolic Domain Knowledge in Deep Learning",
    author = "Hossein Rajaby Faghihi and Quan Guo and Andrzej Uszok and Aliakbar Nafar and Elaheh Raisi and Parisa Kordjamshidi",
    booktitle = "Proceedings of the 2021 Conference on Empirical Methods in Natural Language Processing (EMNLP), Demo Track",
    month = nov,
    year = "2021",
    address = "Online",
    publisher = "Association for Computational Linguistics",
}

@article{nesycoco,
title={NeSyCoCo: A Neuro-Symbolic Concept Composer for Compositional Generalization},
volume={39},
url={https://ojs.aaai.org/index.php/AAAI/article/view/32439},
DOI={10.1609/aaai.v39i4.32439},
number={4},
journal={Proceedings of the AAAI Conference on Artificial Intelligence},
author={Kamali, Danial and Barezi, Elham J. and Kordjamshidi, Parisa},
year={2025},
month={Apr.},
pages={4184-4193}
}

@misc{internvl25,
      title={Expanding Performance Boundaries of Open-Source Multimodal Models with Model, Data, and Test-Time Scaling}, 
      author={Zhe Chen and Weiyun Wang and Yue Cao and Yangzhou Liu and Zhangwei Gao and Erfei Cui and Jinguo Zhu and Shenglong Ye and Hao Tian and Zhaoyang Liu and Lixin Gu and Xuehui Wang and Qingyun Li and Yimin Ren and Zixuan Chen and Jiapeng Luo and Jiahao Wang and Tan Jiang and Bo Wang and Conghui He and Botian Shi and Xingcheng Zhang and Han Lv and Yi Wang and Wenqi Shao and Pei Chu and Zhongying Tu and Tong He and Zhiyong Wu and Huipeng Deng and Jiaye Ge and Kai Chen and Kaipeng Zhang and Limin Wang and Min Dou and Lewei Lu and Xizhou Zhu and Tong Lu and Dahua Lin and Yu Qiao and Jifeng Dai and Wenhai Wang},
      year={2025},
      eprint={2412.05271},
      archivePrefix={arXiv},
      primaryClass={cs.CV},
      url={https://arxiv.org/abs/2412.05271}, 
}

@inproceedings{
kamali2026neptune,
title={Ne{PT}une: A Neuro-Pythonic Framework for Tunable Compositional Reasoning on Vision-Language},
author={Danial Kamali and Parisa Kordjamshidi},
booktitle={The Fourteenth International Conference on Learning Representations},
year={2026},
url={https://openreview.net/forum?id=8H0TkSusWI}
}

@article{sinha2026neurosymbolicframeworks,
  title={Neuro-Symbolic Frameworks: Conceptual Characterization and Empirical Comparative Analysis},
  author={Sinha, Sania and Premsri, Tanawan and Kamali, Danial and Kordjamshidi, Parisa},
  journal={Neurosymbolic Artificial Intelligence},
  year={2026},
  doi={10.1177/29498732261443183},
  url={https://doi.org/10.1177/29498732261443183}
}

@misc{nafar2026agentic,
  title={An Agentic Framework for Neuro-Symbolic Programming},
  author={Nafar, Aliakbar and Chigurupati, Chetan and Kamali, Danial and Karimian, Hamid and Kordjamshidi, Parisa},
  year={2026},
  eprint={2601.00743},
  archivePrefix={arXiv},
  primaryClass={cs.AI},
  doi={10.48550/arXiv.2601.00743},
  url={https://arxiv.org/abs/2601.00743}
}

@misc{3dsrbench,
      title={3DSRBench: A Comprehensive 3D Spatial Reasoning Benchmark}, 
      author={Wufei Ma and Haoyu Chen and Guofeng Zhang and Yu-Cheng Chou and Jieneng Chen and Celso M de Melo and Alan Yuille},
      year={2025},
      eprint={2412.07825},
      archivePrefix={arXiv},
      primaryClass={cs.CV},
      url={https://arxiv.org/abs/2412.07825}, 
}

@misc{space-10,
      title={SpaCE-10: A Comprehensive Benchmark for Multimodal Large Language Models in Compositional Spatial Intelligence}, 
      author={Ziyang Gong and Wenhao Li and Oliver Ma and Songyuan Li and Zhaokai Wang and Songyuan Li and Jiayi Ji and Xue Yang and Gen Luo and Junchi Yan and Rongrong Ji},
      year={2025},
      eprint={2506.07966},
      archivePrefix={arXiv},
      primaryClass={cs.CV},
      url={https://arxiv.org/abs/2506.07966}, 
}

@String(CVPR= {IEEE Conf. Comput. Vis. Pattern Recog.})

@String(ICCV= {Int. Conf. Comput. Vis.})

@String(ICLR = {Int. Conf. Learn. Represent.})

@String(AAAI = {AAAI})

@String(CVPR  = {CVPR})

@String(ICCV  = {ICCV})

@String(ICLR  = {ICLR})

@misc{erik25mmspatial,
      title={MM-Spatial: Exploring 3D Spatial Understanding in Multimodal LLMs}, 
      author={Erik Daxberger and Nina Wenzel and David Griffiths and Haiming Gang and Justin Lazarow and Gefen Kohavi and Kai Kang and Marcin Eichner and Yinfei Yang and Afshin Dehghan and Peter Grasch},
      year={2025},
      eprint={2503.13111},
      archivePrefix={arXiv},
      primaryClass={cs.CV},
      url={https://arxiv.org/abs/2503.13111}, 
}

@inproceedings{liumllms25large,
    title = "Can Multimodal Large Language Models Understand Spatial Relations?",
    author = "Liu, Jingping  and
      Liu, Ziyan  and
      Cen, Zhedong  and
      Zhou, Yan  and
      Zou, Yinan  and
      Zhang, Weiyan  and
      Jiang, Haiyun  and
      Ruan, Tong",
    editor = "Che, Wanxiang  and
      Nabende, Joyce  and
      Shutova, Ekaterina  and
      Pilehvar, Mohammad Taher",
    booktitle = "Proceedings of the 63rd Annual Meeting of the Association for Computational Linguistics (Volume 1: Long Papers)",
    month = jul,
    year = "2025",
    address = "Vienna, Austria",
    publisher = "Association for Computational Linguistics",
    url = "https://aclanthology.org/2025.acl-long.31/",
    doi = "10.18653/v1/2025.acl-long.31",
    pages = "620--632",
    ISBN = "979-8-89176-251-0"
}

@inproceedings{
wu2025spatial,
title={Spatial-{MLLM}: Boosting {MLLM} Capabilities in Visual-based Spatial Intelligence},
author={Diankun Wu and Fangfu Liu and Yi-Hsin Hung and Yueqi Duan},
booktitle={Advances in Neural Information Processing Systems},
year={2025},
url={https://openreview.net/forum?id=RnXS7aK4rK}
}

@inproceedings{chen2024spatialvlm,
  title={Spatialvlm: Endowing vision-language models with spatial reasoning capabilities},
  author={Chen, Boyuan and Xu, Zhuo and Kirmani, Sean and Ichter, Brian and Sadigh, Dorsa and Guibas, Leonidas and Xia, Fei},
  booktitle={Proceedings of the IEEE/CVF Conference on Computer Vision and Pattern Recognition},
  pages={14455--14465},
  year={2024}
}

@article{wang2024picture,
  title={Is a picture worth a thousand words? delving into spatial reasoning for vision language models},
  author={Wang, Jiayu and Ming, Yifei and Shi, Zhenmei and Vineet, Vibhav and Wang, Xin and Li, Sharon and Joshi, Neel},
  journal={Advances in Neural Information Processing Systems},
  volume={37},
  pages={75392--75421},
  year={2024}
}

@inproceedings{kamath2023s,
  title={What’s “up” with vision-language models? Investigating their struggle with spatial reasoning},
  author={Kamath, Amita and Hessel, Jack and Chang, Kai-Wei},
  booktitle={Proceedings of the Conference on Empirical Methods in Natural Language Processing},
  pages={9161--9175},
  year={2023}
}

@inproceedings{wang2025vggt,
  title={Vggt: Visual geometry grounded transformer},
  author={Wang, Jianyuan and Chen, Minghao and Karaev, Nikita and Vedaldi, Andrea and Rupprecht, Christian and Novotny, David},
  booktitle={Proceedings of the Computer Vision and Pattern Recognition Conference},
  pages={5294--5306},
  year={2025}
}

@inproceedings{
ma2025spatialreasoner,
title={SpatialReasoner: Towards Explicit and Generalizable 3D Spatial Reasoning},
author={Wufei Ma and Yu-Cheng Chou and Qihao Liu and Xingrui Wang and Celso M de Melo and Jianwen Xie and Alan Yuille},
booktitle={Advances in Neural Information Processing Systems},
year={2025},
url={https://openreview.net/forum?id=hFaXVjRFHI}
}

@article{fan2025vlm,
  title={VLM-3R: Vision-Language Models Augmented with Instruction-Aligned 3D Reconstruction},
  author={Fan, Zhiwen and Zhang, Jian and Li, Renjie and Zhang, Junge and Chen, Runjin and Hu, Hezhen and Wang, Kevin and Qu, Huaizhi and Wang, Dilin and Yan, Zhicheng and others},
  journal={arXiv preprint arXiv:2505.20279},
  year={2025}
}

@article{cheng2024spatialrgpt,
  title={Spatialrgpt: Grounded spatial reasoning in vision-language models},
  author={Cheng, An-Chieh and Yin, Hongxu and Fu, Yang and Guo, Qiushan and Yang, Ruihan and Kautz, Jan and Wang, Xiaolong and Liu, Sifei},
  journal={Advances in Neural Information Processing Systems},
  volume={37},
  pages={135062--135093},
  year={2024}
}

@misc{mvrobobench,
      title={Seeing Across Views: Benchmarking Spatial Reasoning of Vision-Language Models in Robotic Scenes}, 
      author={Zhiyuan Feng and Zhaolu Kang and Qijie Wang and Zhiying Du and Jiongrui Yan and Shubin Shi and Chengbo Yuan and Huizhi Liang and Yu Deng and Qixiu Li and Rushuai Yang and Arctanx An and Leqi Zheng and Weijie Wang and Shawn Chen and Sicheng Xu and Yaobo Liang and Jiaolong Yang and Baining Guo},
      year={2026},
      eprint={2510.19400},
      archivePrefix={arXiv},
      primaryClass={cs.CV},
      url={https://arxiv.org/abs/2510.19400}, 
}

@misc{allanglesbench,
      title={Seeing from Another Perspective: Evaluating Multi-View Understanding in MLLMs}, 
      author={Chun-Hsiao Yeh and Chenyu Wang and Shengbang Tong and Ta-Ying Cheng and Ruoyu Wang and Tianzhe Chu and Yuexiang Zhai and Yubei Chen and Shenghua Gao and Yi Ma},
      year={2025},
      eprint={2504.15280},
      archivePrefix={arXiv},
      primaryClass={cs.CV},
      url={https://arxiv.org/abs/2504.15280}, 
}

@misc{sparbench,
      title={From Flatland to Space: Teaching Vision-Language Models to Perceive and Reason in 3D}, 
      author={Jiahui Zhang and Yurui Chen and Yanpeng Zhou and Yueming Xu and Ze Huang and Jilin Mei and Junhui Chen and Yu-Jie Yuan and Xinyue Cai and Guowei Huang and Xingyue Quan and Hang Xu and Li Zhang},
      year={2025},
      eprint={2503.22976},
      archivePrefix={arXiv},
      primaryClass={cs.CV},
      url={https://arxiv.org/abs/2503.22976}, 
}

@misc{shiri2024spatialmm,
      title={An Empirical Analysis on Spatial Reasoning Capabilities of Large Multimodal Models}, 
      author={Fatemeh Shiri and Xiao-Yu Guo and Mona Golestan Far and Xin Yu and Gholamreza Haffari and Yuan-Fang Li},
      year={2024},
      eprint={2411.06048},
      archivePrefix={arXiv},
      primaryClass={cs.CV},
      url={https://arxiv.org/abs/2411.06048}, 
}

@article{jia2025omnispatial,
  author = {Mengdi Jia and Zekun Qi and Shaochen Zhang and Wenyao Zhang and Xinqiang Yu and Jiawei He and He Wang and Li Yi},
  title = {Omnispatial: Towards comprehensive spatial reasoning benchmark for vision language models.},
  journal = {arXiv preprint arXiv:2506.03135},
  year = {2025}
}

@article{li2025spatialladder,
  author = {Hongxing Li and Dingming Li and Zixuan Wang and Yuchen Yan and Hang Wu and Wenqi Zhang and Yongliang Shen and Weiming Lu and Jun Xiao and Yueting Zhuang},
  title = {Spatialladder: Progressive training for spatial reasoning in vision-language models.},
  journal = {arXiv preprint arXiv:2510.08531},
  year = {2025}
}

@inproceedings{ma2025spatialllm,
  author = {Wufei Ma and Luoxin Ye and Celso~M de Melo and Alan Yuille and Jieneng Chen},
  title = {Spatialllm: A compound 3d-informed design towards spatially-intelligent large multimodal models.},
  booktitle = {Proceedings of the Computer Vision and Pattern Recognition Conference},
  year = {2025}
}

@misc{qwen2025qwen3vl,
  author = {{Qwen Team}},
  title = {{Qwen3-VL}: Sharper Vision, Deeper Thought, Broader Action},
  year = {2025},
  month = {September}
}

@misc{qwen3.5,
    title  = {{Qwen3.5}: Towards Native Multimodal Agents},
    author = {{Qwen Team}},
    month  = {February},
    year   = {2026},
    url    = {https://qwen.ai/blog?id=qwen3.5}
}

@article{zadeh1965fuzzy,
  author = {Zadeh, Lotfi A.},
  title = {Fuzzy sets},
  journal = {Information and Control},
  volume = {8},
  number = {3},
  pages = {338--353},
  year = {1965}
}

@inproceedings{
wu2025reinforcing,
title={Reinforcing Spatial Reasoning in Vision-Language Models with Interwoven Thinking and Visual Drawing},
author={Junfei Wu and Jian Guan and Kaituo Feng and Qiang Liu and Shu Wu and Liang Wang and Wei Wu and Tieniu Tan},
booktitle={The Thirty-ninth Annual Conference on Neural Information Processing Systems},
year={2026},
url={https://openreview.net/forum?id=yyWeSAsOhs}
}

@inproceedings{yin2025spatial,
  author = {Baiqiao Yin and Qineng Wang and Pingyue Zhang and Jianshu Zhang and Kangrui Wang and Zihan Wang and Jieyu Zhang and Keshigeyan Chandrasegaran and Han Liu and Ranjay Krishna and et~al},
  title = {Spatial mental modeling from limited views.},
  booktitle = {Structural Priors for Vision Workshop at ICCV'25},
  year = {2025}
}

@article{yu2025far,
  author = {Songsong Yu and Yuxin Chen and Hao Ju and Lianjie Jia and Fuxi Zhang and Shaofei Huang and Yuhan Wu and Rundi Cui and Binghao Ran and Zaibin Zhang and et~al},
  title = {How far are vlms from visual spatial intelligence? a benchmark-driven perspective.},
  journal = {arXiv preprint arXiv:2509.18905},
  year = {2025}
}

@misc{pySpatial,
      title={pySpatial: Generating 3D Visual Programs for Zero-Shot Spatial Reasoning}, 
      author={Zhanpeng Luo and Ce Zhang and Silong Yong and Cunxi Dai and Qianwei Wang and Haoxi Ran and Guanya Shi and Katia Sycara and Yaqi Xie},
      year={2026},
      eprint={2603.00905},
      archivePrefix={arXiv},
      primaryClass={cs.CV},
      url={https://arxiv.org/abs/2603.00905}, 
}

@misc{wu2025transductivevisualprogrammingevolving,
      title={Transductive Visual Programming: Evolving Tool Libraries from Experience for Spatial Reasoning}, 
      author={Shengguang Wu and Xiaohan Wang and Yuhui Zhang and Hao Zhu and Serena Yeung-Levy},
      year={2025},
      eprint={2512.20934},
      archivePrefix={arXiv},
      primaryClass={cs.CV},
      url={https://arxiv.org/abs/2512.20934}, 
}

@misc{GCA,
      title={Geometrically-Constrained Agent for Spatial Reasoning}, 
      author={Zeren Chen and Xiaoya Lu and Zhijie Zheng and Pengrui Li and Lehan He and Yijin Zhou and Jing Shao and Bohan Zhuang and Lu Sheng},
      year={2025},
      eprint={2511.22659},
      archivePrefix={arXiv},
      primaryClass={cs.AI},
      url={https://arxiv.org/abs/2511.22659}, 
}

@inproceedings{wang2025spatial457,
  title     = {Spatial457: A Diagnostic Benchmark for 6D Spatial Reasoning of Large Multimodal Models},
  author    = {Wang, Xingrui and Ma, Wufei and Zhang, Tiezheng and de Melo, Celso M and Chen, Jieneng and Yuille, Alan},
  booktitle = {Proceedings of the IEEE/CVF Conference on Computer Vision and Pattern Recognition (CVPR)},
  pages     = {24669--24679},
  year      = {2025},
  url       = {https://arxiv.org/abs/2502.08636}
}

@inproceedings{
  zhang2025do,
  title={Do Vision-Language Models Represent Space and How? Evaluating Spatial Frame of Reference under Ambiguities},
  author={Zheyuan Zhang and Fengyuan Hu and Jayjun Lee and Freda Shi and Parisa Kordjamshidi and Joyce Chai and Ziqiao Ma},
  booktitle={The Thirteenth International Conference on Learning Representations},
  year={2025},
  url={https://openreview.net/forum?id=84pDoCD4lH}
}

@inproceedings{premsri-kordjamshidi-2025-forest,
    title = "{F}o{REST}: Frame of Reference Evaluation in Spatial Reasoning Tasks",
    author = "Premsri, Tanawan  and
      Kordjamshidi, Parisa",
    editor = "Christodoulopoulos, Christos  and
      Chakraborty, Tanmoy  and
      Rose, Carolyn  and
      Peng, Violet",
    booktitle = "Proceedings of the 2025 Conference on Empirical Methods in Natural Language Processing",
    month = nov,
    year = "2025",
    address = "Suzhou, China",
    publisher = "Association for Computational Linguistics",
    url = "https://aclanthology.org/2025.emnlp-main.1772/",
    doi = "10.18653/v1/2025.emnlp-main.1772",
    pages = "34977--35003",
    ISBN = "979-8-89176-332-6"
}

@inproceedings{premsri2026forsaleframereferenceguidedspatial,
  title     = {{FoR-SALE}: Frame of Reference-guided Spatial Adjustment in {LLM}-based Diffusion Editing},
  author    = {Tanawan Premsri and Parisa Kordjamshidi},
  booktitle = {Conference on Language Modeling},
  year      = {2026},
  url       = {https://arxiv.org/abs/2509.23452}
}

@misc{gholami2025spatial,
      title={Spatial Reasoning with Vision-Language Models in Ego-Centric Multi-View Scenes}, 
      author={Mohsen Gholami and Ahmad Rezaei and Zhou Weimin and Sitong Mao and Shunbo Zhou and Yong Zhang and Mohammad Akbari},
      year={2025},
      eprint={2509.06266},
      archivePrefix={arXiv},
      primaryClass={cs.CV},
      url={https://arxiv.org/abs/2509.06266}, 
}

@inproceedings{marsili2025visual,
  title={Visual Agentic AI for Spatial Reasoning with a Dynamic API},
  author={Marsili, Damiano and Agrawal, Rohun and Yue, Yisong and Gkioxari, Georgia},
  booktitle={Proceedings of the Computer Vision and Pattern Recognition Conference},
  pages={19446--19455},
  year={2025}
}

@misc{zhang2026spinbenchperspectiverotationlens,
      title={SpinBench: Perspective and Rotation as a Lens on Spatial Reasoning in VLMs}, 
      author={Yuyou Zhang and Radu Corcodel and Chiori Hori and Anoop Cherian and Ding Zhao},
      year={2026},
      eprint={2509.25390},
      archivePrefix={arXiv},
      primaryClass={cs.CV},
      url={https://arxiv.org/abs/2509.25390}, 
}

@misc{li2025industrynavexploringspatialreasoning,
      title={IndustryNav: Exploring Spatial Reasoning of Embodied Agents in Dynamic Industrial Navigation}, 
      author={Yifan Li and Lichi Li and Anh Dao and Xinyu Zhou and Yicheng Qiao and Zheda Mai and Daeun Lee and Zichen Chen and Zhen Tan and Mohit Bansal and Yu Kong},
      year={2025},
      eprint={2511.17384},
      archivePrefix={arXiv},
      primaryClass={cs.RO},
      url={https://arxiv.org/abs/2511.17384}, 
}

@misc{sohn2025embodied4cmeasuringmattersembodied,
      title={Embodied4C: Measuring What Matters for Embodied Vision-Language Navigation}, 
      author={Tin Stribor Sohn and Maximilian Dillitzer and Jason J. Corso and Eric Sax},
      year={2025},
      eprint={2512.18028},
      archivePrefix={arXiv},
      primaryClass={cs.RO},
      url={https://arxiv.org/abs/2512.18028}, 
}

@misc{zhao2026espire,
      title={ESPIRE: A Diagnostic Benchmark for Embodied Spatial Reasoning of Vision-Language Models}, 
      author={Yanpeng Zhao and Wentao Ding and Hongtao Li and Baoxiong Jia and Zilong Zheng},
      year={2026},
      eprint={2603.13033},
      archivePrefix={arXiv},
      primaryClass={cs.CV},
      url={https://arxiv.org/abs/2603.13033}, 
}

@misc{cvbench,
      title={Cambrian-1: A Fully Open, Vision-Centric Exploration of Multimodal LLMs}, 
      author={Shengbang Tong and Ellis Brown and Penghao Wu and Sanghyun Woo and Manoj Middepogu and Sai Charitha Akula and Jihan Yang and Shusheng Yang and Adithya Iyer and Xichen Pan and Ziteng Wang and Rob Fergus and Yann LeCun and Saining Xie},
      year={2024},
      eprint={2406.16860},
      archivePrefix={arXiv},
      primaryClass={cs.CV},
      url={https://arxiv.org/abs/2406.16860}, 
}

@misc{spatialsense,
      title={SpatialSense: An Adversarially Crowdsourced Benchmark for Spatial Relation Recognition}, 
      author={Kaiyu Yang and Olga Russakovsky and Jia Deng},
      year={2019},
      eprint={1908.02660},
      archivePrefix={arXiv},
      primaryClass={cs.CV},
      url={https://arxiv.org/abs/1908.02660}, 
}

@misc{gsrbench,
      title={GSR-BENCH: A Benchmark for Grounded Spatial Reasoning Evaluation via Multimodal LLMs}, 
      author={Navid Rajabi and Jana Kosecka},
      year={2024},
      eprint={2406.13246},
      archivePrefix={arXiv},
      primaryClass={cs.CL},
      url={https://arxiv.org/abs/2406.13246}, 
}

@misc{embspatialbench,
      title={EmbSpatial-Bench: Benchmarking Spatial Understanding for Embodied Tasks with Large Vision-Language Models}, 
      author={Mengfei Du and Binhao Wu and Zejun Li and Xuanjing Huang and Zhongyu Wei},
      year={2024},
      eprint={2406.05756},
      archivePrefix={arXiv},
      primaryClass={cs.AI},
      url={https://arxiv.org/abs/2406.05756}, 
}

@inproceedings{basicspatialabilities,
   title={Defining and Evaluating Visual Language Models’ Basic Spatial Abilities: A Perspective from Psychometrics},
   url={http://dx.doi.org/10.18653/v1/2025.acl-long.567},
   DOI={10.18653/v1/2025.acl-long.567},
   booktitle={Proceedings of the 63rd Annual Meeting of the Association for Computational Linguistics (Volume 1: Long Papers)},
   publisher={Association for Computational Linguistics},
   author={Xu, Wenrui and Lyu, Dalin and Wang, Weihang and Feng, Jie and Gao, Chen and Li, Yong},
   year={2025},
   pages={11571–11590} }

@misc{sat,
      title={SAT: Dynamic Spatial Aptitude Training for Multimodal Language Models}, 
      author={Arijit Ray and Jiafei Duan and Ellis Brown and Reuben Tan and Dina Bashkirova and Rose Hendrix and Kiana Ehsani and Aniruddha Kembhavi and Bryan A. Plummer and Ranjay Krishna and Kuo-Hao Zeng and Kate Saenko},
      year={2025},
      eprint={2412.07755},
      archivePrefix={arXiv},
      primaryClass={cs.CV},
      url={https://arxiv.org/abs/2412.07755}, 
}

@misc{legopuzzles,
      title={LEGO-Puzzles: How Good Are MLLMs at Multi-Step Spatial Reasoning?}, 
      author={Kexian Tang and Junyao Gao and Yanhong Zeng and Haodong Duan and Yanan Sun and Zhening Xing and Wenran Liu and Kaifeng Lyu and Kai Chen},
      year={2025},
      eprint={2503.19990},
      archivePrefix={arXiv},
      primaryClass={cs.AI},
      url={https://arxiv.org/abs/2503.19990}, 
}

@misc{videomsr,
      title={Video-MSR: Benchmarking Multi-hop Spatial Reasoning Capabilities of MLLMs}, 
      author={Rui Zhu and Xin Shen and Shuchen Wu and Chenxi Miao and Xin Yu and Yang Li and Weikang Li and Deguo Xia and Jizhou Huang},
      year={2026},
      eprint={2601.09430},
      archivePrefix={arXiv},
      primaryClass={cs.CV},
      url={https://arxiv.org/abs/2601.09430}, 
}

@misc{s3bench,
      title={See, Remember, Explore: A Benchmark and Baselines for Streaming Spatial Reasoning}, 
      author={Yuxi Wei and Wei Huang and Qirui Chen and Lu Hou and Xiaojuan Qi},
      year={2026},
      eprint={2603.23864},
      archivePrefix={arXiv},
      primaryClass={cs.CV},
      url={https://arxiv.org/abs/2603.23864}, 
}

@misc{gemini31,
  title        = {Gemini 3.1 Pro Model Card},
  author       = {{Google DeepMind}},
  year         = {2026},
  month        = feb,
  howpublished = {\url{https://deepmind.google/models/model-cards/gemini-3-1-pro/}},
  note         = {Published February 2026}
}

@misc{gemini25,
      title={Gemini 2.5: Pushing the Frontier with Advanced Reasoning, Multimodality, Long Context, and Next Generation Agentic Capabilities}, 
      author={Gheorghe Comanici and Eric Bieber and Mike Schaekermann and Ice Pasupat and Noveen Sachdeva and Inderjit Dhillon and Marcel Blistein and Ori Ram and Dan Zhang and Evan Rosen and Luke Marris and Sam Petulla and Colin Gaffney and Gemini Team},
      year={2025},
      eprint={2507.06261},
      archivePrefix={arXiv},
      primaryClass={cs.CL},
      url={https://arxiv.org/abs/2507.06261}, 
}

@misc{clark2026molmo2openweightsdata,
      title={Molmo2: Open Weights and Data for Vision-Language Models with Video Understanding and Grounding}, 
      author={Christopher Clark and Jieyu Zhang and Zixian Ma and Jae Sung Park and Mohammadreza Salehi and Rohun Tripathi and Sangho Lee and Zhongzheng Ren and Chris Dongjoo Kim and Yinuo Yang and Vincent Shao and Yue Yang and Weikai Huang and Ziqi Gao and Taira Anderson and Jianrui Zhang and Jitesh Jain and George Stoica and Winson Han and Ali Farhadi and Ranjay Krishna},
      year={2026},
      eprint={2601.10611},
      archivePrefix={arXiv},
      primaryClass={cs.CV},
      url={https://arxiv.org/abs/2601.10611}, 
}

@misc{singh2026openaigpt5card,
      title={OpenAI GPT-5 System Card}, 
      author={{OpenAI Team}},
      year={2026},
      eprint={2601.03267},
      archivePrefix={arXiv},
      primaryClass={cs.CL},
      url={https://arxiv.org/abs/2601.03267}, 
}

@misc{remyxai2025spaceqwen25vl3b,
  title        = {{SpaceQwen2.5-VL-3B-Instruct}},
  author       = {{Remyx AI}},
  year         = {2025},
  howpublished = {\url{https://huggingface.co/remyxai/SpaceQwen2.5-VL-3B-Instruct}},
  note         = {Hugging Face model repository; accessed 2026-05-07}
}

@misc{remyxai2025spaceom,
  title        = {{SpaceOm}},
  author       = {{Remyx AI}},
  year         = {2025},
  howpublished = {\url{https://huggingface.co/remyxai/SpaceOm}},
  note         = {Hugging Face model repository; accessed 2026-05-07}
}

@misc{remyxai2025spacethinkerqwen25vl3b,
  title        = {{SpaceThinker-Qwen2.5VL-3B}},
  author       = {{Remyx AI}},
  year         = {2025},
  howpublished = {\url{https://huggingface.co/remyxai/SpaceThinker-Qwen2.5VL-3B}},
  note         = {Hugging Face model repository; accessed 2026-05-07}
}

@misc{nvidia2025cosmosreason1physicalcommonsense,
      title={Cosmos-Reason1: From Physical Common Sense To Embodied Reasoning}, 
      author={{NVIDIA} and Alisson Azzolini and Junjie Bai and Hannah Brandon and Jiaxin Cao and Prithvijit Chattopadhyay and Huayu Chen and Jinju Chu and Yin Cui and Jenna Diamond and Yifan Ding and Liang Feng and Francesco Ferroni and Rama Govindaraju and Jinwei Gu and Siddharth Gururani and Imad El Hanafi and Zekun Hao and Jacob Huffman and Jingyi Jin and Brendan Johnson and Rizwan Khan and George Kurian and Elena Lantz and Nayeon Lee and Zhaoshuo Li and Xuan Li and Maosheng Liao and Tsung-Yi Lin and Yen-Chen Lin and Ming-Yu Liu and Xiangyu Lu and Alice Luo and Andrew Mathau and Yun Ni and Lindsey Pavao and Wei Ping and David W. Romero and Misha Smelyanskiy and Shuran Song and Lyne Tchapmi and Andrew Z. Wang and Boxin Wang and Haoxiang Wang and Fangyin Wei and Jiashu Xu and Yao Xu and Dinghao Yang and Xiaodong Yang and Zhuolin Yang and Jingxu Zhang and Xiaohui Zeng and Zhe Zhang},
      year={2025},
      eprint={2503.15558},
      archivePrefix={arXiv},
      primaryClass={cs.AI},
      url={https://arxiv.org/abs/2503.15558}, 
}

@article{depthanything3,
  title={Depth Anything 3: Recovering the visual space from any views},
  author={Haotong Lin and Sili Chen and Jun Hao Liew and Donny Y. Chen and Zhenyu Li and Guang Shi and Jiashi Feng and Bingyi Kang},
  journal={arXiv preprint arXiv:2511.10647},
  year={2025}
}

@inproceedings{
sam3,
title={{SAM} 3: Segment Anything with Concepts},
author={Nicolas Carion and Laura Gustafson and Yuan-Ting Hu and Shoubhik Debnath and Ronghang Hu and Didac Suris Coll-Vinent and Chaitanya Ryali and Kalyan Vasudev Alwala and Haitham Khedr and Andrew Huang and Jie Lei and Tengyu Ma and Baishan Guo and Arpit Kalla and Markus Marks and Joseph Greer and Meng Wang and Peize Sun and Roman R{\"a}dle and Triantafyllos Afouras and Effrosyni Mavroudi and Katherine Xu and Tsung-Han Wu and Yu Zhou and Liliane Momeni and RISHI HAZRA and Shuangrui Ding and Sagar Vaze and Francois Porcher and Feng Li and Siyuan Li and Aishwarya Kamath and Ho Kei Cheng and Piotr Dollar and Nikhila Ravi and Kate Saenko and Pengchuan Zhang and Christoph Feichtenhofer},
booktitle={The Fourteenth International Conference on Learning Representations},
year={2026},
url={https://openreview.net/forum?id=r35clVtGzw}
}

@inproceedings{yang2025mmsi,
  title={MMSI-Bench: A Benchmark for Multi-Image Spatial Intelligence},
  author={Yang, Sihan and Xu, Runsen and Xie, Yiman and Yang, Sizhe and Li, Mo and Lin, Jingli and Zhu, Chenming and Chen, Xiaochen and Duan, Haodong and Yue, Xiangyu and Lin, Dahua and Wang, Tai and Pang, Jiangmiao},
  booktitle={The Fourteenth International Conference on Learning Representations (ICLR)},
  year={2026}
}

@inproceedings{cut3r,
        Author = {Qianqian Wang and Yifei Zhang and Aleksander Holynski and Alexei A. Efros and Angjoo Kanazawa},
        Title = {Continuous 3D Perception Model with Persistent State},
        Year = {2025},
          booktitle={Proceedings of the IEEE/CVF Conference on Computer Vision and Pattern Recognition (CVPR)},
          pages={10510--10522},
        }

@misc{tiger,
      title={TIGeR: Tool-Integrated Geometric Reasoning in Vision-Language Models for Robotics},
      author={Yi Han and Enshen Zhou and Shanyu Rong and Jingkun An and Pengwei Wang and Zhongyuan Wang and Cheng Chi and Lu Sheng and Shanghang Zhang},
      year={2026},
      eprint={2510.07181},
      archivePrefix={arXiv},
      primaryClass={cs.RO},
      url={https://arxiv.org/abs/2510.07181},
}

@inproceedings{mirzaee-kordjamshidi-2023-disentangling,
    title = "Disentangling Extraction and Reasoning in Multi-hop Spatial Reasoning",
    author = "Mirzaee, Roshanak  and Kordjamshidi, Parisa",
    booktitle = "Findings of the Association for Computational Linguistics: EMNLP 2023",
    month = dec,
    year = "2023",
    address = "Singapore",
    publisher = "Association for Computational Linguistics",
    url = "https://aclanthology.org/2023.findings-emnlp.221",
    doi = "10.18653/v1/2023.findings-emnlp.221",
    pages = "3379--3397"
}

@inproceedings{mirzaee-etal-2021-spartqa,
    title = "{SPARTQA}: A Textual Question Answering Benchmark for Spatial Reasoning",
    author = "Mirzaee, Roshanak  and Rajaby Faghihi, Hossein  and Ning, Qiang  and Kordjamshidi, Parisa",
    booktitle = "Proceedings of the 2021 Conference of the North American Chapter of the Association for Computational Linguistics: Human Language Technologies",
    month = jun,
    year = "2021",
    address = "Online",
    publisher = "Association for Computational Linguistics",
    url = "https://aclanthology.org/2021.naacl-main.364",
    doi = "10.18653/v1/2021.naacl-main.364",
    pages = "4582--4598"
}

@misc{rajpal2026spatialreasoningmodalityswitching,
      title={Spatial Reasoning via Modality Switching Between Language and Symbolic Representation},
      author={Shreya Rajpal and Tanawan Premsri and Parisa Kordjamshidi},
      year={2026},
      eprint={2606.31285},
      archivePrefix={arXiv},
      primaryClass={cs.AI},
      url={https://arxiv.org/abs/2606.31285},
}

@misc{deepseekai2026deepseekv4highlyefficientmilliontoken,
      title={DeepSeek-V4: Towards Highly Efficient Million-Token Context Intelligence}, 
      author={{DeepSeek-AI}},
      year={2026},
      eprint={2606.19348},
      archivePrefix={arXiv},
      primaryClass={cs.CL},
      url={https://arxiv.org/abs/2606.19348}, 
}

@inproceedings{ma-etal-2026-breaking,
  title = {Breaking Down and Building Up: Mixture of Skill-Based Vision-and-Language Navigation Agents},
  author = {Ma, Tianyi and Zhang, Yue and Wang, Zehao and Kordjamshidi, Parisa},
  editor = {Liakata, Maria and Moreira, Viviane P. and Zhang, Jiajun and Jurgens, David},
  booktitle = {Proceedings of the 64th Annual Meeting of the {A}ssociation for {C}omputational {L}inguistics (Volume 1: Long Papers)},
  month = jul,
  year = {2026},
  address = {San Diego, California, United States},
  publisher = {Association for Computational Linguistics},
  url = {https://aclanthology.org/2026.acl-long.595/},
  doi = {10.18653/v1/2026.acl-long.595},
  pages = {13035--13065},
  ISBN = {979-8-89176-390-6}
}

@misc{wang2026orientv2unifyingorientation,
  title={Orient Anything V2: Unifying Orientation and Rotation Understanding},
  author={Zehan Wang and Ziang Zhang and Jiayang Xu and Jialei Wang and Tianyu Pang and Chao Du and HengShuang Zhao and Zhou Zhao},
  year={2026},
  eprint={2601.05573},
  archivePrefix={arXiv},
  primaryClass={cs.CV},
  url={https://arxiv.org/abs/2601.05573},
}

@misc{vteam2026glm45vglm41vthinkingversatilemultimodal,
      title={GLM-4.5V and GLM-4.1V-Thinking: Towards Versatile Multimodal Reasoning with Scalable Reinforcement Learning},
      author={V Team and Wenyi Hong and Wenmeng Yu and Xiaotao Gu and Guo Wang and Guobing Gan and Haomiao Tang and Jiale Cheng and Ji Qi and Junhui Ji and Lihang Pan and Shuaiqi Duan and Weihan Wang and Yan Wang and Yean Cheng and Zehai He and Zhe Su and Zhen Yang and Ziyang Pan and Aohan Zeng and Baoxu Wang and Bin Chen and Boyan Shi and Changyu Pang and Chenhui Zhang and Da Yin and Fan Yang and Guoqing Chen and Haochen Li and Jiale Zhu and Jiali Chen and Jiaxing Xu and Jiazheng Xu and Jing Chen and Jinghao Lin and Jinhao Chen and Jinjiang Wang and Junjie Chen and Leqi Lei and Letian Gong and Leyi Pan and Mingdao Liu and Mingde Xu and Mingzhi Zhang and Qinkai Zheng and Ruiliang Lyu and Shangqin Tu and Sheng Yang and Shengbiao Meng and Shi Zhong and Shiyu Huang and Shuyuan Zhao and Siyan Xue and Tianshu Zhang and Tianwei Luo and Tianxiang Hao and Tianyu Tong and Wei Jia and Wenkai Li and Xiao Liu and Xiaohan Zhang and Xin Lyu and Xinyu Zhang and Xinyue Fan and Xuancheng Huang and Yadong Xue and Yanfeng Wang and Yanling Wang and Yanzi Wang and Yifan An and Yifan Du and Yiheng Huang and Yilin Niu and Yiming Shi and Yu Wang and Yuan Wang and Yuanchang Yue and Yuchen Li and Yusen Liu and Yutao Zhang and Yuting Wang and Yuxuan Zhang and Zhao Xue and Zhengxiao Du and Zhenyu Hou and Zihan Wang and Peng Zhang and Debing Liu and Bin Xu and Juanzi Li and Minlie Huang and Yuxiao Dong and Jie Tang},
      year={2026},
      eprint={2507.01006},
      archivePrefix={arXiv},
      primaryClass={cs.CV},
      url={https://arxiv.org/abs/2507.01006},
}

@misc{gemmateam2026gemma4,
    title={Gemma 4 Technical Report},
    author={Gemma Team},
    year={2026},
    eprint={2607.02770},
    archivePrefix={arXiv},
    primaryClass={cs.CL},
    url={https://arxiv.org/abs/2607.02770},
}

\appendix
\section{Ethical Considerations and Artifacts}

This work uses and creates scientific artifacts, including vision-language models, 3D perception models, existing spatial reasoning benchmarks, and the proposed 3D FORCE benchmark. Existing artifacts are used for research evaluation and are cited in the relevant sections. 3D FORCE is a synthetic diagnostic benchmark built from rendered scenes, scene graphs, and template-based spatial expressions. It does not contain human subjects, personally identifying information, or real-world images of people. Generated instances were spot-checked for labeling errors and inappropriate content.

3D FORCE is intended for research on compositional spatial reasoning and should not be used as evidence that a system is safe for real-world robotic, navigation, or surveillance deployment.
We will release the 3D FORCE dataset for non-commercial research use under the Creative Commons Attribution-NonCommercial 4.0 International license (CC BY-NC 4.0), and the SATURN code under the Apache License 2.0.
The release will include only artifacts for which we have redistribution rights; third-party source assets, models, and benchmarks remain subject to their original licenses and terms of use.

\section{Details of Anchor-Aware Spatial Predicate Computation}
\label{app:spatial_predicates}

\paragraph{Anchor representation.}
The spatial engine represents objects, cameras, agents, and virtual viewers through a unified anchor abstraction.
Let
\[
\mathcal{E}=\{e_1,\ldots,e_M\}
\]
denote the set of entities available to the symbolic executor.
Each entity \(e_i\) is associated with an anchor state
\[
a_i=(\mathbf{x}_i,\mathbf{R}_i),
\]
where \(\mathbf{x}_i\in\mathbb{R}^3\) is a 3D position and \(\mathbf{R}_i\in\mathbb{R}^{3\times 3}\) is an orientation in the global coordinate frame.
Each anchor state induces a frame of reference (FoR).
The selected FoR anchor is denoted by
\[
a=(\mathbf{x}_a,\mathbf{R}_a),
\]
which may correspond to an object, camera, agent, or virtual viewer constructed during program execution.

For implementation, we use the semantic axes induced by each entity anchor:
\[
\mathbf{r}_i=\mathbf{R}_i\mathbf{e}_{\mathrm{right}},\quad
\mathbf{u}_i=\mathbf{R}_i\mathbf{e}_{\mathrm{up}},\quad
\mathbf{q}_i=\mathbf{R}_i\mathbf{e}_{\mathrm{front}},
\]
where \(\mathbf{r}_i\), \(\mathbf{u}_i\), and \(\mathbf{q}_i\) denote the right, up, and forward directions of entity \(e_i\)'s anchor state.
All object, camera, agent, and virtual-viewer anchors are converted into this semantic-axis representation before predicate computation, avoiding ambiguity from different native camera or object coordinate conventions.

\paragraph{Anchor-conditioned predicates.}
For a selected FoR anchor \(a\) and spatial relation \(r\), the spatial engine produces an anchor-conditioned predicate matrix
\[
S^a_r\in[0,1]^{|\mathcal{E}|\times|\mathcal{E}|}.
\]
Each entry \(S^a_r[i,j]\) scores whether entity \(e_i\) is in relation \(r\) to entity \(e_j\), evaluated in the FoR induced by anchor \(a\).

Object-centric predicates are a special case in which the reference object supplies the FoR anchor.
When \(e_j\) is an oriented object, we define
\[
S^{\mathrm{obj}}_r[i,j]\equiv S^{a_j}_r[i,j].
\]
Thus, \(S^{\mathrm{obj}}_r[i,j]\) evaluates the relation between \(e_i\) and \(e_j\) from \(e_j\)'s own perspective.

\paragraph{General soft predicate scoring.}
For an anchor-conditioned relation, the target-reference displacement between entities \(e_i\) and \(e_j\) is expressed in the FoR induced by anchor \(a\).
Assuming \(\mathbf{R}_a\) maps coordinates from anchor \(a\)'s local frame to the global frame, we compute
\[
\Delta^a_{ij}=\mathbf{R}_a^\top(\mathbf{x}_i-\mathbf{x}_j).
\]
We also express the local orientations of the two entity anchors in the same FoR:
\[
\mathbf{R}^a_i=\mathbf{R}_a^\top\mathbf{R}_i,\quad
\mathbf{R}^a_j=\mathbf{R}_a^\top\mathbf{R}_j.
\]
Here, \(\Delta^a_{ij}\) is the displacement from reference entity \(e_j\) to target entity \(e_i\), expressed in anchor \(a\)'s local coordinate frame.
The matrices \(\mathbf{R}^a_i\) and \(\mathbf{R}^a_j\) are the orientations of entities \(e_i\) and \(e_j\), also expressed in anchor \(a\)'s FoR.

Each spatial relation is implemented through a relation-specific evidence function
\[
h_r(\Delta^a_{ij},\mathbf{R}^a_i,\mathbf{R}^a_j),
\]
which is mapped to a soft predicate score:
\[
S^a_r[i,j]
=
\sigma\!\left(
\frac{
h_r(\Delta^a_{ij},\mathbf{R}^a_i,\mathbf{R}^a_j)-m_r
}{\tau_r}
\right).
\]
Here, \(m_r\) is a relation-specific margin and \(\tau_r\) controls the softness of the predicate.
This formulation covers directional and orientation-based predicates using the same interface.

\paragraph{Scene-scale normalization.}
To make positional margins comparable across scenes, we normalize spatial evidence by a robust scene scale.
Let
\[
s_{\mathrm{scene}}
=
\operatorname{quantile}_{0.9}
\bigl\{
\|\mathbf{x}_p-\mathbf{x}_q\|_2
:
p<q
\bigr\},
\]
with a small constant \(\varepsilon\) for numerical stability.
The normalized displacement is
\[
\bar{\Delta}^a_{ij}
=
\frac{\Delta^a_{ij}}{s_{\mathrm{scene}}+\varepsilon}.
\]
Pairwise directional predicates use \(\bar{\Delta}^a_{ij}\); egocentric direction labels use bearing evidence that depends only on the direction of the displacement and is therefore scale-invariant by construction; orientation predicates use dot-product evidence between anchor forward axes and therefore do not require scene-scale normalization.

\paragraph{Default hyperparameters.}
We use a small set of fixed hyperparameters across all experiments.
These values are not tuned per benchmark or per task.
For directional predicates, we use
\[
m_{\mathrm{dir}}=0.0,\quad \tau_{\mathrm{dir}}\approx 0.0714.
\]
For directional-combination predicates, we use
\[
m_{\mathrm{comb}}=0.0,\quad \tau_{\mathrm{comb}}\approx 0.0714.
\]
For egocentric direction labels, \(m_{d}=0\) and \(\tau_{d}=1\); the bearing evidence carries its own scale.
For orientation-based predicates, we use
\[
m_{\parallel}=0.70,\quad
m_{\perp}=0.30,\quad
\tau_{\mathrm{ori}}\approx 0.0714.
\]

These defaults are summarized in Table~\ref{tab:spatial_hparams} and are fixed across all experiments; we do not tune them per benchmark.

\begin{table}[t]
\centering
\small
\setlength{\tabcolsep}{5pt}
\renewcommand{\arraystretch}{1.1}
\begin{tabular}{lcc}
\toprule
\textbf{Predicate type} & \textbf{Margin} & \textbf{Temperature} \\
\midrule
Directional & \(m_{\mathrm{dir}}=0.0\) & \(\tau_{\mathrm{dir}}\approx 0.0714\) \\
Directional combination & \(m_{\mathrm{comb}}=0.0\) & \(\tau_{\mathrm{comb}}\approx 0.0714\) \\
Direction label & \(m_{d}=0.0\) & \(\tau_{d}=1\) \\
Parallel & \(m_{\parallel}=0.70\) & \(\tau_{\mathrm{ori}}\approx 0.0714\) \\
Perpendicular & \(m_{\perp}=0.30\) & \(\tau_{\mathrm{ori}}\approx 0.0714\) \\
\bottomrule
\end{tabular}
\caption{Default hyperparameters for soft spatial predicates. Directional predicates project the local displacement onto the anchor's semantic axes; orientation predicates use dot-product evidence between anchor forward vectors. Direction labels use bearing-alignment evidence \(h_d=2\ln\cot(\delta_d/2)\).}
\label{tab:spatial_hparams}
\end{table}

\paragraph{Directional predicates.}
Directional relations are computed from signed projections of the normalized displacement \(\bar{\Delta}^a_{ij}\) onto the local axes of the selected FoR anchor \(a\).
Because \(\bar{\Delta}^a_{ij}\) is already expressed in anchor \(a\)'s local coordinate frame, we use the canonical semantic axes.
In a \(Y\)-up right-handed coordinate system, we define
\[
h_{\mathrm{right}}(\bar{\Delta}^a_{ij})
=
\langle \bar{\Delta}^a_{ij}, \mathbf{e}_{\mathrm{right}}\rangle,
\]
\[
h_{\mathrm{left}}(\bar{\Delta}^a_{ij})
=
-\langle \bar{\Delta}^a_{ij}, \mathbf{e}_{\mathrm{right}}\rangle,
\]
\[
h_{\mathrm{above}}(\bar{\Delta}^a_{ij})
=
\langle \bar{\Delta}^a_{ij}, \mathbf{e}_{\mathrm{up}}\rangle,
\]
\[
h_{\mathrm{below}}(\bar{\Delta}^a_{ij})
=
-\langle \bar{\Delta}^a_{ij}, \mathbf{e}_{\mathrm{up}}\rangle,
\]
\[
h_{\mathrm{front}}(\bar{\Delta}^a_{ij})
=
-\langle \bar{\Delta}^a_{ij}, \mathbf{e}_{\mathrm{front}}\rangle,
\]
\[
h_{\mathrm{behind}}(\bar{\Delta}^a_{ij})
=
\langle \bar{\Delta}^a_{ij}, \mathbf{e}_{\mathrm{front}}\rangle.
\]
Here \textsc{front} follows occluder semantics: ``\(e_i\) is in front of \(e_j\)'' means \(e_i\) lies nearer the observer along the anchor's viewing axis.
The corresponding soft predicate is
\[
S^a_r[i,j]
=
\sigma\!\left(
\frac{
h_r(\bar{\Delta}^a_{ij})-m_{\mathrm{dir}}
}{\tau_{\mathrm{dir}}}
\right).
\]

Directional combinations such as \textsc{front-left}\ and \textsc{behind-right} are computed from the same local displacement vector.
For a relation \(r\) composed of component directions \(\mathcal{C}(r)\), we define
\[
h_r(\bar{\Delta}^a_{ij})
=
\min_{c\in\mathcal{C}(r)}
h_c(\bar{\Delta}^a_{ij}),
\]
and score it as
\[
S^a_r[i,j]
=
\sigma\!\left(
\frac{
h_r(\bar{\Delta}^a_{ij})-m_{\mathrm{comb}}
}{\tau_{\mathrm{comb}}}
\right).
\]
This implements combined directions as soft geometric conjunctions without requiring additional tool calls or query-specific geometric programs.

\paragraph{Egocentric direction labels.}
In addition to the pairwise axial predicates above, \method exposes egocentric direction \emph{labels} that classify where a target lies relative to an anchor: eight horizontal directions with target bearings \(\theta_d\in\{0^\circ,\pm45^\circ,\pm90^\circ,\pm135^\circ,180^\circ\}\), plus \textsc{above} and \textsc{below}. Their evidence is bearing alignment: let \(\delta_d\) be the angle between the displacement \(\Delta^a_{ij}\) --- projected onto the anchor's horizontal plane for horizontal labels --- and the label's target direction. The evidence and hyperparameters are
\[
h_d = 2\ln\cot(\delta_d/2),
\qquad
m_d = 0,
\qquad
\tau_d = 1,
\]
which under the general scoring interface reduces to the closed form
\[
S^a_d=\sigma(h_d)=\cos^2(\delta_d/2)=\tfrac12\bigl(1+\cos\delta_d\bigr);
\]
the implementation computes this closed form directly. Because \(\delta_d\) depends only on the direction of the displacement, label scores are scale-invariant and require no scene-scale normalization, and diagonal labels are ordinary labels at \(\pm45^\circ\) bearings rather than conjunctions. \(\arg\max_d S^a_d\) is exactly nearest-direction classification, so the label family acts as a soft direction classifier.

\paragraph{Orientation-based predicates.}
Orientation predicates compare local orientation axes of entity anchors.
Let
\[
\mathbf{q}_i^a=\mathbf{R}_a^\top\mathbf{R}_i\mathbf{e}_{\mathrm{front}},
\quad
\mathbf{q}_j^a=\mathbf{R}_a^\top\mathbf{R}_j\mathbf{e}_{\mathrm{front}}
\]
denote the forward axes of entities \(e_i\) and \(e_j\), expressed in the FoR induced by anchor \(a\).

A parallel predicate is scored as
\[
S^a_{\mathrm{parallel}}[i,j]
=
\sigma\!\left(
\frac{
|\langle \mathbf{q}_i^a,\mathbf{q}_j^a\rangle|-m_{\parallel}
}{\tau_{\mathrm{ori}}}
\right).
\]
A perpendicular predicate is scored as
\[
S^a_{\mathrm{perpendicular}}[i,j]
=
\sigma\!\left(
\frac{
m_{\perp}-|\langle \mathbf{q}_i^a,\mathbf{q}_j^a\rangle|
}{\tau_{\mathrm{ori}}}
\right).
\]
A fixed-angle predicate with target angle \(\theta^\star\) is scored using dot products:
\[
S^a_{\theta^\star}[i,j]
=
\sigma\!\left(
\frac{
m_\theta-\left|(\mathbf{q}_i^a)^\top\mathbf{q}_j^a-\cos\theta^\star\right|
}{\tau_{\mathrm{ori}}}
\right).
\]
This implements relations such as \textsc{parallel}, \textsc{perpendicular}, and fixed-angle predicates such as \(60^\circ\), without explicitly computing angles; \(m_\theta\) is the margin of the instantiated relation, \(m_\parallel\) or \(m_\perp\) in Table~\ref{tab:spatial_hparams}.
If signed orientation is needed, the predicate can instead use cross-product evidence.

\paragraph{Constructed anchors.}
In addition to object- and camera-induced anchors, \method supports virtual anchors instantiated at arbitrary 3D positions and orientations.
Existing anchors can also be transformed through rotation and translation.
A constructed anchor supplies a frame of reference to the same predicate interface without enlarging the entity set \(\mathcal{E}\), allowing the symbolic program to evaluate spatial relations from arbitrary viewpoints.

Overall, the spatial engine is predicate-extensible: any relation that can be expressed as a soft evidence function over estimated anchor states can be added as an \(h_r\).
Our current instantiation covers directional, directional-combination, and orientation-based predicates.

\section{Symbolic Composition Functions}
\label{app:composition_functions}

A component of \method is the symbolic executor, which runs the LLM-generated program. SATURN uses the hybrid soft-execution model from prior Pythonic neuro-symbolic reasoning, adapting it to compose both semantic grounding scores and geometry-derived spatial predicate scores.

\paragraph{Soft Compositional Reasoning.}
To reason about visual concepts themselves, we employ a set of soft logical operations based on fuzzy logic principles similar to~\citep{hsu2024s,nesycoco,kamali2026neptune}. Instead of operating on binary true/false values, these operations work directly on the uncertainty scores obtained from the grounding interface and the spatial engine. This is implemented through custom data structures that encapsulate scores for a given predicate and overload standard Python operators, such as \texttt{\&} for AND, \texttt{|} for OR, and \texttt{not} for negation, to perform the corresponding logical operations. Details of these operations are shown in Table~\ref{tab:expressions}. The table lists both a differentiable form and the form the executor actually evaluates; since \method performs inference only, we use the non-differentiable executed forms. For example, when the program executes $\alpha_x \,\&\, \alpha_y$, our framework takes the element-wise minimum of the two corresponding score vectors, implementing the fuzzy t-norm for conjunction. Similarly, relational composition between an object-centric score vector $\alpha_x$ and a relational score matrix $\beta_{xy}$ preserves uncertainty while propagating object-level evidence through spatial relations.

\begin{table*}[h!]
\centering
\resizebox{0.85\textwidth}{!}{%
\begin{tabular}{@{}lllcc@{}}
\toprule
\textbf{Syntax} & \textbf{Logical Form} & \textbf{Description} & \textbf{Differentiable Form} & \textbf{Inference} \\
\midrule
$\alpha_x.\texttt{exists}()$ & $\exists x \, \alpha_x$ & Existential quantification & $\max(\alpha_x)$ & $\max(\alpha_x)$ \\
$\alpha_x.\texttt{forall}()$ & $\forall x \, \alpha_x$ & Universal quantification & $\min(\alpha_x)$ & $\mathbf{1}\!\left[\min_x \alpha_x > 0.5\right]$ \\
$\alpha_x \,\&\, \alpha_y$ & $\alpha_x \land \alpha_y$ & Logical conjunction & $\min(\alpha_x, \alpha_y)$ & $\min(\alpha_x, \alpha_y)$ \\
$\alpha_x \,\&\, \beta_{xy}$ & $\alpha_x \land \beta_{xy}$ & Relational conjunction & $\min(\alpha_x, \beta_{xy})$ & $\min(\alpha_x, \beta_{xy})$ \\
$\alpha_x \,|\, \alpha_y$ & $\alpha_x \lor \alpha_y$ & Logical disjunction & $\max(\alpha_x, \alpha_y)$ & $\max(\alpha_x, \alpha_y)$ \\
$\alpha_x.\texttt{implies}(\alpha_y)$ & $\alpha_x \rightarrow \alpha_y$ & Logical implication & $\max(1-\alpha_x, \alpha_y)$ & $\max(1-\alpha_x, \alpha_y)$ \\
$\texttt{not } \alpha_x$ & $\neg \alpha_x$ & Logical negation & $1 - \alpha_x$ & $1 - \alpha_x$ \\

$\alpha_x.\texttt{iota}(\text{var})$ & $\iota(\text{var}, \alpha)$ & Best-match & $\frac{\alpha - \min \alpha}{\max \alpha - \min \alpha + \varepsilon}$ & $\frac{\alpha - \min \alpha}{\max \alpha - \min \alpha + \varepsilon}$ \\

$\alpha_x.\texttt{count}()$ & $\text{count}(\alpha_x)$ & Count & $\sum_x \alpha_x$ & $\sum_x \mathbf{1}\!\left[\alpha_x \ge 0.5\right]$ \\
$s_1 == s_2$ & $s_1 = s_2$ & Equality & $\sigma\!\left(\dfrac{2\gamma - |s_1-s_2|}{2\gamma\tau}\right)$ & $\sigma\!\left(\dfrac{2\gamma - |s_1-s_2|}{2\gamma\tau}\right)$ \\
$s_1 > s_2$ & $s_1 > s_2$ & Inequality & $\sigma\!\left(\dfrac{s_1-s_2+\gamma}{\tau}\right)$ & $\mathbf{1}\!\left[s_1 > s_2\right]$ \\

\bottomrule
\end{tabular}
}
\caption{Soft compositional operators used in symbolic execution. Here, $\alpha$ represents an object-centric 1D score vector, $\beta$ relational kD ($k\geq2$) score matrix, $\tau=0.25$ is a temperature parameter, $\gamma=0.25$ is a margin.}
\label{tab:expressions}
\end{table*}

\paragraph{Imperative Reasoning.}
Our symbolic executor leverages a standard Python interpreter to handle the program's overall structure and control flow. This is possible because we define iteration and Boolean operations on concept objects, allowing the generated programs to express complex procedural logic, including conditionals, loops, and variable assignments. As a result, the framework retains the expressive power of a general-purpose programming language while allowing the underlying predicates to remain uncertainty-aware. Some operations, such as counting over string sets, may break the computation graph, but this hybrid design enables the system to reason flexibly under uncertainty while still supporting the program structures needed for compositional spatial reasoning.

\begin{table*}[t]
    \adjustbox{max width=\textwidth, center}{%
    \centering
    \small
    \setlength{\tabcolsep}{4pt}
    \begin{tabular}{lccccccc}
        \toprule
        \textbf{Benchmark}
        & \textbf{Multi-view}
        & \multicolumn{2}{c}{\textbf{FoR}}
        & \textbf{FoR Mixing}
        & \textbf{Compositionality}
        & \multicolumn{2}{c}{\textbf{Multi-hop}} \\
        \cmidrule(lr){3-4} \cmidrule(lr){7-8}
        &
        & \textbf{Camera}
        & \textbf{Object}
        & \textbf{}
        & \textbf{Type}
        & \textbf{Chains}
        & \textbf{Controlled depth} \\
        \midrule
        CLEVR~\citep{johnson2017clevr}
        & \xmark & \cmark & \xmark & \xmark & Relation & \cmark & \cmark \\
        BLINK~\citep{BLINK}
        & \cmark & \cmark & \xmark & \xmark & Perception & \xmark & \xmark \\
        SpaCE-10~\citep{space-10}
        & \xmark & \cmark & \xmark & \xmark & Skill & \xmark & \xmark \\
        3DSRBench~\citep{3dsrbench}
        & \xmark & \cmark & \cmark & \xmark & 3D skill & \xmark & \xmark \\
        SPHERE~\citep{sphere}
        & \xmark & \cmark & \cmark & \xmark & Skill & \xmark & \xmark \\
        ViewSpatial~\citep{viewspatial}
        & \xmark & \cmark & \cmark & \xmark & Perspective & \xmark & \xmark \\
        Spatial457~\citep{wang2025spatial457}
        & \xmark & \cmark & \cmark & \xmark & 6D relation & \xmark & \xmark \\
        FoREST$^\dagger$~\citep{premsri-kordjamshidi-2025-forest}
        & \xmark & \cmark & \cmark & \xmark & Perspective & \xmark & \xmark \\
        MMSI-Bench~\citep{yang2025mmsi}
         & \cmark & \cmark & \cmark & \xmark & Perspective & \cmark & \xmark \\
        OmniSpatial~\citep{jia2025omnispatial}
        & \xmark & \cmark & \cmark & \xmark & Skill & \xmark & \xmark \\
        MindCube~\citep{yin2025spatial}
        & \cmark & \cmark & \cmark & \xmark & Mental-model & \xmark & \xmark \\
        SPINBENCH~\citep{zhang2026spinbenchperspectiverotationlens}
        & \cmark & \cmark & \cmark & \xmark & Perspective & \xmark & \xmark \\
        COMFORT~\citep{zhang2025do}
& \xmark & \cmark & \cmark & \xmark & Perspective / FoR & \xmark & \xmark \\
Spatial-MM~\citep{shiri2024spatialmm}
& \xmark & \cmark & \cmark & \xmark & Relation & \cmark & \xmark \\
SPAR-Bench~\citep{sparbench}
& \cmark & \cmark & \xmark & \xmark & Skill & \xmark & \xmark \\
Ego3D-Bench~\citep{gholami2025spatial}
& \cmark & \cmark & \xmark & \xmark & 3D skill & \xmark & \xmark \\

MV-RoboBench~\citep{mvrobobench}
& \cmark & \cmark & \xmark & \xmark & Skill & \xmark & \xmark \\
        \midrule
        \textsc{\bench} (ours)
        & \cmark & \cmark & \cmark & \cmark & Relation & \cmark & \cmark \\
        \bottomrule

    \end{tabular}
    }
    \caption{
    Comparison of spatial reasoning benchmarks across evaluation capabilities. \cmark{} marks a capability as an explicitly evaluated axis; \xmark{} marks absence or only incidental support. \textbf{Multi-view}: whether the benchmark evaluates multiple images, views, or viewpoint changes together as part of the task design. \textbf{FoR}: whether questions are anchored to the camera/viewer frame or to an oriented object's intrinsic frame. \textbf{FoR Mixing}: whether a single query composes relations stated in different reference frames. \textbf{Compositionality Type}: \emph{Relation} chains atomic spatial relations between specific objects; \emph{Skill} composes heterogeneous capabilities such as recognition, localization, counting, or functional reasoning; \emph{3D skill} evaluates 3D spatial capabilities such as height, location, orientation, and multi-object reasoning; \emph{6D relation} evaluates spatial relations involving 3D position and orientation; \emph{Mental-model} requires constructing an internal scene representation across partial views; \emph{Perspective} evaluates viewpoint or frame-of-reference transformations; \emph{Perception} evaluates low-level visual perception skills. \textbf{Multi-hop / Chains}: questions chain $\geq$2 atomic spatial relations between specific objects. \textbf{Multi-hop / Controlled depth}: hop count and topology are explicit, controlled axes. $^\dagger$FoREST is primarily a text/LLM frame-of-reference benchmark.
    }
\label{tab:benchmark-capabilities}
\end{table*}

\section{Comparison to Existing Spatial Reasoning Benchmarks}
  \label{app:benchmark_comparison}

  Existing spatial reasoning benchmarks vary along several orthogonal axes:
  whether they evaluate multi-view inputs, whether they support both
  camera-centric and object-centric frames of reference (FoRs), whether they
  allow a single query to mix FoRs, the type of compositionality they
  evaluate, and whether they control multi-hop depth and topology.
  Table~\ref{tab:benchmark-capabilities} summarizes these axes for the most
  closely related benchmarks. \bench is the only benchmark in this set that
  explicitly controls all four axes simultaneously: multi-view, FoR mixing
  within a single query, relation-level compositionality, and controlled
  multi-hop depth.

\section{Benchmark Construction and Validation}
\label{app:benchmark_construction}

\subsection{Image Rendering Setup}
Our benchmark generation pipeline uses Blender3D as the simulation environment for image rendering.
We include 50 diverse backgrounds to increase visual variety across scenes. The scene generation process contains 12 object categories: airliner, biplane, double bus, fighter plane, minivan, school bus, scooter motorcycle, sedan, SUV, truck, horse, and tank.
Each object is assigned one of five colors—blue, brown, green, red, and yellow—with assignments chosen to preserve visual realism. For example, horses are always assigned the color brown rather than arbitrary colors from the palette.

\subsection{Image and Scene Graph Generation}

To render the underlying scenes, we use the image-generation pipeline of Spatial457~\citep{wang2025spatial457} with several modifications, including higher-resolution images, less ambiguous object categories and names, and a revised color palette.

For the standard multi-view setting, we render additional views by placing cameras at approximately uniform azimuth intervals around the scene.
For two, three, and four views, cameras are separated by roughly \(180^\circ\), \(120^\circ\), and \(90^\circ\), respectively.
This produces controlled viewpoint variation while keeping the query-relevant objects visible across the views.
For the partial-view setting, camera positions and orientations are randomized to create incomplete individual observations.
We constrain this randomization so that every pair of views shares at least one visible object, and the union of all views contains all objects required by the query.
Thus, partial-view examples require integrating information across images rather than solving the query from a single complete view. Each scene contains 6-12 objects.

Scene graphs are then constructed using each object’s 3D world coordinates (\texttt{X}, \texttt{Y}, \texttt{Z}), 2D image coordinates (\texttt{X'}, \texttt{Y'}), and \texttt{pose} direction derived from 3D orientation. Based on the resulting scene graph, we render additional views to create multi-image subsets of 2, 3, and 4 images, ensuring that all objects remain visible across all views.

\subsection{Benchmark Instance Generation}

\textbf{\puzzlebench}
To generate \puzzlebench, we use a simulation environment in which a set of images, along with their corresponding scene graphs, is rendered.
For each image, we randomly select $k$ objects from the scene graph to generate object descriptions, $T$, and sample $l$ object pairs from $k$ to generate the spatial relation descriptions $R$.
The object and relation descriptions are generated based on a set of sentence templates.
The number of hops of reasoning is limited to a maximum of five.
These instances constitute the set of positive examples in our benchmark.
To generate negative examples, we mutate one or more relations in the relation set by reversing their directions (e.g., "A to the left of B" is changed to "B is to the left of A").
For each image, the descriptions are generated multiple times under different camera perspectives to support multi-view evaluation.  Before generation, each instance is randomly assigned to either the positive or negative set, which determines whether the generated spatial configuration is satisfiable.
For each instance, the maximum number of objects and spatial relations is 5.
In total, we generate 1,150 questions, including 972 standard multi-view questions and 178 partial-view questions. Detailed statistics are provided in Table~\ref{tab:hop_distribution} and~\ref{tab:perspective_distribution}.

\textbf{\refbench}
To generate \refbench instances, we first construct a random topology graph for each instance to encode multi-hop reasoning, where nodes represent entities and edges represent spatial relations.
We consider three topological structures, chain, star, and hybrid, as illustrated in Figure~\ref{fig:benchmark_overview}.
Next, we identify objects and spatial relations that align with the constructed topology using a brute-force search over all feasible object–relation combinations in the rendered scene graph.
We enforce that the full relational chain is necessary to identify the target, such that any partial subset of relations is insufficient, and ensure that each instance contains a unique solution.
The number of hops of reasoning is limited to a maximum of six.
After selecting the objects and relations, we generate referring expressions using predefined spatial-relation templates conditioned on the topology graph.
Similar to \puzzlebench, the descriptions are generated multiple times under different camera perspectives to support multi-view evaluation.  Separate parameter configurations are defined for each topology structure used to encode multi-hop reasoning. For the chain structure, we allow up to 5 reasoning hops. For the star structure, the number of hops is limited to 2–4 to ensure that the complete query is required to identify the target object, while any partial query remains insufficient. For the hybrid structure, we generate only 4-hop and 6-hop instances, as the hybrid topology requires an even number of reasoning hops for valid construction.
In total, we generate 2,088 questions for \refbench. Detailed statistics are provided in Table~\ref{tab:hop_distribution} and~\ref{tab:perspective_distribution}. We do not define train/dev/test splits because 3D FORCE is intended as an evaluation-only diagnostic benchmark for zero-shot compositional spatial reasoning.

\begin{table*}[t]
\centering
\small
\begin{tabular}{c|ccccccc}
    \toprule
    \textbf{Benchmark} & \textbf{0-Hop} & \textbf{1-Hop} & \textbf{2-Hop} & \textbf{3-Hop} & \textbf{4-Hop} & \textbf{5-Hop} & \textbf{6-Hop} \\
    \midrule
    \puzzlebench & 150 & 158 & 199 & 223 & 204 & 216 & - \\
    \refbench & 200 & 320 & 360 & 356 & 342 & 280 & 230 \\
    \bottomrule
    \end{tabular}
    \caption{Distribution of reasoning hops in \puzzlebench and \refbench. \refbench additionally includes 6-hop reasoning instances.}
    \label{tab:hop_distribution}
\end{table*}

\begin{table*}[t]
\centering
\small
\begin{tabular}{l|cc}
    \toprule
    \textbf{Setting} & \textbf{\puzzlebench} & \textbf{\refbench} \\
    \midrule
    Single-image & 313 & 524 \\
    Multi-image & 285 & 631 \\
    Object-centric & 200 & 435 \\
    Object-centric + Single-image & 228 & 154 \\
    Object-centric + Multi-image & 124 & 344 \\
    \midrule
    1 View & 232 & 436 \\
    2 Views & 228 & 463 \\
    3 Views & 267 & 548 \\
    4 Views & 423 & 641 \\
    \bottomrule
    \end{tabular}
    \caption{Distribution of perspective settings and number of views in \puzzlebench and \refbench.}
    \label{tab:perspective_distribution}
\end{table*}

\begin{figure*}[t]
    \centering
    \includegraphics[width=\linewidth]{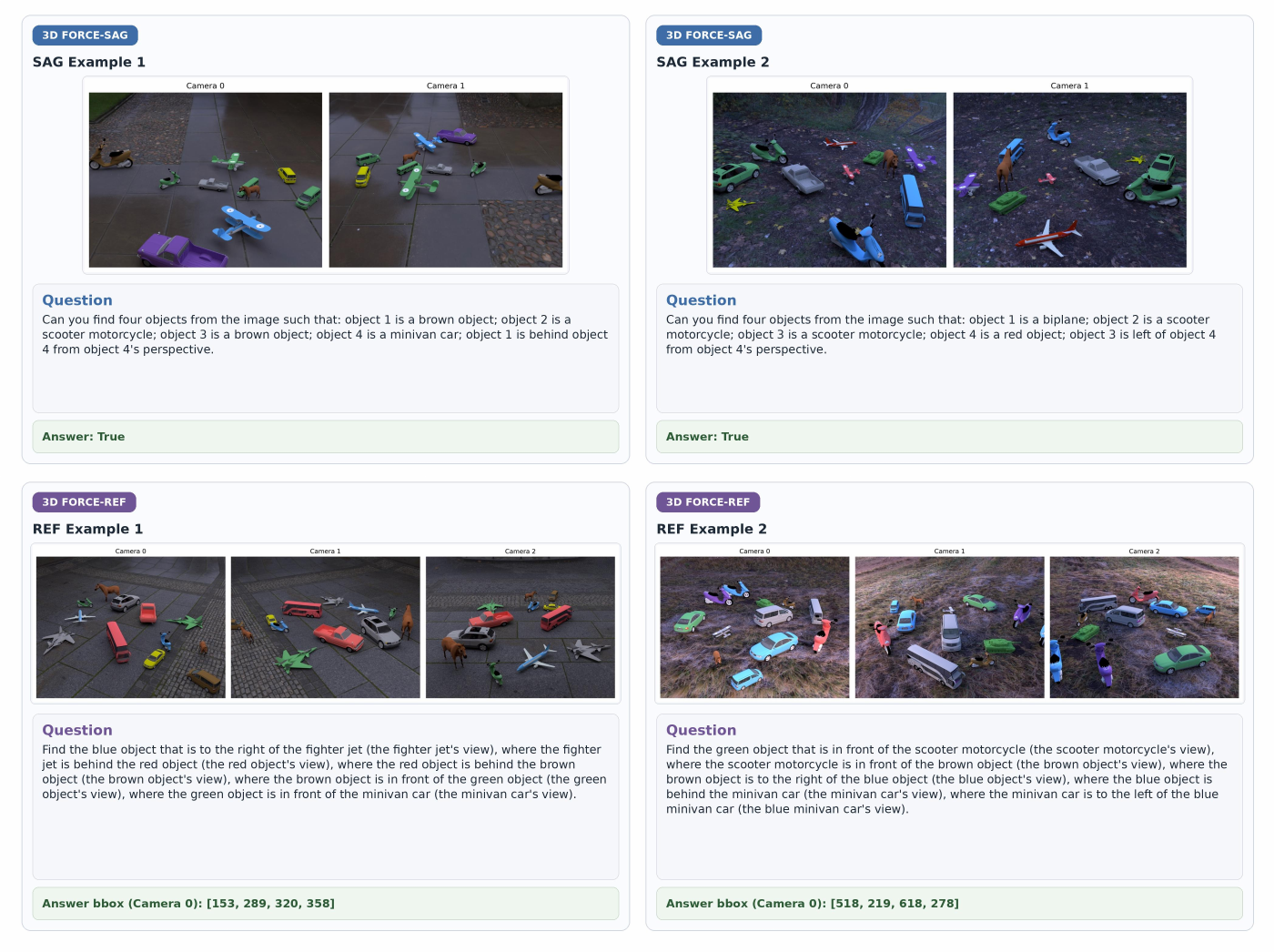}
    \caption{Examples from \puzzlebench (top) and \refbench (bottom), showing two examples from each benchmark together with the corresponding ground-truth answers.}
    \label{fig:3dforce_examples}
\end{figure*}

\subsection{Examples}

We provide the examples of \bench, including \puzzlebench and \refbench in Figure~\ref{fig:3dforce_examples}.

\section{Text-Stated Camera-Pose Constraint Extraction}
\label{app:pose_constraints}

Some multi-view inputs contain explicit natural-language facts about the camera setup.
For example, a question may state that one image was captured after rotating the camera \(90^\circ\) from another image, or that several images were captured from the same physical position.
These statements are not target spatial relations to be inferred from the scene; they are auxiliary geometric facts about the camera poses.
We therefore extract them before code generation and apply them to the reconstructed camera states used by the spatial engine.

\paragraph{Constraint extraction.}
Given the query \(Q\), the extractor returns a set of structured pose constraints.
We use a lightweight keyword filter to avoid calling the extractor when \(Q\) does not mention camera-pose information.
When the filter fires, a VLM-based extractor classifies candidate clauses as either stated camera facts or hypothetical observer motions.
Only stated camera facts are extracted.
For instance, ``image 2 was taken after rotating the camera \(90^\circ\) from image 1'' is extracted, while ``if I stand at image 2 and turn right'' is treated as an imagined observer motion and ignored.

The current implementation supports two constraint types:
\[
\mathcal{K}=\mathcal{K}_r\cup\mathcal{K}_p .
\]
A rotation constraint \((u,v,\theta)\in\mathcal{K}_r\) states that camera \(v\)'s orientation is related to camera \(u\)'s orientation by a yaw angle \(\theta\).
A same-position constraint \(S\in\mathcal{K}_p\) states that all cameras in \(S\) share the same world-frame translation.
Extracted constraints are schema-validated before application; records with invalid camera indices or malformed yaw values are discarded.

\paragraph{Rotation propagation.}
Let \(\Gamma^0=\{(\mathbf{x}_v^0,\mathbf{R}_v^0)\}_{v=1}^{V}\) be the camera states produced by image-based reconstruction.
\method selects a single anchor camera \(a\): the lowest-indexed camera appearing in any rotation constraint. Yaw offsets are composed along the constraint graph by a breadth-first traversal from \(a\).
The anchor rotation is initialized from \(\mathbf{R}_a^0\).

When multiple constrained cameras provide evidence for the anchor, \method can refine the anchor by back-projecting each camera's image-derived rotation to an implied anchor rotation.
For a camera \(v\) connected to \(a\), let \(\theta_{a\rightarrow v}\) be the composed yaw along the path from \(a\) to \(v\).
The implied anchor rotation from camera \(v\) is
\[
\hat{\mathbf{R}}_a^{(v)}
=
\mathbf{R}_{\mathrm{yaw}}(-\theta_{a\rightarrow v})
\mathbf{R}_v^0 .
\]
The anchor rotation \(\tilde{\mathbf{R}}_a\) is obtained by averaging these implied rotations on \(\mathrm{SO}(3)\)\ with uniform weights.
If this averaging step is disabled, \(\tilde{\mathbf{R}}_a=\mathbf{R}_a^0\).

The remaining cameras reachable from the anchor are then assigned by forward propagation:
\[
\tilde{\mathbf{R}}_v
=
\mathbf{R}_{\mathrm{yaw}}(\theta_{a\rightarrow v})
\tilde{\mathbf{R}}_a .
\]
Cameras not reachable from any rotation constraint keep their original image-derived orientations.

\paragraph{Same-position constraints.}
Same-position constraints modify translations but not rotations.
For each group \(S\in\mathcal{K}_p\), \method moves every camera in the group onto the position of its lowest-indexed member. Since spatial predicates depend only on relative positions, the choice of the shared location does not change predicate scores; using the lowest-indexed member keeps camera \(0\) at the origin of the reconstruction frame.
Cameras not included in any same-position group keep their original translations.

\paragraph{Output.}
The output is a refined camera-state set
\[
\tilde{\Gamma}
=
\{(\tilde{\mathbf{x}}_v,\tilde{\mathbf{R}}_v)\}_{v=1}^{V}.
\]
These refined camera states replace the initial camera states in the scene representation, so all downstream spatial predicates and generated programs operate on the corrected camera poses.
This procedure applies text-stated constraints as hard geometric facts rather than as soft penalties; there is no joint optimization objective or loss-weight tradeoff.

\section{Experimental Details}
\label{app:experimental_setting}

\subsection{External Benchmarks}

\subsubsection{MindCube}

Throughout the main paper we refer to this evaluation set simply as MindCube. The specific subset is the 1K split of \citet{yin2025spatial}'s benchmark (also referred to as MindCube-tiny in the original release), which evaluates spatial mental modeling from limited visual observations. The benchmark tests whether a model can construct and manipulate a 3D representation of a scene from a small set of 2D images. We use MindCube as an external multi-view benchmark to evaluate whether \method transfers beyond our controlled \refbench and \puzzlebench benchmarks.

MindCube contains three primary sub-tasks. \textbf{Rotation} evaluates whether a model can infer the complete environment from partial visual information and maintain consistency across sequential views. \textbf{Around} evaluates whether a model can infer the scene from a novel viewpoint, requiring interpolation and extrapolation over an implicit 3D scene representation. \textbf{Among} evaluates whether a model can reason about 3D spatial arrangements from four orthogonal views with substantial occlusion, requiring consistency across perspectives and relative-position reasoning over partially hidden objects.

\subsubsection{MMSI-Bench}
\label{sec:mmsi}

MMSI-Bench~\citep{yang2025mmsi} evaluates multi-image spatial reasoning, where models must integrate information across multiple views rather than answer from a single image.
We use MMSI as a real-world transfer benchmark to test whether \method's explicit spatial predicates and symbolic composition remain useful outside our controlled \bench setting.
Following the original benchmark, we report performance on four subcategories:
\textit{Positional Relationships}, which evaluates relative positions among objects, cameras, and semantic regions across views;
\textit{Motion}, which tests reasoning about object or camera movement;
\textit{Attribution}, which evaluates spatially relevant object properties such as shape, size, or length;
and \textit{Multi-Step Reasoning}, which requires chaining multiple spatial cues to answer a query.
Together, these categories provide a complementary real-world evaluation of multi-view spatial understanding and compositional reasoning.

\subsection{Implementation Details}

\method is evaluated without task-specific training or fine-tuning.
The program generator is provided with API documentation and in-context examples; therefore, the setting is zero-shot with respect to benchmark supervision but not prompt-free.
We use Qwen3-VL-8B-Instruct~\citep{qwen2025qwen3vl} for visual interpretation and semantic scoring, and DeepSeek-V4-Flash~\citep{deepseekai2026deepseekv4highlyefficientmilliontoken} in non-thinking mode for Python program generation.
Unless otherwise stated, program generation uses greedy decoding with temperature \(0\) and a maximum output length of 16,000 tokens.

For perception, we use SAM3~\citep{sam3} for segmentation, VGGT~\citep{wang2025vggt} as the default depth and 3D reconstruction module, and OAv2~\citep{wang2026orientv2unifyingorientation} for object orientation estimation.
VGGT is used as the default geometry backbone to keep the comparison with prior 3D visual-programming methods fair.
The spatial engine converts the resulting object and camera states into soft perspective-aware predicate tensors, which are composed by the Pythonic symbolic executor.

All \method variants use the same perception modules, program-generation setup, and symbolic executor unless explicitly stated.
Spatial predicate margins, temperatures, and symbolic composition operators are fixed across benchmarks and are not tuned per dataset.
For hard symbolic variants, predicate scores are thresholded at \(0.5\).
For experiments reporting mean and standard deviation, we run \method three times and report mean \(\pm\) standard deviation.
For paired comparisons between two variants evaluated on the same items, we report the difference in accuracy with a 95\% confidence interval from a bootstrap over items with 10,000 resamples, and a $p$-value from an exact McNemar test on the discordant pairs. Pairing on items removes between-item variance, so these intervals are tighter than the run-to-run standard deviations above and answer a different question: whether one variant beats the other on the same questions, not how much either varies across runs.
Variation across runs comes from nondeterminism in perception and model-serving backends; decoding settings are kept fixed.

We use vLLM 0.19.1 for VLM inference and Ray for scalable serving of tool calls.
The main \method pipeline runs on four GPUs in our implementation: 2 GPUs for VLM inference and 2 GPUs for perception/tool execution.
For throughput, the full experimental sweep was run on an 8-H100 node. The per-query cost in Table~\ref{tab:latency} was measured on A6000 GPUs.
The total compute budget for each MindCube and MMSI is approximately 24 GPU-hours, including perception, VLM inference, and program generation.
For proprietary or API-based baselines, model sizes and infrastructure are not disclosed by the providers; for open-weight models, we report the public model names and parameter scales in the main experiments.

\section{Additional Results}

\subsection{Backbone VLM and Reasoning Mode}\label{app:backbone}
\begin{figure}[h]
\centering
\includegraphics[width=\columnwidth]{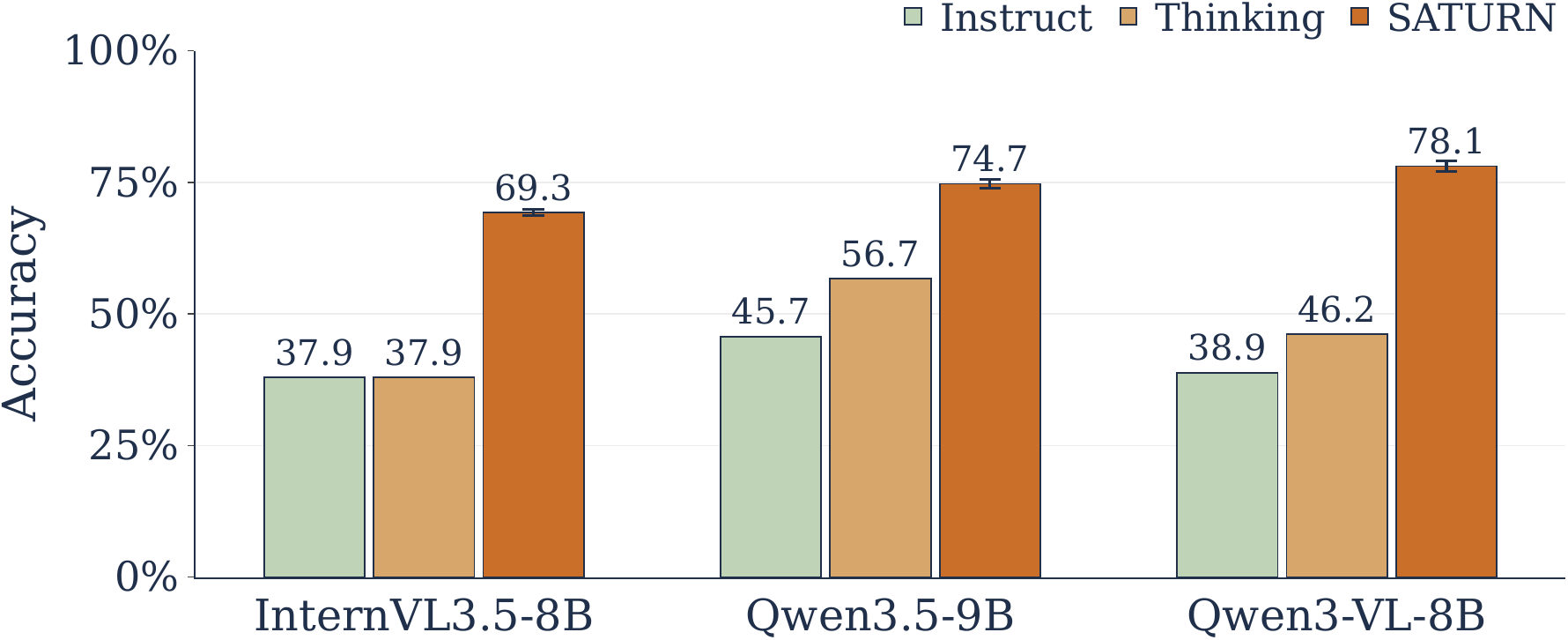}
\caption{Effect of VLM backbone and reasoning mode on MindCube. Instruct and Thinking are single runs; the \method{} bar is the mean over three runs with the standard deviation as the error bar (0.54, 0.91, and 1.05 points for InternVL3.5-8B, Qwen3.5-9B, and Qwen3-VL-8B).}
\label{fig:reasoning_mode_ablation}
\end{figure}

\noindent\textbf{Backbone VLM.}
Figure~\ref{fig:reasoning_mode_ablation} compares three ways of using the same VLM families on MindCube: CoT prompting of instruction model, thinking-mode, and using the VLM as the perception backbone inside \method. Thinking mode provides limited gains over instruction prompting: it improves the Qwen-based models, but gives no gain for InternVL3.5. In contrast, \method substantially improves all three backbones, with minimum \(+29\) percentage points gain. These results suggest that the main improvement comes from explicit 3D predicate computation and symbolic composition, rather than from stronger language-side reasoning alone. At the same time, the remaining variation across backbones indicates that \method is still affected by the quality of the underlying VLM perception.

\subsection{Images and Camera Frames}\label{app:nk}

\begin{figure*}[t]
    \centering
    \includegraphics[width=\textwidth]{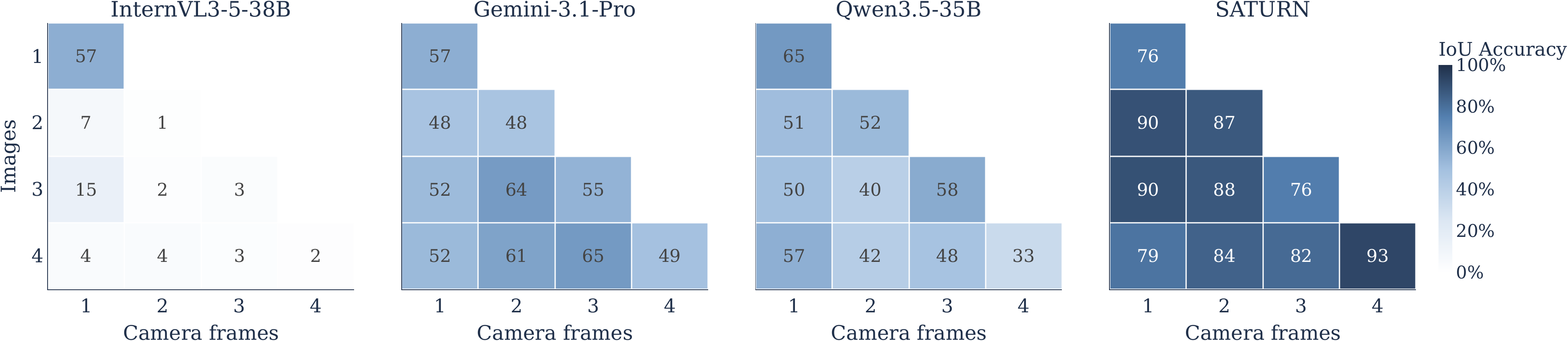}
    \caption{Accuracy on \refbenchshort{} by number of images (rows) and number of camera frames of reference the query composes (columns), on camera-perspective queries. Cells above the diagonal do not exist.}
    \label{fig:nk_heatmap}
\end{figure*}

\begin{figure*}[t]
    \centering
    \includegraphics[width=\textwidth]{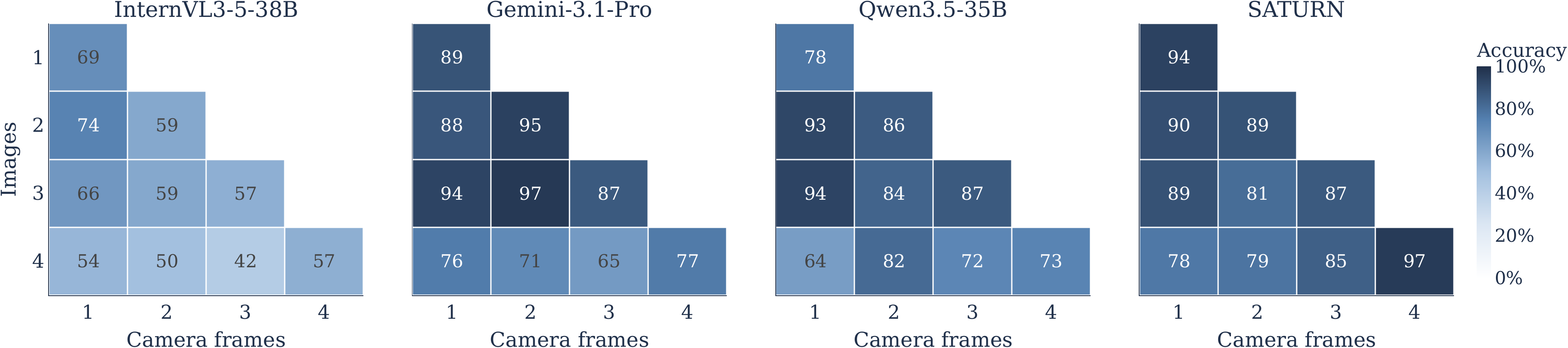}
    \caption{Accuracy on \puzzlebenchshort{} by number of images (rows) and number of camera frames of reference composed (columns), on camera-perspective queries.}
    \label{fig:nk_heatmap_sag}
\end{figure*}

Two scale axes are separate from reasoning depth: the number of images a query spans, and the number of camera frames of reference it composes. They are coupled in the benchmark, since a query composes at most as many frames as it has images, so Figures~\ref{fig:nk_heatmap} and~\ref{fig:nk_heatmap_sag} hold both visible. Both restrict to camera-perspective queries, so the column is an exact frame count; pooling all perspectives would make the axis ambiguous, because the zero-frame column holds both object-anchored and camera-anchored queries.

On \refbenchshort, \method{} shows no systematic trend on either axis, staying between \(76\%\) and \(93\%\). The VLMs share no pattern. InternVL3.5-38B falls from \(57\%\) with a single image to at most \(15\%\) with more, a step at the input boundary rather than a gradient. Qwen3.5-35B declines along both axes, from \(65\%\) to \(33\%\). Gemini-3.1-Pro stays between \(48\%\) and \(65\%\). The same holds on \puzzlebenchshort.

\label{app:nk_sag}

\subsection{Reasoning Topology}\label{app:topo}

\begin{figure*}[t]
    \centering
    \includegraphics[width=\textwidth]{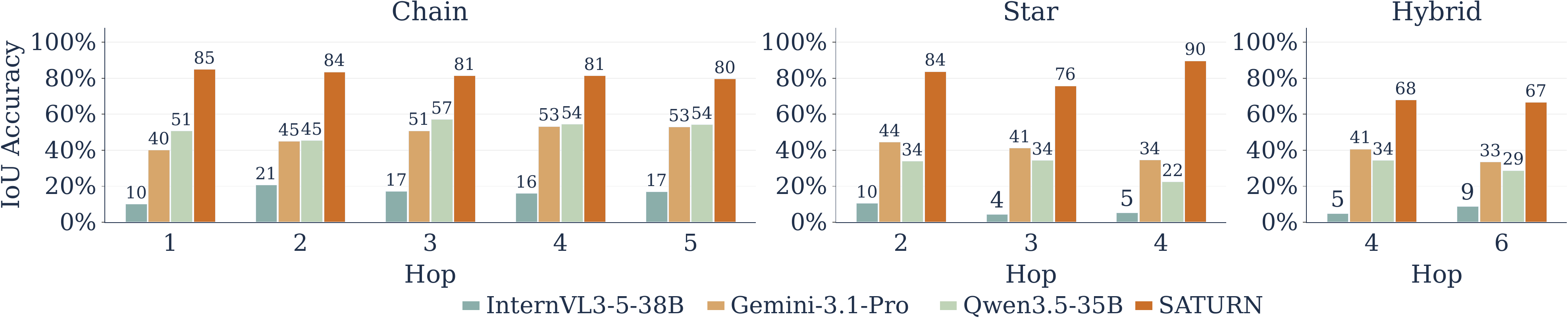}
    \caption{Accuracy on \refbenchshort categorized by the topology structures used to encode multi-hop reasoning, chain (left), star (center), and hybrid (right), evaluated across different reasoning-hop lengths for top-performing models.}
    \label{fig:topo-hop-analysis}
\end{figure*}

Figure~\ref{fig:topo-hop-analysis} separates the topologies pooled in Figure~\ref{fig:hop-analysis}. \textit{Chain} is flat across hop counts, \textit{star} degrades as hops grow for the VLMs, and \textit{hybrid} is the weakest structure for every model. At a fixed four hops, every model scores lower on \textit{hybrid} than on \textit{chain}: \method{} falls from \(81\%\) to \(68\%\), Qwen3.5-35B from \(54\%\) to \(34\%\), Gemini-3.1-Pro from \(53\%\) to \(41\%\), and InternVL3.5-38B from \(16\%\) to \(5\%\).

\subsection{Per-Query Cost by Stage}\label{app:cost}

Table~\ref{tab:stage_cost} breaks down the per-query cost reported in Table~\ref{tab:latency}. Scene estimation dominates, and it can be amortized when several queries concern the same scene.

\begin{table}[h]
\centering
\small
\setlength{\tabcolsep}{5pt}
\renewcommand{\arraystretch}{1.05}
\begin{tabular}{lr}
\toprule
\textbf{Stage} & \textbf{Mean (s)} \\
\midrule
Scene estimation    & 97.0 \\
Program generation  & 5.1 \\
Symbolic execution  & 2.1 \\
Scene assembly      & 0.7 \\
\midrule
Total               & 105 \\
\bottomrule
\end{tabular}
\caption{Per-query cost of \method on MindCube, by stage.}
\label{tab:stage_cost}
\end{table}

\section{Symbolic Baselines}
\label{app:symbolic_baselines}

We compare against three tool-augmented systems. Each was released with its own
backbone and its own perception stack, so a direct comparison would confound the
reasoning method with the models it happens to call. We therefore run GCA and pySpatial with
the same perception backbone as \method, Qwen3-VL-8B-Instruct, and change nothing
else in their reasoning. For TIGeR, we use its own fine-tuned weights. This appendix records what each system does, the
configuration we ran it in, and what we could not match.

\textbf{Object boxes on \refbenchshort.} On \refbenchshort{} we provide all three
tool-augmented baselines with the benchmark's candidate object boxes. Each answers by
selecting a candidate, and we score the selected box against the target.

\subsection{GCA}
GCA~\citep{GCA} formalizes a query into a reference-frame constraint and then has
a vision-language model emit tool calls and Python code against a fixed tool set:
3D reconstruction, open-vocabulary detection, box-to-3D projection, pose estimation,
scale estimation, OCR, and motion analysis.

The released configuration uses Qwen3-VL-235B-A22B-Thinking for both the reasoner
and the code generator. We run it with Qwen3-VL-8B-Instruct as the reasoner and
DeepSeek-V4-Flash as the code generator, matching \method's own backbone and generator,
and route its perception calls to the same SAM3, VGGT, and OAv2 services, so
the two systems see the same detections and the same reconstruction.

\subsection{pySpatial}
pySpatial~\citep{pySpatial} has a code agent compose calls to reconstruction,
camera-pose recovery, and novel-view synthesis. The program does not compute spatial
relations; it renders one or more new views, and a vision-language model answers from
the original images together with the rendered output.

We run it with Qwen3-VL-8B-Instruct as the answering model and DeepSeek-V4-Flash
as the code agent, with VGGT for reconstruction.

\subsection{TIGeR}
TIGeR~\citep{tiger} is a model rather than a prompting method: GLM-4.1V-Thinking~\citep{vteam2026glm45vglm41vthinkingversatilemultimodal}
fine-tuned on 300K tool-invocation traces, so that it emits geometric code and calls
external libraries for exact computation. Because the behavior is trained into the
checkpoint, its backbone cannot be swapped the way the other two can. We serve the
released checkpoint and route its perception calls to the shared VGGT and SAM3
services, which is the only part of its stack we can match to \method.

TIGeR is therefore the one baseline whose reasoning model differs from ours by
design. We include it because it is the closest published system that computes
geometry rather than estimating it, and we mark the mismatch wherever we report its
numbers.

\section{Error Analysis}
\label{app:error_analysis}

\subsection{Object Orientation Error}\label{app:orientation}

Table~\ref{tab:orientation_error} reports orientation accuracy under ground-truth boxes, isolating orientation estimation from detection. Errors are small in the median, and the mean is raised by a small share of front--back flips on categories that are close to front--back symmetric.

\begin{table}[h]
\centering
\small
\setlength{\tabcolsep}{5pt}
\renewcommand{\arraystretch}{1.05}
\begin{tabular}{lr}
\toprule
\textbf{Orientation error (20{,}818 objects)} & \textbf{Value} \\
\midrule
Median yaw error            & $5.4^\circ$ \\
Mean yaw error              & $11.4^\circ$ \\
Mean excluding front--back flips & $6.9^\circ$ \\
Within $22.5^\circ$         & 95.4\% \\
Front--back flips           & 2.7\% \\
\bottomrule
\end{tabular}
\caption{Object orientation error on \refbenchshort, measured under ground-truth boxes.}
\label{tab:orientation_error}
\end{table}

\subsection{Program Reliability}\label{app:program_reliability}

Table~\ref{tab:program_reliability} reports what happens to the 5{,}288 programs \method generated across the four benchmarks. \emph{Parse}, \emph{Run}, and \emph{Answer} are measured after retries, so we also report first-attempt success, the share of queries that needed no regeneration. Program generation is not a bottleneck, because the program only composes predicates from a small API.

\begin{table}[h]
\centering
\small
\setlength{\tabcolsep}{4pt}
\renewcommand{\arraystretch}{1.05}
\begin{tabular}{lrcccc}
\toprule
\textbf{Benchmark} & \textbf{N} & \textbf{Parse} & \textbf{Run} & \textbf{Answer} & \textbf{First try} \\
\midrule
MindCube & 1050 & 100.0 & 100.0 & 99.9  & 99.3 \\
MMSI     & 1000 & 99.7  & 99.4  & 98.7  & 94.5 \\
\puzzlebenchshort & 1150 & 100.0 & 100.0 & 100.0 & 100.0 \\
\refbenchshort    & 2088 & 100.0 & 99.8  & 99.8  & 99.2 \\
\bottomrule
\end{tabular}
\caption{Program reliability (\%). \emph{Parse}, \emph{Run}, and \emph{Answer} are measured after retries; \emph{First try} is the share of queries answered without regenerating the program.}
\label{tab:program_reliability}
\end{table}

\section{Paired Significance of the Baseline Gaps}
\label{app:significance}

Table~\ref{tab:mindcube_mmsi} reports the mean and standard deviation over three \method runs.
Those standard deviations describe run-to-run variation; they do not say whether \method beats a
baseline on the same questions. For that we pair the two systems on the id-intersection of their
predictions and test the difference directly.

We use one \method run per benchmark for the pairing and use the same run for every baseline comparison within that benchmark. For each baseline we report the accuracy
difference on the matched set, a 95\% confidence interval from a paired case-resampling bootstrap
with 10{,}000 resamples, and a two-sided exact McNemar test on the discordant pairs. Because the
pairing removes between-question variance, these intervals are much tighter than the run-to-run
standard deviations, and they answer a different question.

\begin{table}[h]
\centering
\small
\setlength{\tabcolsep}{2.5pt}

\begin{tabular}{lcccc}
\toprule
& \multicolumn{2}{c}{\textbf{MindCube}} & \multicolumn{2}{c}{\textbf{MMSI}} \\
\cmidrule(lr){2-3}\cmidrule(lr){4-5}
\textbf{Baseline} & $\Delta$ & 95\% CI & $\Delta$ & 95\% CI \\
\midrule
GCA\textsubscript{8B} & +46.9 & [43.0, 50.6] & +11.4 & [7.5, 15.4] \\
pySpatial & +26.0 & [22.5, 29.5] & +21.1 & [16.8, 25.3] \\
TIGeR & +52.2 & [48.6, 55.7] & +23.4 & [19.3, 27.5] \\
VLM + scene state & +18.3 & [15.0, 21.5] & +9.5 & [5.9, 13.1] \\
VLM (direct) & +40.3 & [36.7, 43.9] & +19.5 & [15.4, 23.6] \\
\bottomrule
\end{tabular}
\caption{Accuracy difference (\method $-$ baseline) in percentage points on the matched set,
with a 95\% paired-bootstrap interval. Every difference is significant at $p<0.001$ by an exact
McNemar test. $n=1050$ on MindCube and $n=1000$ on MMSI, except pySpatial on MMSI ($n=914$).}
\label{tab:paired_significance}
\end{table}

\section{Qualitative Examples}
\label{app:qualitative}

Here we show six representative samples drawn from
\bench-REF, MindCube, and MMSI: three where \method{} produces the
correct answer on a hard query and three where it fails, followed by a full
end-to-end trace (Example~G).

\definecolor{satgreen}{HTML}{1F9D55}
\definecolor{satred}{HTML}{CC1F1A}
\definecolor{satboxbg}{HTML}{F5F7FA}
\definecolor{satboxrule}{HTML}{B8C2CC}
\definecolor{satboxtitle}{HTML}{2A4365}

\lstdefinestyle{satqualcode}{%
  basicstyle=\scriptsize\ttfamily,%
  language=Python,%
  showstringspaces=false,%
  keywordstyle=\color{magenta},%
  commentstyle=\color{codegreen}\itshape,%
  stringstyle=\color{codepurple},%
  numbers=none,%
  frame=none,%
  framesep=0pt,%
  backgroundcolor=\color{satboxbg!50},%
  breaklines=true,%
  columns=fullflexible,%
  aboveskip=2pt,%
  belowskip=2pt,%
}

\newtcolorbox{satqualbox}[2][satboxtitle]{%
  enhanced,%
  width=\textwidth,%
  colback=white,%
  colframe=#1,%
  boxrule=0.6pt,%
  arc=3pt,%
  left=8pt,right=8pt,top=6pt,bottom=6pt,%
  title=#2,%
  fonttitle=\bfseries,%
  coltitle=white,%
  colbacktitle=#1,%
  toptitle=2pt,bottomtitle=2pt,lefttitle=8pt,righttitle=8pt,%
  titlerule=0pt,%
}

\begin{figure*}[t!]
\centering
\begin{satqualbox}[satgreen]{{Example~A: 3D~FORCE-REF, 6-hop hybrid (success)}}
\textbf{Setup.} 4 cameras, 12 detected objects, 6-hop hybrid topology, every relation FoR-tagged. {\small\textit{(benchmark: 3D~FORCE-REF; sample id: \texttt{hybrid\_d6\_multiview\_4\_multi\_cam\_4\_with\_bbox\_q108})}}

\vspace{4pt}
\noindent
\textbf{Question.} \textit{Find the yellow object that is behind the green object (camera 2 view) and to the left of the red object (the red object's view), where the green object is behind the brown object (camera 3 view) and the brown object is to the left of the red double-decker bus (camera 3 view) and the red object is in front of the sedan car (camera 0 view) and the sedan car is in front of the scooter motorcycle (camera 1 view).}

\vspace{6pt}
\noindent
\includegraphics[width=\linewidth]{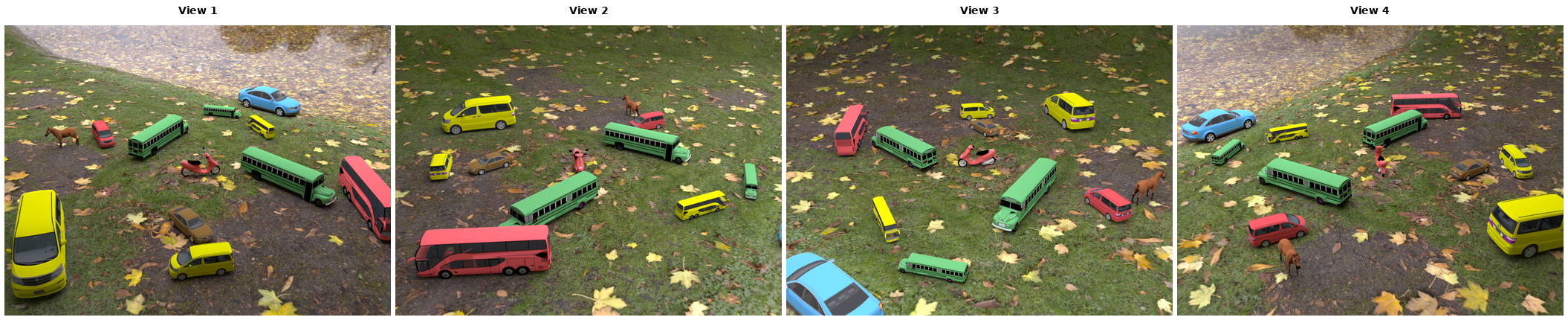}

\vspace{4pt}
{\footnotesize\textbf{Planner groundings.}}

{\footnotesize
\setlength{\tabcolsep}{4pt}
\renewcommand{\arraystretch}{0.95}
\begin{tabular}{@{}p{0.20\linewidth} p{0.70\linewidth} @{\hspace{4pt}}c@{}}
\toprule
\textbf{Phrase} & \textbf{Description} & \textbf{Unique} \\
\midrule
objects in the image & - & \textcolor{satred}{\xmark} \\
\bottomrule
\end{tabular}
}

\vspace{6pt}

    \begin{lstlisting}[style=satqualcode]
    is_yellow = score("is the object in the red bounding box a yellow object?")
    is_green = score("is the object in the red bounding box a green object?")
    is_red = score("is the object in the red bounding box a red object?")
    is_brown = score("is the object in the red bounding box a brown object?")
    is_bus = score("is the object in the red bounding box a double-decker bus?")
    is_sedan = score("is the object in the red bounding box a sedan car?")
    is_scooter = score("is the object in the red bounding box a scooter motorcycle?")

    cam2_behind = scene.cameras[2].third_person.behind
    cam3_behind = scene.cameras[3].third_person.behind
    cam3_left = scene.cameras[3].third_person.left
    cam0_front = scene.cameras[0].third_person.front
    cam1_front = scene.cameras[1].third_person.front

    scooter = is_scooter("x1").iota("x1")
    sedan = (is_sedan("x2") & cam1_front("x2", "x1") & scooter("x1")).iota("x2")
    red_obj = (is_red("x3") & cam0_front("x3", "x2") & sedan("x2")).iota("x3")
    brown_obj = (is_brown("x4") & cam3_left("x4", "x5") & is_bus("x5") & cam3_behind("x6", "x4") & is_green("x6")).iota("x4")
    green_obj = (is_green("x6") & cam3_behind("x6", "x4") & brown_obj("x4")).iota("x6")

    target = (is_yellow("x7") & cam2_behind("x7", "x6") & green_obj("x6") & scene.obj_left("x7", "x3") & red_obj("x3")).iota("x7")
    return int(target.argmax())
\end{lstlisting}

\vspace{2pt}
\noindent
\textbf{\method{} answer:} \texttt{11} \quad
\textbf{Ground truth:} \texttt{box [435, 573, 608, 675]} \quad
\textcolor{satgreen}{\cmark\textbf{Success}}

\vspace{4pt}
\noindent
{\footnotesize\textbf{Index to box.} \refbenchshort is scored by IoU against a
box, while the program returns an entity index. In this setting the scene is
seeded from the released boxes, so entity \(i\) is box \(i\): index \texttt{11}
resolves to \texttt{[435, 573, 608, 675]}, which is the ground-truth box.}
\end{satqualbox}
\end{figure*}

\begin{figure*}[t!]
\centering
\begin{satqualbox}[satred]{{Example~B: 3D~FORCE-REF, 2-hop star (failure)}}
\textbf{Setup.} 2 cameras, 7 detected objects, 2-hop star topology, both anchors object-centric. {\small\textit{(benchmark: 3D~FORCE-REF; sample id: \texttt{star\_d2\_multiview\_2\_multi\_cam\_0\_q1247})}}

\vspace{4pt}
\noindent
\textbf{Question.} \textit{Find the green object that is in front of the red sedan car (the red sedan car's view) and to the left of the tank (the tank's view).}

\vspace{6pt}
\noindent
\includegraphics[width=\linewidth]{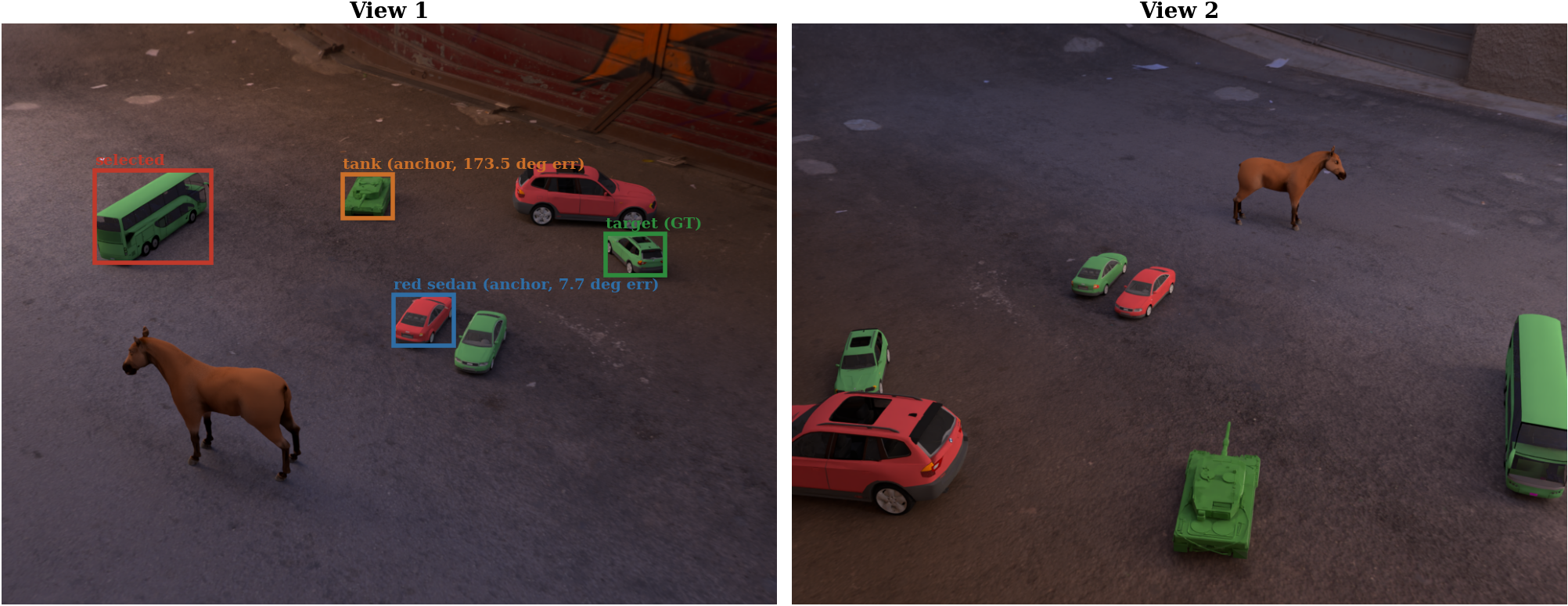}

\vspace{6pt}

\begin{lstlisting}[style=satqualcode]
    is_green = score("is the object in the red bounding box a green object?")
    is_red = score("is the object in the red bounding box a red object?")
    is_sedan = score("is the object in the red bounding box a sedan car?")
    is_tank = score("is the object in the red bounding box a tank?")

    red_sedan = (is_red("x1") & is_sedan("x1")).iota("x1")
    tank = is_tank("x1").iota("x1")

    target = (is_green("x3") &
              scene.obj_front("x3", "x1") & red_sedan("x1") &
              scene.obj_left("x3", "x2") & tank("x2")).iota("x3")

    return int(target.argmax())
\end{lstlisting}

\vspace{2pt}
\noindent
\textbf{\method{} answer:} \texttt{box [123, 193, 277, 316]} \quad
\textbf{Ground truth:} \texttt{box [798, 277, 877, 332]} \quad
\textcolor{satred}{\xmark\textbf{Failure}}

\vspace{4pt}
\noindent
\textbf{Diagnosis.} The program is correct; the scene it reads is not. One anchor is accurate and the other is inverted: the red sedan's estimated front azimuth is off by $7.7^{\circ}$, while the tank's is off by $173.5^{\circ}$ ($76.2^{\circ}$ ground truth against $262.7^{\circ}$ estimated). Every other object in the scene is within $8^{\circ}$. The accurate anchor does not discriminate, because both green candidates lie in front of the red sedan in the ground-truth and estimated scenes alike. The flipped anchor does: the target lies $5.6$\,m to the tank's left in the ground-truth scene but $4.5$\,m to its right in the estimated one, and the selected double-decker bus inverts the same way, so the program returns it. The tank's estimated front is nearly parallel to the optical axis in both views, the configuration in which front and back are least separable.
\end{satqualbox}
\end{figure*}

\begin{figure*}[t!]
\centering
\begin{satqualbox}[satgreen]{{Example~C: MindCube Among, 4-view perspective shift (success)}}
\textbf{Setup.} 4 real-world room photos at 90$^{\circ}$ rotations; query references image~2's viewpoint explicitly. {\small\textit{(benchmark: MindCube-Among; sample id: \texttt{among\_group525\_q1\_2\_1})}}

\vspace{4pt}
\noindent
\textbf{Question.} \textit{Based on these four images (image 1, 2, 3, and 4) showing the black chair from different viewpoints (front, left, back, and right), with each camera aligned with room walls and partially capturing the surroundings: If I am standing at the same spot and facing the same direction as shown in image 2, what is behind me? A. Grey sofa B. Office area C. Two chairs on the corridor D. Cabinet desk along a corridor}

\vspace{6pt}
\noindent
\includegraphics[width=\linewidth]{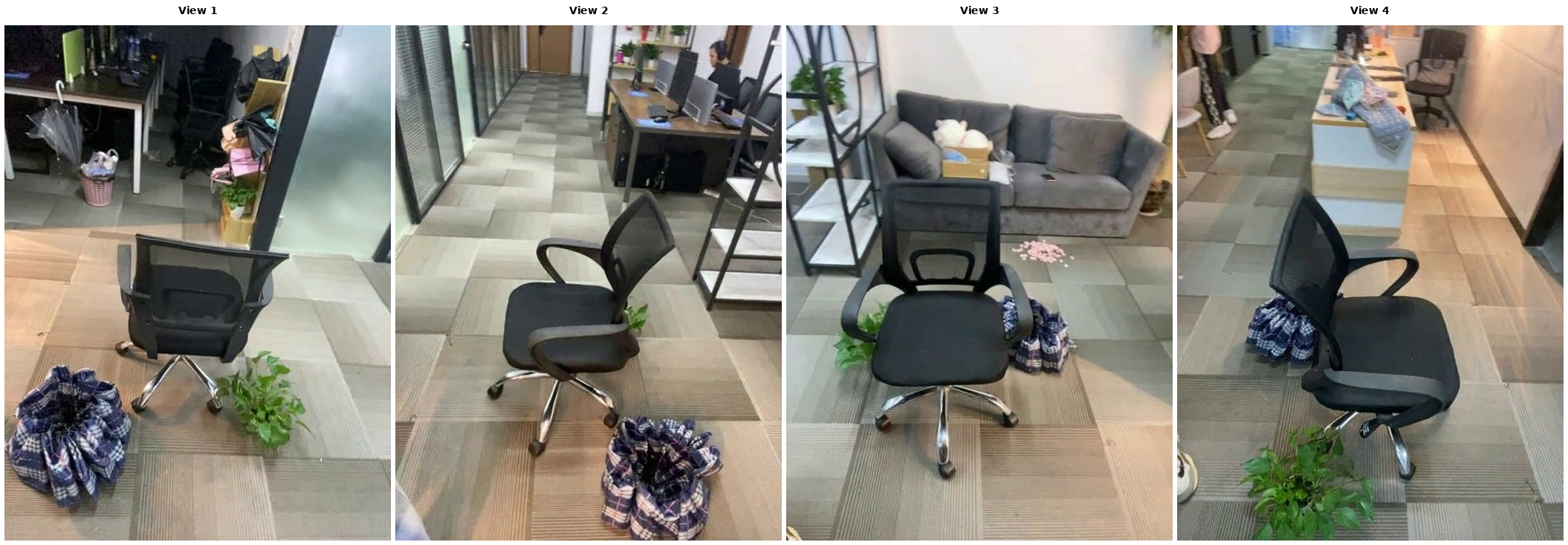}

\vspace{4pt}
{\footnotesize\textbf{Planner groundings.}}

{\footnotesize
\setlength{\tabcolsep}{4pt}
\renewcommand{\arraystretch}{0.95}
\begin{tabular}{@{}p{0.20\linewidth} p{0.70\linewidth} @{\hspace{4pt}}c@{}}
\toprule
\textbf{Phrase} & \textbf{Description} & \textbf{Unique} \\
\midrule
black chair & \textit{black mesh-back office chair with chrome base and wheels, viewed from the front} & \textcolor{satgreen}{\cmark} \\
grey sofa & \textit{large grey upholstered sofa with cushions, positioned against a wall} & \textcolor{satgreen}{\cmark} \\
office area & \textit{workspace with desks, computers, and office chairs in the background} & \textcolor{satgreen}{\cmark} \\
two chairs on the corridor & \textit{two office chairs side by side in a hallway, visible in the background} & \textcolor{satgreen}{\cmark} \\
cabinet desk along a corridor & \textit{long white cabinet desk with items on top, situated along a hallway} & \textcolor{satgreen}{\cmark} \\
\bottomrule
\end{tabular}
}

\vspace{6pt}

\begin{lstlisting}[style=satqualcode]

    # Allocentric anchor at camera 1 (image 2, 0-indexed)
    anchor = scene.frame(position=scene.cameras[1].position,
                         orientation=scene.cameras[1].orientation)

    # INVERSE Rule 1: geometric argmax FIRST - pick the scene object most behind me
    target_idx = anchor.first_person.behind[:scene.objects_count].argmax()

    # Score each option phrase against the chosen target's box (RAW score, target is fixed)
    options = {"A": "grey sofa", "B": "office area", "C": "two chairs on the corridor", "D": "cabinet desk along a corridor"}
    return max(options, key=lambda k:
               float(score(f"is the object in the red bounding box a {options[k]}?")[target_idx]))
\end{lstlisting}

\vspace{2pt}
\noindent
\textbf{\method{} answer:} \texttt{D} \quad
\textbf{Ground truth:} \texttt{D} \quad
\textcolor{satgreen}{\cmark\textbf{Success}}

\vspace{4pt}
\noindent
\textbf{Explanation.} The question pins the reasoning anchor to image~2 (camera index~1). The generated program constructs the anchor as \texttt{scene.frame(position=scene.cameras[1].position, orientation=scene.cameras[1].orientation)} and then takes the geometric argmax along the anchor's \emph{behind} axis. This is exactly the anchor-conditioned predicate interface the method paper claims as its central architectural primitive - a single call constructs a virtual viewer, after which standard predicates are applied.
\end{satqualbox}
\end{figure*}

\begin{figure*}[t!]
\centering
\begin{satqualbox}[satred]{{Example~D: MindCube Rotation, 3-view 180$^{\circ}$ flip (failure)}}
\textbf{Setup.} 3 real-world bedroom photos at 0/90/180$^{\circ}$; text-stated rotation constraints extracted. {\small\textit{(benchmark: MindCube-Rotation; sample id: \texttt{rotation\_group005\_q1\_2})}}

\vspace{4pt}
\noindent
\textbf{Question.} \textit{These three images (image 1, 2, and 3) show the same scene from three different viewpoints. The image 2 was taken after turning the camera 90 degrees to the right (clockwise) from the position of image 1. For image 3, the camera was turned another 90 degrees right, so it's basically facing the opposite direction of image 1. Based on these three images: If I am standing at the same spot and facing the same direction as shown in image 1, what is to my behind? A. Study table and black chair B. Window C. Bed}

\vspace{6pt}
\noindent
\includegraphics[width=\linewidth]{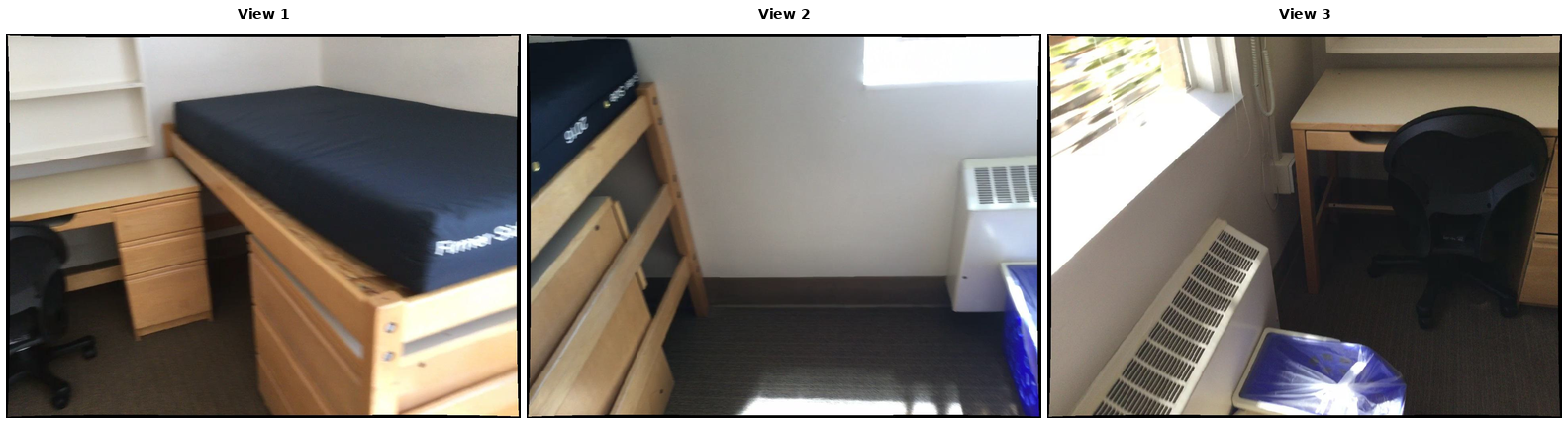}

\vspace{4pt}
{\footnotesize\textbf{Planner groundings.}}

{\footnotesize
\setlength{\tabcolsep}{4pt}
\renewcommand{\arraystretch}{0.95}
\begin{tabular}{@{}p{0.20\linewidth} p{0.70\linewidth} @{\hspace{4pt}}c@{}}
\toprule
\textbf{Phrase} & \textbf{Description} & \textbf{Unique} \\
\midrule
study table and black chair & \textit{light-wood desk with a black office chair tucked under it} & \textcolor{satgreen}{\cmark} \\
window & \textit{window with blinds casting bright light, located on the left wall} & \textcolor{satgreen}{\cmark} \\
bed & \textit{wooden bunk bed with a dark blue mattress, positioned against the right wall} & \textcolor{satgreen}{\cmark} \\
\bottomrule
\end{tabular}
}

\vspace{6pt}

\begin{lstlisting}[style=satqualcode]
    anchor = scene.frame(position=scene.cameras[0].position,
                         orientation=scene.cameras[0].orientation)
    target_idx = anchor.first_person.behind[:scene.objects_count].argmax()

    options = {"A": "study table and black chair", "B": "window", "C": "bed"}
    return max(options, key=lambda k:
               float(score(f"is the object in the red bounding box a {options[k]}?")[target_idx]))
\end{lstlisting}

\vspace{2pt}
\noindent
\textbf{\method{} answer:} \texttt{B} \quad
\textbf{Ground truth:} \texttt{A} \quad
\textcolor{satred}{\xmark\textbf{Failure}}

\vspace{4pt}
\noindent
\textbf{Diagnosis.} The pose-constraint extractor correctly captured the rotations (yaw $90^{\circ}$ from cam~0 to cam~1 and $180^{\circ}$ from cam~0 to cam~2), and the planner correctly anchored at camera~0. The failure occurs at the perception layer: from camera~0's viewpoint, the geometric argmax along the \emph{behind} axis matches ``window'' in view 3 rather than the ground-truth option~A (``study table and black chair'').
\end{satqualbox}
\end{figure*}

\begin{figure*}[t!]
\centering
\begin{satqualbox}[satgreen]{{Example~E: MMSI Multi-Step Reasoning, 8 cameras (success)}}
\textbf{Setup.} 8 consecutive first-person frames; cardinal direction is defined relative to an object's intrinsic front (the nightstand). {\small\textit{(benchmark: MMSI-MSR; sample id: \texttt{298})}}

\vspace{4pt}
\noindent
\textbf{Question.} \textit{Where is the trash can in the room if the nightstand near the window faces north relative to the TV?
Options: A: Near the east wall, B: Southeast corner, C: Near the west wall, D: Northwest corner}

\vspace{6pt}
\noindent
\includegraphics[width=\linewidth]{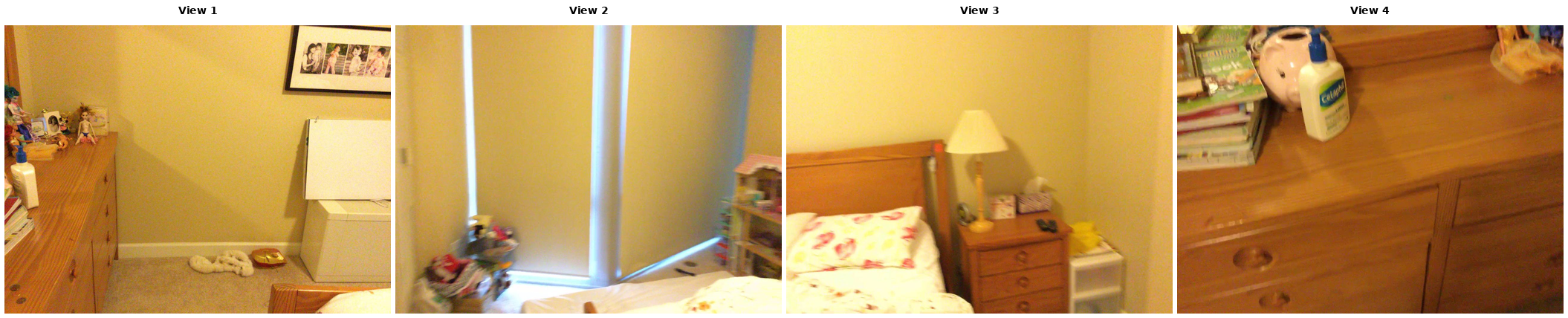}

\vspace{4pt}
{\footnotesize\textbf{Planner groundings.}}

{\footnotesize
\setlength{\tabcolsep}{4pt}
\renewcommand{\arraystretch}{0.95}
\begin{tabular}{@{}p{0.20\linewidth} p{0.70\linewidth} @{\hspace{4pt}}c@{}}
\toprule
\textbf{Phrase} & \textbf{Description} & \textbf{Unique} \\
\midrule
TV & \textit{black flat-screen television on top of a wooden dresser} & \textcolor{satgreen}{\cmark} \\
nightstand & \textit{wooden nightstand with a lamp and tissues, next to the bed} & \textcolor{satgreen}{\cmark} \\
window & \textit{large window with vertical blinds, letting in bright light} & \textcolor{satgreen}{\cmark} \\
trash can & \textit{green plastic trash can with a lid, on the floor} & \textcolor{satgreen}{\cmark} \\
\bottomrule
\end{tabular}
}

\vspace{6pt}

\begin{lstlisting}[style=satqualcode]

    # Find the TV and nightstand
    tv_idx = score("is the object in the red bounding box a black flat-screen television on top of a wooden dresser?").iota("x1").argmax()
    nightstand_idx = score("is the object in the red bounding box a wooden nightstand with a lamp and tissues, next to the bed?").iota("x1").argmax()
    trash_can_idx = score("is the object in the red bounding box a green plastic trash can with a lid, on the floor?").iota("x1").argmax()

    # The nightstand faces north relative to the TV
    # This means the direction from TV to nightstand is north
    north_vec = scene.objects[nightstand_idx].position - scene.objects[tv_idx].position
    scene.set_cardinal_vector(north_vec)

    # Build anchor at the TV with its intrinsic orientation
    anchor = scene.frame(
        position=scene.objects[tv_idx].position,
        orientation=scene.objects[tv_idx].orientation
    )

    # Score each cardinal direction for the trash can
    options = {"A": "east", "B": "south-east", "C": "west", "D": "north-west"}
    return max(options, key=lambda k: float(anchor.first_person(options[k])[trash_can_idx]))
\end{lstlisting}

\vspace{2pt}
\noindent
\textbf{\method{} answer:} \texttt{A} \quad
\textbf{Ground truth:} \texttt{A} \quad
\textcolor{satgreen}{\cmark\textbf{Success}}

\vspace{4pt}
\noindent
\textbf{Explanation.} The query defines north as ``the direction the nightstand faces relative to the TV''. The code generator interprets this as a position-difference cardinal binding, emitting \texttt{north\_vec = nightstand.position $-$ tv.position} and binding the world-frame \emph{north} axis to this vector. The TV-anchored frame is then constructed, and the trash-can's position is evaluated against the cardinal axes derived from this north. This matches the MMSI ground-truth convention for object-anchored cardinal bindings, and is the interpretation that Example~F fails to reproduce on a structurally identical query.
\end{satqualbox}
\end{figure*}

\begin{figure*}[t!]
\centering
\begin{satqualbox}[satred]{{Example~F: MMSI Multi-Step Reasoning, 8 cameras (failure)}}
\textbf{Setup.} 8 consecutive frames, 4 detected objects; cardinal direction is defined by the same template as Example~E (\emph{X near window faces north relative to Y}), so the contrast pair holds query structure fixed. {\small\textit{(benchmark: MMSI-MSR; sample id: \texttt{223})}}

\vspace{4pt}
\noindent
\textbf{Question.} \textit{The bedside table near the window faces north relative to the television; where is the light switch in the room located?
Options: A: Southeast corner, B: Northwest corner, C: Near the east wall, D: Near the west wall}

\vspace{6pt}
\noindent
\includegraphics[width=\linewidth]{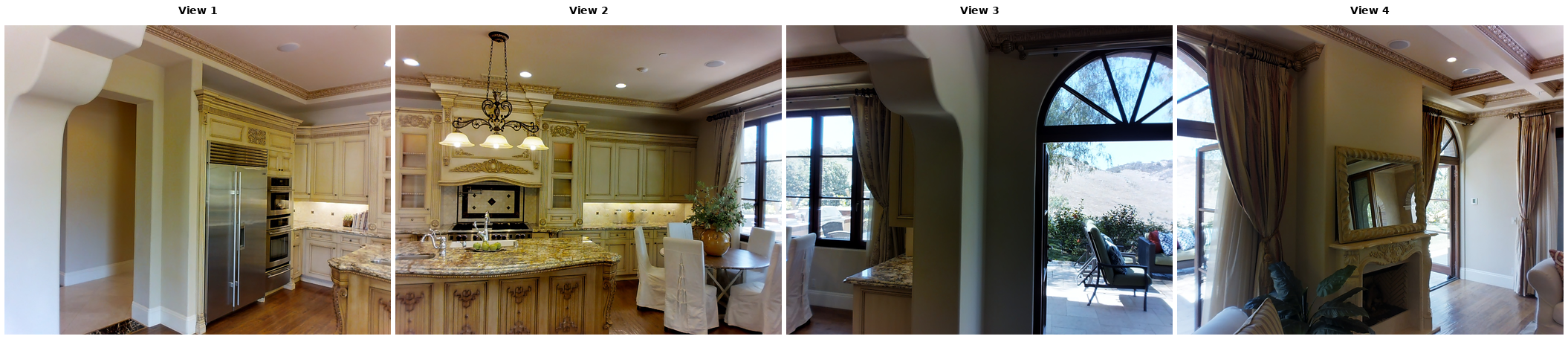}

\vspace{4pt}
{\footnotesize\textbf{Planner groundings.}}

{\footnotesize
\setlength{\tabcolsep}{4pt}
\renewcommand{\arraystretch}{0.95}
\begin{tabular}{@{}p{0.20\linewidth} p{0.70\linewidth} @{\hspace{4pt}}c@{}}
\toprule
\textbf{Phrase} & \textbf{Description} & \textbf{Unique} \\
\midrule
television & \textit{black flat-screen TV on top of a wooden dresser} & \textcolor{satgreen}{\cmark} \\
bedside table & \textit{wooden nightstand with a lamp and tissues, next to the bed} & \textcolor{satgreen}{\cmark} \\
window & \textit{large window with vertical blinds, letting in bright light} & \textcolor{satgreen}{\cmark} \\
light switch & \textit{white rectangular switch plate on a beige wall, near a doorway} & \textcolor{satgreen}{\cmark} \\
\bottomrule
\end{tabular}
}

\vspace{6pt}

\begin{lstlisting}[style=satqualcode]

    # Get indices for the objects
    bedside_table_idx = score("is the object in the red bounding box a bedside table?").iota("x1").argmax()
    television_idx = score("is the object in the red bounding box a television?").iota("x1").argmax()
    light_switch_idx = score("is the object in the red bounding box a light switch?").iota("x1").argmax()

    # The bedside table faces north, so its front vector is north
    scene.set_cardinal_vector(scene.objects[bedside_table_idx].front_vec)

    # Build anchor at the television with its intrinsic orientation
    anchor = scene.frame(
        position=scene.objects[television_idx].position,
        orientation=scene.objects[television_idx].orientation
    )

    # Score each cardinal direction for the light switch
    options = {"A": "south-east", "B": "north-west", "C": "east", "D": "west"}
    return max(options, key=lambda k: float(anchor.first_person(options[k])[light_switch_idx]))
\end{lstlisting}

\vspace{2pt}
\noindent
\textbf{\method{} answer:} \texttt{D} \quad
\textbf{Ground truth:} \texttt{C} \quad
\textcolor{satred}{\xmark\textbf{Failure}}

\vspace{4pt}
\noindent
\textbf{Diagnosis.} The failure is at code generation. For the cardinal-anchor clause ``the bedside table faces north relative to the television'', the code generator emitted \texttt{scene.set\_cardinal\_vector(bedside\_table.front\_vec)} --- using the table's own facing axis as north --- rather than the position-difference interpretation \texttt{table.position $-$ tv.position} that the structurally identical Example~E correctly used. The two programs differ in one line, but the resulting world-frame north rotates by roughly $90^{\circ}$, which is enough to flip east and west in the final geometric resolution.
\end{satqualbox}
\end{figure*}

\begin{figure*}[t!]
\centering
\begin{satqualbox}[satgreen]{{Example~G: MindCube Rotation, end-to-end trace (success)}}
\textbf{Setup.} 3 views, 4 detected objects. {\small\textit{(sample id: \texttt{rotation\_group025\_q1\_4})}}

\vspace{4pt}
\noindent
\textbf{Question.} \textit{The image 2 was taken after turning the camera $90^{\circ}$ to the right from image 1, and image 3 another $90^{\circ}$ right. If I stand where image 1 was taken, facing the same direction, then turn $90^{\circ}$ to the left, what is to my left?} \textbf{A.} sink \quad \textbf{B.} wooden frame mirror \quad \textbf{C.} door

\vspace{5pt}
\noindent
\centerline{\includegraphics[width=0.68\linewidth]{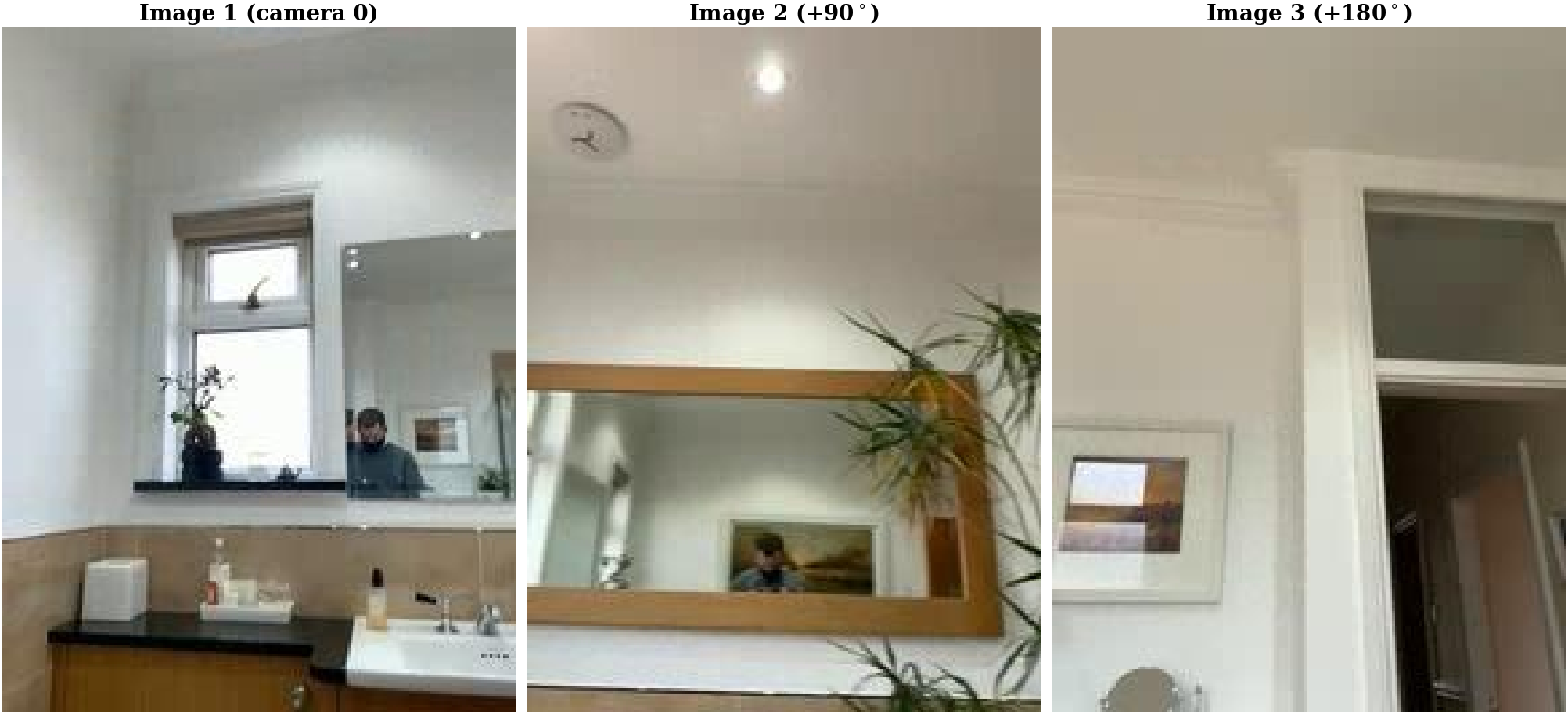}}

\vspace{5pt}
{\footnotesize\textbf{(1) Parser groundings.} Each question phrase is expanded into a description and grounded.}

{\footnotesize
\setlength{\tabcolsep}{4pt}\renewcommand{\arraystretch}{0.95}
\begin{tabular}{@{}p{0.20\linewidth} p{0.58\linewidth} c c@{}}
\toprule
\textbf{Phrase} & \textbf{Description} & \textbf{Verified} & \textbf{Objects} \\
\midrule
sink & white ceramic sink with chrome faucet on a black countertop & \textcolor{satgreen}{\cmark} & \{0\} \\
wooden frame mirror & large rectangular mirror with a light wood frame mounted on the wall & \textcolor{satgreen}{\cmark} & \{1\} \\
door & white interior door with a glass panel, partially open, revealing a hallway & \textcolor{satgreen}{\cmark} & \{2, 3\} \\
\bottomrule
\end{tabular}}

\vspace{5pt}
{\footnotesize\textbf{(2) Pose constraints and estimated scene.} The extractor reads the stated turns from the question and applies them to the cameras: \texttt{rotation(cam\,0$\to$1, yaw $90^{\circ}$)} and \texttt{rotation(cam\,0$\to$2, yaw $180^{\circ}$)}. The table shows the camera headings before and after they are applied.}

{\footnotesize
\setlength{\tabcolsep}{5pt}\renewcommand{\arraystretch}{0.95}
\begin{tabular}{@{}clrrc@{}}
\toprule
\textbf{Id} & \textbf{Label} & \textbf{Centre $(x,y,z)$} & \textbf{Azimuth} & \textbf{Views} \\
\midrule
0 & sink                & $(+0.24, -0.40, +1.08)$ & $-178.7^{\circ}$ & 1 \\
1 & wooden frame mirror & $(+0.81, +0.18, +0.13)$ & $-80.0^{\circ}$ & 2 \\
2 & door (partially open)   & $(-0.57, +0.23, -0.40)$ & $18.4^{\circ}$  & 3 \\
3 & door (hallway behind)   & $(-0.54, +0.26, -0.33)$ & $0.4^{\circ}$ & 3 \\
\midrule
\multicolumn{5}{@{}l@{}}{\textit{camera headings, reconstruction only:} $c_0$ $-0.0^{\circ}$, $c_1$ $101.4^{\circ}$, $c_2$ $-155.1^{\circ}$ \quad (turns $101.4^{\circ}$, $103.5^{\circ}$)} \\
\multicolumn{5}{@{}l@{}}{\textit{after applying the constraints:} $c_0$ $11.6^{\circ}$, $c_1$ $101.6^{\circ}$, $c_2$ $-168.4^{\circ}$ \quad (turns $90.0^{\circ}$, $90.0^{\circ}$)} \\
\bottomrule
\end{tabular}}

\vspace{5pt}
{\footnotesize\textbf{(3) Soft semantic scores.} One score vector per \texttt{score()} call, over the four objects. The door phrase matches objects 2 and 3 equally; thresholding would have to discard one.}

{\footnotesize
\setlength{\tabcolsep}{6pt}\renewcommand{\arraystretch}{0.95}
\begin{tabular}{@{}lrrrr@{}}
\toprule
\texttt{score(\dots)} & \textbf{obj 0} & \textbf{obj 1} & \textbf{obj 2} & \textbf{obj 3} \\
\midrule
``\dots a sink?''                & \textbf{0.971} & 0.024 & 0.064 & 0.023 \\
``\dots a wooden frame mirror?'' & 0.072 & \textbf{0.924} & 0.469 & 0.407 \\
``\dots a door?''                & 0.003 & 0.090 & \textbf{0.905} & \textbf{0.905} \\
\bottomrule
\end{tabular}}

\vspace{6pt}
{\footnotesize\textbf{(4) Generated program.} Boilerplate wrappers elided.}

\begin{lstlisting}[style=satqualcode]
    sink_idx   = score("is the object in the red bounding box a sink?").iota("x1").argmax()
    mirror_idx = score("is the object in the red bounding box a wooden frame mirror?").iota("x1").argmax()
    door_idx   = score("is the object in the red bounding box a door?").iota("x1").argmax()

    # stand at camera 0, face its direction, then turn 90 degrees left
    anchor = scene.frame(position=scene.cameras[0].position,
                         orientation=scene.cameras[0].orientation).rotate(yaw=-90)

    sink_left   = float(anchor.first_person.left[sink_idx])
    mirror_left = float(anchor.first_person.left[mirror_idx])
    door_left   = float(anchor.first_person.left[door_idx])

    options = {"A": sink_left, "B": mirror_left, "C": door_left}
    return max(options, key=options.get)
\end{lstlisting}

\vspace{2pt}
\noindent
\textbf{(5) \method{} answer:} \texttt{C} \quad \textbf{Ground truth:} \texttt{C} \quad \textcolor{satgreen}{\cmark\textbf{Correct}}

\vspace{4pt}
\noindent
\textbf{Trace.} The program computes no geometry: it anchors on camera~0, rotates by the turn the question describes, and reads \texttt{first\_person.left}. Uncertainty survives to that point, which matters because the door grounding split into two objects scoring $0.905$ each.
\end{satqualbox}
\end{figure*}

\paragraph{Patterns across the examples.}
Two observations recur. \emph{First}, on the controlled benchmark
the wins and the failures look almost identical at the planner and
code-generation stages --- the same number of relations, the same
program structure --- so the empirical gap originates inside the
soft predicate layer rather than at the LLM interface. \emph{Second},
on real-world data, perception remains the dominant failure source:
when an abstract anchor (``the kitchen'', ``the office area'') has
to be reified as a 3D point, segmentation noise translates into a
rotated cardinal axis that propagates into the final answer.

\section{Prompt Templates}
\label{app:prompts}

\method uses natural-language prompts to interface its symbolic executor
with external language and vision-language models. We provide the grounder
and pose-constraint extractor prompts below.

\subsection{Grounder Prompt}
\label{app:prompt_grounder}
The grounder prompt drives the VLM that parses the natural-language
question and emits a structured noun-phrase plan: which object
categories and descriptions to detect, and from which camera views.
The prompt contains a compact specification of the expected output
schema together with a small set of in-context examples illustrating
how to extract role-indexed object phrases and assign cam-id cues. The
full template is shown in Figure~\ref{fig:grounder-prompt}.

\begin{figure*}[h]
\centering
\begin{minipage}{\textwidth}
\begin{lstlisting}[style=promptstyle]
You are the planner for a code-generating vision agent.

Two data sources the downstream code reads from:
    scene.objects[i]   - populated by SAM3 from the phrases YOU emit
    scene.cameras[k]   - populated by VGGT. Camera indices are ZERO-indexed;
                       human-facing "image N" / "Figure N" / "Photo N" labels
                       are ONE-indexed.

INDEX INVARIANT (REQUIRED):
  Every view reference is emitted as a PAIR of fields:
      `image`  - 1-indexed (matches the question's wording: image 1, Figure 2, ...)
      `cam_id` - 0-indexed (the actual scene.cameras[k] index)
  The pair MUST satisfy:
      cam_id == image - 1
  Examples:
      image 1 -> cam_id 0     image 2 -> cam_id 1     image 5 -> cam_id 4
  Use null in BOTH fields together when no specific view applies (motion,
  multi-view objects). A pair where only one of {image, cam_id} is null, or
  where cam_id != image - 1, is REJECTED by the validator and you will be
  asked to refine.

Rule: ground exactly the objects the code needs from scene.objects. Nothing more.

Fields per grounding:
  phrase       canonical lowercase noun phrase ("chair") - determiner optional.
               If the same noun appears in two views, emit ONE grounding with
               the relevant view, not two.
  description  short, VISUALLY DISTINCTIVE phrase that uniquely identifies
               THIS specific object instance. Combine 2-4 of color, material,
               shape, size, surface details, position-in-scene cues. NEVER
               echo the phrase itself - description must add visual content.
               This is SAM3's ONLY disambiguation signal.
  image        1-indexed image number where this object is most clearly
               visible. Use null only for motion / ordering / multi-view
               objects tracked across all frames.
  cam_id       0-indexed camera index. MUST equal image - 1. Use null when
               image is null.
  is_region    true for walls / sides / rooms; false (default) for discrete objects.
  unique       false when the question references multiple instances; true otherwise.

Skip-reason tags (use exactly these):
  "option label"      NON-OBJECT text from options (Northeast, A/B/C/D, +X/-Z).
                      Physical objects named as options ("A: chair") -> GROUND.
  "abstract concept"  computed by code, not detected (path, shape, letter, order, pattern)
  "non-object"        pronouns, actions, the-pictures, the-order, your-path,
                      I / me / my / myself / the observer / the photographer / the viewer
                      (the observer is implicit; never ground as an object).
  "self-referential"  the photographer's camera or viewpoint. Includes "camera",
                      "camera in Photo N", "my viewpoint", "camera taking / that
                      took / mounted / capturing the picture", and "vehicle/car with
                      the camera" when the vehicle IS the camera-bearing observer. The
                      camera's pose is scene.cameras[k] - never ground as scene.objects[i].
  "covered elsewhere" only for true noun duplicates ("TV" + "television")

Output ONE JSON block:
{
  "setup_caption": "<A concise, observational description of the setup: what the images depict, how the frames relate to each other (multi-view of one space, sequential frames showing motion, single anchor view, etc.), and any observer/self-reference cue downstream code needs (e.g. the camera IS the vehicle). Describe what is shown and what the question states as a given fact. Do not speculate about the quantities the question is asking for - if the question asks for a direction, do not name one; if it asks for an ordering, do not commit to one; if it asks which option is correct, do not pick. The caption is a museum placard, not a solution.>",
  "program_sketch": "<one sentence naming the operations and the objects they need>",
  "object_groundings": [
    {"phrase", "description", "image": <int or null>, "cam_id": <int or null>, "is_region", "unique"}, ...
  ],
  "skipped_phrases": [ {"phrase", "reason"}, ... ]
}

{examples}

QUESTION:
{question}
\end{lstlisting}
\end{minipage}
\caption{Grounder prompt used by the VLM that extracts noun-phrase groundings and camera assignments from the question.}
\label{fig:grounder-prompt}
\end{figure*}

\subsection{Pose Constraint Extractor}
\label{app:prompt_constraints}
The pose-constraint extractor classifies clauses in the question into
stated camera facts versus hypothetical observer motions, and emits
schema-validated yaw rotations and same-position groups for the
downstream geometric refinement step
(Appendix~\ref{app:pose_constraints}). The full template is shown in
Figure~\ref{fig:constraints-prompt}.

\begin{figure*}[h]
\centering
\begin{minipage}{\textwidth}
\begin{lstlisting}[style=promptstyle]
You read a multi-view spatial-reasoning question and identify CAMERA POSE
RELATIONSHIPS that the question text states explicitly.

Your job is to extract these relationships into a structured JSON list. The
extracted constraints will be used to refine noisy 3D reconstruction estimates
of where the cameras are pointed. You are NOT answering the question - only
mining the question text for camera-relative-pose facts.

=========================================================================
The core distinction: stated camera fact  
=========================================================================

STATED CAMERA FACT  - describes how the cameras themselves were
 positioned or oriented when the images were taken. These are ground truth about the physical setup. EXTRACT.

 Linguistic signatures (grammatical past / passive about the camera):
   - "image 2 was taken after rotating the camera 90 from image 1"
   - "image 3 is the opposite direction of image 1"
   - "the camera location is fixed across all images"
   - "all three views were captured from the same spot"
   - "the four images were shot in a 90 rotation pattern around X"

=========================================================================
What counts as a camera pose constraint
=========================================================================

Two constraint types are supported:

  1. ROTATION
       Camera-to-camera angular relationship stated as a fact.

       -> {"type": "rotation", "from_cam": <int>, "to_cam": <int>, "yaw": <float>}

       Sign convention: positive yaw = clockwise rotation viewed from above.
       "Counter-clockwise" or "to the left" -> negative yaw.
       "Opposite direction" -> yaw = 180.

  2. SAME_POSITION
       Two or more cameras share the same world-frame position (only their
       orientations differ).

       -> {"type": "same_position", "cams": [0, 1, 2]}

       Only emit when the question text states all the listed cameras were
       AT one location. "Different viewpoints", "around the object",
       "from front/back/left/right of X", "captured at the corners of the
       room" all imply DIFFERENT positions - never same_position.

If the question text does not explicitly state any pose relationship between
cameras, emit an empty constraints list.

=========================================================================
Camera indices
=========================================================================

Cameras are 0-indexed:
    "image 1" / "first photo" / "view 1" -> camera 0
    "image 2" / "second photo"           -> camera 1
    "image 3" / "third photo"            -> camera 2
    ...

=========================================================================
Output schema  (think first, then emit constraints)
=========================================================================

Emit a single JSON object with TWO fields, in this order:

{
  "reasoning": "<short, structured thinking>",
  "constraints": [ ... ]
}

The "reasoning" field should:

  1. List every clause in the question that looks like it MIGHT describe a
     camera pose or viewpoint.
  2. For each clause, classify it as STATED-CAMERA-FACT or
     OBSERVER-HYPOTHETICAL using the test above ("if I delete this clause,
     do I still know where the cameras are?").
  3. Conclude with what (if anything) gets extracted.
Keep "reasoning" terse - bullets or one paragraph. No prose padding.

Then "constraints" contains the JSON list. Use the literal field names
"reasoning", "constraints", "type", "from_cam", "to_cam", "yaw", "cams".
No markdown fences, no extra prose outside the JSON object.

{examples}
{questions}
\end{lstlisting}
\end{minipage}
\caption{Pose constraint extractor prompt.}
\label{fig:constraints-prompt}
\end{figure*}

\end{document}